\PassOptionsToPackage{table}{xcolor}
\documentclass[10pt,twocolumn,letterpaper]{article}

\usepackage{cvpr}              % To produce the CAMERA-READY version
\usepackage[accsupp]{axessibility}  % Improves PDF readability for those with disabilities.
\usepackage{threeparttable}
\usepackage{multirow}
\usepackage[table]{xcolor}
\definecolor{cvprblue}{rgb}{0.21,0.49,0.74}
\definecolor{forgerytask}{rgb}{0.404, 0.682, 0.420}
\definecolor{forgerytype}{rgb}{0.486, 0.745, 0.788}
\definecolor{forgerymodel}{rgb}{0.392, 0.604, 0.784}
\usepackage[pagebackref,breaklinks,colorlinks,allcolors=cvprblue]{hyperref}

\def\paperID{6301} % *** Enter the Paper ID here
\def\confName{CVPR}
\def\confYear{2025}

\title{Forensics-Bench: A Comprehensive Forgery Detection Benchmark Suite for Large Vision Language Models}

\author{Jin Wang$^{1,*}$, Chenghui Lv$^{5,4,*}$, Xian Li$^{6,4}$, Shichao Dong$^7$, Huadong Li$^8$, Kelu Yao$^4$, Chao Li$^4$,\\Wenqi Shao$^3$, Ping Luo$^{1, 2, \dag}$\\
$^1$The University of Hong
Kong  $^2$HKU Shanghai Intelligent Computing Research Center\\
$^3$Shanghai AI Laboratory  $^4$Zhejiang Laboratory $^5$Hangzhou Institute for Advanced Study\\
$^6$Zhejiang University $^7$Alibaba, Beijing, China  $^8$MEGVII Technology\\
}

\begin{document}
\maketitle
{
	\renewcommand{\thefootnote}{\fnsymbol{footnote}}
    \footnotetext[1]{Equal contribution (primary contact: wj0529@connect.hku.hk)}
	\footnotetext[2]{Corresponding author}
}

\begin{abstract}
Recently, the rapid development of AIGC has significantly boosted the diversities of fake media spread in the Internet, posing unprecedented threats to social security, politics, law, and etc.
To detect the ever-increasingly \textbf{diverse} malicious fake media in the new era of AIGC, recent studies have proposed to exploit Large Vision Language Models (LVLMs) to design \textbf{robust} forgery detectors due to their impressive performance on a \textbf{wide} range of multimodal tasks.
However, it still lacks a comprehensive benchmark designed to comprehensively assess LVLMs' discerning capabilities on forgery media.
To fill this gap, we present Forensics-Bench, a new forgery detection evaluation benchmark suite to assess LVLMs across massive forgery detection tasks, requiring comprehensive recognition, location and reasoning capabilities on diverse forgeries.
Forensics-Bench comprises $63,292$ meticulously curated multi-choice
visual questions, covering $112$ unique forgery detection types from $5$ perspectives: forgery semantics, forgery modalities, forgery tasks, forgery types and forgery models.
We conduct thorough evaluations on $22$ open-sourced LVLMs and $3$ proprietary models GPT-4o, Gemini 1.5 Pro, and Claude 3.5 Sonnet, highlighting the significant challenges of comprehensive forgery detection posed by Forensics-Bench.
We anticipate that Forensics-Bench will motivate the community to advance the frontier of LVLMs, striving for all-around forgery detectors in the era of AIGC.
The deliverables will be updated \href{https://Forensics-Bench.github.io/}{here}.
\end{abstract}    
\section{Introduction}
\label{sec:intro}

In recent years, with the rapid development of AI-generated content (AIGC) technology \citep{goodfellow2014generative,ho2020denoising,esser2021taming}, the barrier to creating fake media has been significantly lowered for the general public. 
As a result, a large amount of various synthetic media has flooded the Internet, which poses unprecedented threats to politics, 
law, and social security, such as the malicious dissemination of deepfakes \citep{deepfakes,faceswap} and misinformation \citep{luo2021newsclippings}. 
To address such situations, researchers have proposed numerous forgery detection methods \citep{Dong_2023_CVPR,SBI,face-x-ray,amazon,shao2023detecting,guo2023hierarchical}, aiming to filter out synthetic media as much as possible.
Nevertheless, the synthetic media nowadays can be incredibly \textbf{diverse}, which may encompass different modalities, depict various semantics, and be created/manipulated with different AI models, etc. 
Thus, designing a \textbf{generalized} forgery detector with such comprehensive discerning capabilities becomes a crucial and urgent task in this new era of AIGC, posing significant challenges to the research community.

\begin{figure*}[t]
  \centering
   \includegraphics[width=0.92\textwidth]{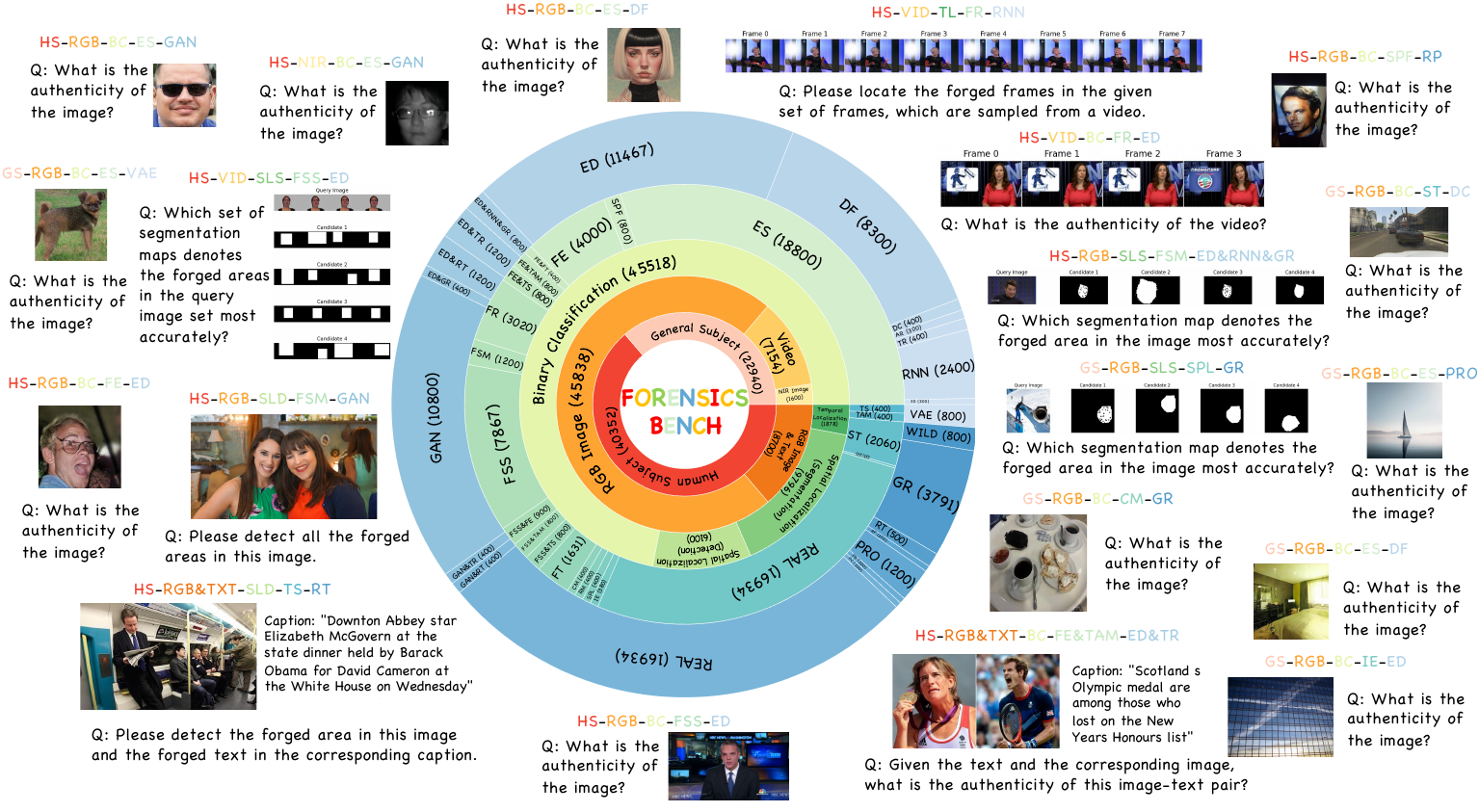}
   \caption{Overview of Forensics-Bench. Forensics-Bench consists of $63K$ samples, covering $112$ unique forgery detection types from five different perspectives characterizing forgeries. As shown in different rings from inside out, these five perspectives include \textbf{\textcolor{red}{forgery semantics}, \textcolor{orange}{forgery modalities}, \textcolor{forgerytask}{forgery tasks}, \textcolor{forgerytype}{forgery types}} and \textbf{\textcolor{forgerymodel}{forgery models}}. From the view of \textbf{\textcolor{red}{forgery semantics}}, Forensics-Bench includes data featuring: \textit{HS $\rightarrow$ Human Subject} and \textit{GS $\rightarrow$ General subject}. From the view of \textbf{\textcolor{orange}{forgery modalities}}, Forensics-Bench includes data featuring: \textit{RGB$\&$TXT$\rightarrow$RGB Image $\&$ Text}, \textit{VID$\rightarrow$Video} and etc. From the view of \textbf{\textcolor{forgerytask}{forgery tasks}}, Forensics-Bench includes data featuring: \textit{TL$\rightarrow$Temporal Localization}, \textit{SLS$\rightarrow$Spatial Localization (Segmentation)}, \textit{BC$\rightarrow$Binary Classification} and etc. From the view of \textbf{\textcolor{forgerytype}{forgery types}}, Forensics-Bench includes data featuring: \textit{TS$\rightarrow$Text Swap}, \textit{FSS$\rightarrow$Face Swap (Single Face)}, \textit{ES$\rightarrow$Entire Synthesis} and etc. From the view of \textbf{\textcolor{forgerymodel}{forgery models}}, Forensics-Bench includes data generated from: \textit{GAN$\rightarrow$Generative Adversarial Models}, \textit{DF$\rightarrow$Diffusion models}, \textit{VAE$\rightarrow$Variational Auto-Encoders} and etc. Forensics-Bench enables comprehensive evaluations of LVLMs on versatile forgery detection types in the evolving era of AIGC. Please see Appendix \ref{sec:abbreviation} for more detailed abbreviations.}
   \label{fig:Forensics-Bench}
\end{figure*}

In the meantime, Large Vision Language Models (LVLMs) \citep{gpt4o,gemini,internvl_chat,liu2023llava,Qwen-VL,internlmxcomposer,zhang2023llamaadapter} have achieved remarkable progress in a \textbf{wide} range of multimodal tasks, such as visual recognition and visual captioning, which reignites the discussion of artificial general intelligence (AGI) \citep{morrisposition}. 
These promising generalization capabilities make LVLMs a compelling solution for distinguishing increasingly diverse synthetic media \cite{chang2024antifakepromptprompttunedvisionlanguagemodels,jin2024fakenewsdetectionmanipulation,jia2024chatgptdetectdeepfakesstudy,qi2024sniffer,liu2024fka,zhang2024common}. 
However, it still lacks a comprehensive evaluation benchmark to assess LVLMs' ability to recognize synthetic media, which hinders the applications of LVLMs on forgery detection and thus further impedes the continuous progress of LVLMs towards the next level of AGI \citep{morrisposition}. 
To this end, a line of work \citep{shi2024shield,li2024fakebench,zhou2024diffusyn,liu2024mmfakebench,wang2024mfc} attempted to bridge this gap with different evaluation benchmarks, but they only cover a limited range of synthetic media, such as out-of-context forgeries \citep{luo2021newsclippings} and diffusion-based forgeries \citep{rombach2022high}, which restricts their comprehensiveness in revealing the full extent of LVLMs' forgery detection capabilities.

To drive the research in this direction, we introduce Forensics-Bench, a new forgery detection benchmark suite for comprehensively evaluating the capabilities of LVLMs in the context of forgery detection.
To this end, Forensics-Bench is meticulously curated to cover forgeries as various as possible, consisting of $63$K multi-choice visual questions and covering $112$ unique forgery detection types in statistics.
Specifically, the \textbf{breadth} of Forensics-Bench addresses five perspectives: 
1) \textit{Different forgery modalities}, including RGB images, NIR images, videos and texts.
2) \textit{Various semantics}, covering both human subjects and other general subjects.
3) \textit{Created/manipulated with different AI models}, such as GANs, diffusion models, VAE and etc. 
4) \textit{Various task types}, including forgery binary classification, forgery spatial localization and forgery temporal localization.
5) \textit{Versatile forgery types}, such as face swap, face attribute editing, face reenactment and etc. 
Such diversity in Forensics-Bench demands LVLMs to equip with comprehensive capabilities on discerning various forgeries, underscoring the significant challenges posed by AIGC technology nowadays.
Please see Figure \ref{fig:Forensics-Bench} for details.

% 我们评测的模型 + 我们的结论（偏科，有些做得好）
In experiments, we evaluated $22$ publicly available LVLMs and $3$ proprietary models based on Forensics-Bench to comprehensively compare their capabilities in forgery detection.
We summarized our findings as follows.
\begin{itemize}
    \item We find that Forensics-Bench presented significant challenges to state-of-the-art LVLMs, the best of which only achieved $66.7\%$ overall accuracy, highlighting the unique difficulty of robust forgery detection.
    \item Among various forgery types, LVLMs demonstrated a significant bias in their performance: they excelled at certain forgery types like spoofing and style translation (close to $100\%$), but performed poorly on others like face swap (multiple faces) and face editing (lower than $55\%$). This outcome uncovered the partial understanding LVLMs have regarding different forgery types.
    \item In different forgery detection tasks, LVLMs generally demonstrated better performance on classification tasks, while struggling with spatial and temporal localization tasks.
    \item For forgeries synthesized by popular AI models, we find that current LVLMs performed better on forgeries output by diffusion models compared with those generated by GANs. Such results exposed the limitations in LVLMs' capabilities in discerning forgeries provided by different AI models.
\end{itemize}

Overall, the contributions of this paper are summarized as follows.
i) We introduce a novel benchmark, Forensics-Bench, compiling $112$ diverse forgery detection types based on five perspectives characterizing different forgeries.
ii) We conduct a thorough and comprehensive evaluation of $25$ state-of-the-art LVLMs on Forensics-Bench. Through extensive experiments, we discover significant variability in LVLMs' performance across different forgery detection types, exposing the limitations of their capabilities.
iii) We provide multi-faceted analytical experiments related to forgery detection, such as forgery detection under perturbations and forgery attribution, further showcasing the limitations of LVLMs' understanding of forgeries.
We hope that Forensics-Bench can aid researchers in gaining a deeper understanding of the LVLMs' capabilities on forgery detection, offering insights for future designs and solutions.

% contribution

%-------------------------------------------------------------------------

\begin{figure}[t]
  \centering
   \includegraphics[width=0.9\columnwidth]{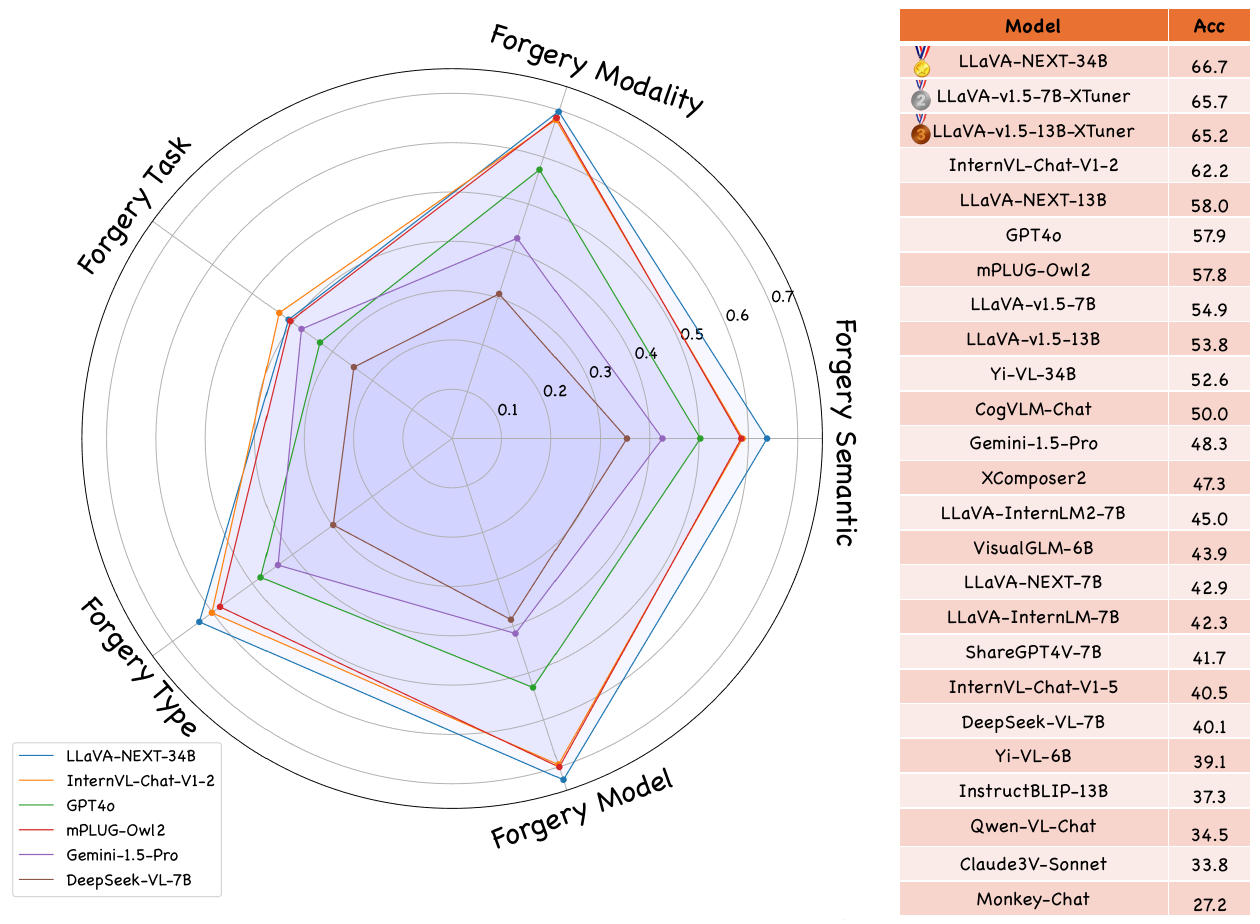}
   \caption{Forensics-Bench evaluation results of Large Vision Language Models (LVLMs). We visualize evaluation results of representative LVLMs in five Forensics-Bench perspectives on the left side and present the overall leaderboard results on the right side. For detailed quantitative results, please refer to Table \ref{tab: overall_results}.}
   \label{fig:Forensics-Benchtype-leadboard}
\end{figure}

\section{Related Work}
\label{sec:related_work}

\subsection{Forgery Detection}
With the rapid advancement of AIGC technology \citep{goodfellow2014generative,ho2020denoising,esser2021taming}, synthetic media has become increasingly realistic and indistinguishable to the public. 
Malicious users can easily exploit these techniques to spread fake news, forge judicial evidence, and tarnish the reputations of celebrities, posing significant challenges to social security.
In response, many researchers have proposed a variety of methods to detect synthetic media, aiming to ensure the authenticity and reliability of collected content \cite{ding2020swapped,tariq2018detecting,marra2018detection,dong2022explaining}. 
However, previous approaches often lacked generalization capabilities, struggling to maintain performance when faced with unseen forgeries \cite{ff++,celeb,dfdc,yandf40,bhattacharyya2024diffusion}. 
To this end, numerous methods \cite{multi,single-center-loss,improving,Dong_2023_CVPR,SBI,face-x-ray,amazon,shao2023detecting,guo2023hierarchical,yan2024transcending,yan2023ucf,feng2023self} have been proposed, expecting to increase the generalization capabilities of current forgery detectors. 
However, this challenge was significantly amplified by the recent evolution of diverse AIGC techniques \citep{goodfellow2014generative,ho2020denoising,esser2021taming,yang2024cogvideox}, which drastically increased the diversity and complexity of forgeries.
Therefore, to support the development of robust forgery detectors, designing a comprehensive evaluation benchmark for all-round forgery detection has become an urgent necessity.

\begin{table}[t]
    \centering
    \scriptsize
    \scalebox{0.75}{%
        \begin{tabular}{c|cccccc}
            \toprule
            \multirow{2}{*}{Benchmark} &\multicolumn{6}{c}{Forgery Data Collection} \\
             \cmidrule{2-7}
            & \# Sample & Semantic & \# Modality & \# Task & \# Type & \# Model\\
            \cmidrule{1-1}\cmidrule{2-7}
            FakeBench \cite{li2024fakebench} & 54K & Huamn \& General & 2 & 4 & 1 & 4 \\
            % DiffuSyn Bench \cite{zhou2024diffusyn} & 2K & \textbf{-} & 1 & 1 & 4 & 3 \\
            % \cite{liu2024fka} &  & &&&&  \\
            MMFakeBench \cite{liu2024mmfakebench} &11K &Human&2 &1&12& 12\\
            MFC-Bench \cite{wang2024mfc} & 35K & Human \& General & 2& 1 & 9& 11\\
            \cmidrule{1-1}\cmidrule{2-7}
            Forensics-Bench & 63K & Human \& General & 4
 & 4 & 21 & 22 \\
            \bottomrule
        \end{tabular}%
    }
     \caption{The comparison between Forensics-Bench and existing evaluation benchmarks for Large Vision Language Models in the context of forgery detection. }
    \label{tab:comparison-bench}
\end{table}

\subsection{LVLMs and Benchmark}
With the rapid advancement of Large Language Models (LLMs) \citep{touvron2023llama2openfoundation,yang2023baichuan2openlargescale,2023internlm,chatgpt,ouyang2022training,chowdhery2022palm}, there was a growing interest among researchers in enhancing the visual understanding capabilities of these models. 
The key to developing LVLMs lies in aligning visual content with language based on the foundation of LLMs. 
CLIP \citep{radford2021learning} was one of the pioneering works in this area, exploring contrastive learning with a large corpus of image-text pairs to align visual and language representations in a unified latent space. 
Subsequently, to improve the reception and comprehension of visual content by LLMs, Mini-GPT4 \citep{zhu2023minigpt} directly connected the visual encoder with a frozen LLM using a multilayer perceptron (MLP). 
More recently, there has been a shift towards fine-tuning existing models through instruction tuning techniques \citep{liu2023llava,liu2023improvedllava,instructblip,internlmxcomposer2,mplug,luo2023cheap,chen2023sharegpt4v}. 
For instance, LLaVA \citep{liu2023llava} fine-tuned models by constructing a $158K$ instruction-following dataset, ultimately endowing the model with excellent visual understanding capabilities.

To accurately assess the real capabilities of these LVLMs, a comprehensive and challenging benchmark \citep{liu2023mmbench,fu2023mme,schwenk2022okvqa,yue2023mmmu,sharma2018conceptual,li2023seed,yu2023mm,wu2023q} is essential. Early single-task benchmarks \citep{marino2019ok,sharma2018conceptual,goyal2017making}, such as MS-COCO \citep{sharma2018conceptual} and VQA \citep{goyal2017making}, typically evaluated LVLMs on specific aspects. As LVLMs became capable of handling an increasing variety of tasks, more comprehensive benchmarks \citep{liu2023mmbench,fu2023mme,schwenk2022okvqa,yue2023mmmu,li2023seed,yu2023mm,wu2023q,cheng2023can}, like MM-Bench \citep{liu2023mmbench}, MMMU \citep{yue2023mmmu}, and MMT-Bench \citep{mmtbench}, have been proposed to evaluate models across multiple dimensions. These benchmarks, with their broad range of tasks, comprehensively tested the true abilities of LVLMs, advancing our understanding of their capability boundaries and providing critical insights for further improvements. 
However, in the context of forgery detection, there still lacks a comprehensive evaluation benchmark to holistically assess the forgery detection capabilities of LVLMs in fine-grained perspectives.

\subsection{Forgery Detection and LVLMs}
In recent years, leveraging Large Vision-Language Models (LVLMs) for forgery detection has gained significant attention due to their exceptional capabilities in understanding versatile visual content. 
A series of studies \citep{lin2024detectingmultimediageneratedlarge,wu2023cheapfakedetectionllmusing,chang2024antifakepromptprompttunedvisionlanguagemodels,shi2024shield,jin2024fakenewsdetectionmanipulation,jia2024chatgptdetectdeepfakesstudy,qi2024sniffer,liu2024fka} have demonstrated the effectiveness of LVLMs in forgery detection tasks. 
To better evaluate the capabilities of LVLMs in this domain, several benchmarks \citep{shi2024shield,li2024fakebench,zhou2024diffusyn,liu2024mmfakebench,wang2024mfc} have been introduced to assess their effectiveness. However, these evaluation benchmarks were usually limited in scope, assessing only partial capabilities of LVLMs in the field of forgery detection. 
To address this limitation, we propose Forensics-Bench, which compiles $63$K multimodal questions across $112$ diverse forgery detection types, providing a comprehensive evaluation testbed for current state-of-the-art LVLMs.

%-------------------------------------------------------------------------

\section{Forensics-Bench}
\label{sec:FDBench}

In this section, we present the main components of Forensics-Bench. 
In Section \ref{sec3: bmk_design}, we first introduce the design principles of Forensics-Bench.
Next, in Section \ref{sec3: data_collection}, we elaborate the detailed construction process of Forensics-Bench and give a brief overview of Forensics-Bench.

\subsection{Benchmark Design} %Formulation of Task Paradigms
\label{sec3: bmk_design}
To provide a comprehensive testbed for Large Vision Language Models (LVLMs) in the context of forgery detection, we propose to design our Forensics-Bench from five perspectives to characterize different forgeries, consisting of forgery semantics, forgery modalities, forgery tasks, forgery types, and forgery models.

\noindent \textbf{Forgery semantics}. 
Human subjects have been a long-standing focus of previous forgery detection studies \cite{ff++,jiang2020deeperforensics,celeb,dfdc}, considering the significant threats of deepfakes posed to social security.
Besides, considering the unprecedented capabilities of recent generative models, such as Stable Diffusion Models \cite{rombach2022high,podell2023sdxl}, newly-generated media can contain versatile semantics given the free space of textual prompts. 
Thus, it is of great importance that future forgery detectors should show no bias towards certain types of content illustrated in forged images/videos. 
Therefore, we propose to classify data into human subjects and other general subjects in Forensics-Bench, to evaluate LVLMs' performance when encountering different semantics. 

\noindent \textbf{Forgery modalities}. In recent years, the modalities of forgeries have become increasingly various \cite{cai2023av,khalid2021fakeavceleb,ff++,celeb,shao2023detecting,wang2022forgerynir,zellers2019defending}, which may cause greater impact on our daily life.
To this end, we propose to characterize the forgery data based on different modalities, including RGB images, NIR images, videos, and texts, to assess whether LVLMs show significant performance variations across these modalities, thereby revealing their preferences. 

\noindent \textbf{Forgery tasks}. 
The general goal of forgery detection is to recognize whether the given media is real or fake, namely, conducting binary classifications \cite{ff++,ojha2023towards,wang2020cnn,wang2023dire}. 
Besides, previous studies \cite{he2021forgerynet,le2021openforensics} have focused on localizing the forged areas on images/videos, providing more finer and explainable information for users.
To this end, from this perspective, we propose to cover $4$ common forgery detection tasks, namely forgery binary classification, forgery spatial localization (segmentation masks), forgery spatial localization (bounding boxes) and forgery temporal localization.
This design encompasses most forgery detection scenarios, allowing for a detailed comparison of different LVLMs' performance across various tasks. 

\noindent \textbf{Forgery types}. With a plethora of AIGC techniques coming in handy, malicious users nowadays can apply diverse operations into different forgeries.
In this paper, we refer these kinds of operations as different forgery types.
For instance, users can conduct face swap or face reenactment to any given human subject videos \cite{ff++}, or even directly synthesizing the whole video from scratch \cite{pku_yuan_lab_and_tuzhan_ai_etc_2024_10948109,yang2024cogvideox}. 
To this end, we propose to classify the forgery data from the perspective of forgery types in Forensics-Bench.
This includes entire synthesis, face spoofing, face editing, text attribute manipulation, text swap, face reenactment, face swap (multiple faces), face swap (single face), copy-move, removal, splicing, image enhancement, out-of-context, style translation, and different combinations of the above operations.
Please see Appendix \ref{sec:details_of_FDBENCH} for detailed descriptions of these forgery types.
This comprehensive design can help to evaluate whether LVLMs maintain robust performance when faced with diverse and complex forgery types.

\noindent \textbf{Forgery models}. Another important research direction in forgery detection is to identity the forgery model that is applied to the given input \cite{he2021forgerynet,girish2021towards,yang2022deepfake}. 
This may presents continuous challenges to future forgery detectors as the generative models are constantly proposed and may be applied to generate forgeries.
To this end, in our benchmark, we propose to collect samples generated from popular AI models, which includes Diffusion models, Encoder-Decoder, graphics-based models, GANs, VAEs, RNNs and etc.
Please see Appendix \ref{sec:details_of_FDBENCH} for detailed descriptions.
This design can help to carefully analyze LVLMs' performance across data from various AI model sources.

\begin{figure}[t]
  \centering
   \includegraphics[width=0.92\columnwidth]{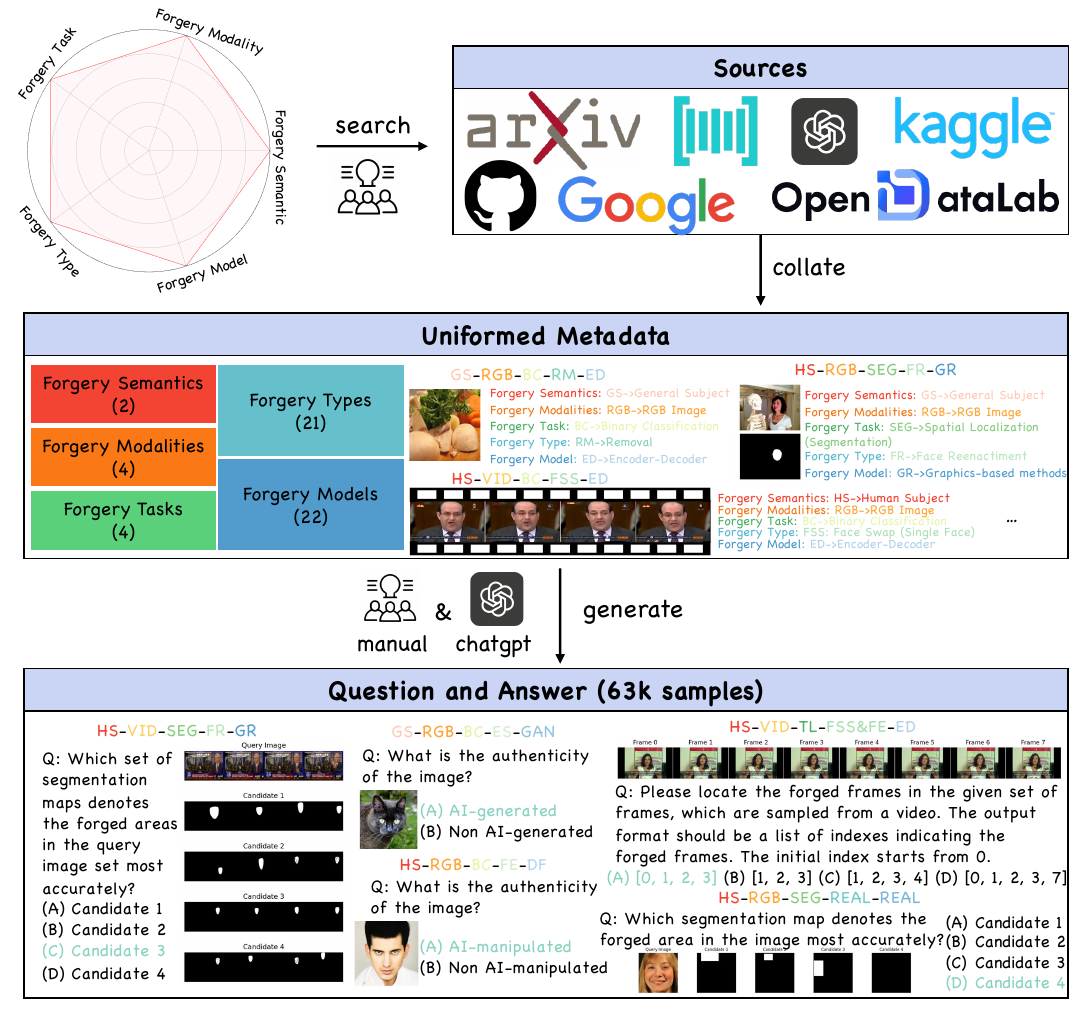}
   \caption{An illustration of the pipeline for data collection of Forensics-Bench. First, from the designed $5$ perspectives of Forensics-Bench, we searched the related public available dataset from the Internet. Then, we collated the retrieved dataset into a uniformed metadata format. Finally, we either manually transformed original data into handcrafted Questions\&Answers (Q\&A) or proceed the Q\&A transformation with the aid of ChatGPT. Forensics-Bench supports evaluations over a diverse kinds of forgeries across various perspectives. Please zoom in for better visualizations.}
   \label{fig:Forensics-Bench-pipeline}
\end{figure}

\subsection{Data Collection}
\label{sec3: data_collection}
Based on the above comprehensive benchmark design, we then collect our Forensics-Bench in a top-down hierarchy.
As shown in Fig. \ref{fig:Forensics-Bench-pipeline}, all co-authors first brainstorm and list common forgery semantics, forgery modalities, forgery tasks, forgery types and forgery models shown in pervious literature. 
Then, we retrieve the relevant public datasets for each listed item, covering forgery detections types as many as possible.
Finally, we construct corresponding multi-choice Questions\&Answers (Q\&A) samples manually or with the aid of ChatGPT.

\noindent \textbf{Dataset search}.
Considering the widespread of our benchmark design, we have gathered a comprehensive collection of datasets in the field of forgery detection, sourced from public available datasets and academic repositories. 
These sources are strategically selected to encompass a wide array of forgery data, ensuring the breadth of our dataset. 
Furthermore, we incorporate both synthetic and real-world data to approximately reflect the practical and complex challenges faced in contemporary forgery detections.
These data sources are carefully listed in Appendix \ref{sec:data structure}.
After collection, we conduct a comprehensive cleansing and filtering process to enhance the quality and utility of the dataset. 
Note that we only exploit the test/validation set of the public datasets for data cleansing, ensuring that the collected data in Forensics-Bench is not seen by LVLMs as much as possible.
Duplicates and low-quality samples are manually identified and removed to uphold high data quality standards. 
The forgery data across all sources is then standardized, followed by meticulous annotations and transformations to ensure correctness throughout the entire dataset construction.

\noindent \textbf{Metadata structure}.
We standardized all the cleansed data to generate metadata, depicting the necessary meta-information of each forgery detection type. 
Typically, we randomly select $200$ samples for each forgery detection type per public dataset for testing, recording key information within the metadata structure such as forgery types, forgery tasks, forgery models and etc. 
Considering the popularity of different forgery models, some forgery detection types may source from multiple public datasets, such as forgeries synthesized by diffusion models and GANs.
Additionally, other details like image resolution and text-image pairings are also documented. 
For samples in the format of video modality, we uniformly extract frames from each forged video to record in the metadata. 
Ultimately, this metadata is exploited to generate multi-choice Questions\&Answers (Q\&A), thus assessing the capabilities of LVLMs in the context of forgery detection.

\noindent \textbf{Question and Answer Generation}. 
Based on the recorded metadata, we then generate corresponding multi-choice Questions\&Answers based on manual rules or ChatGPT.
Some examples are illustrated in Fig. \ref{fig:Forensics-Bench-pipeline}. 
For example, for forgery binary classification task, some choices may include AI-generated/non AI-generated, or AI-manipulated/non AI-manipulated, depending on whether the media is entirely synthesized by AI models or modified by AI models based on real media.
For forgery spatial localization tasks (segmentation masks), the wrong choices are generated by adding random perturbations to the ground truth.
These choice designs can help reduce the ambiguity of the generated Questions\&Answers and increase the relevance between correct answers and wrong answers, ensuring the fairness on the evaluation results of Forensics-Bench.

\noindent \textbf{Dataset Statistics}.
Finally, Forensics-Bench consists of $63292$ data samples, covering $46358$ forgery samples and $16934$ real samples, following the designed ratios of pervious datasets \cite{ff++,celeb,dfdc} where the forgery samples accounts for the majority given the diversities of AIGC techniques.
In Forensics-Bench, we cover $2$ forgery semantics, $4$ forgery modalities, $4$ forgery tasks, $21$ forgery types and $22$ forgery models, comprehensively evaluating the perception, location and reasoning capabilities of LVLMs in the context of forgery detection.
A detailed comparison with previous evaluation benchmarks for Large Vision Language Models in the context of forgery detection is provided in Table \ref{tab:comparison-bench}.
To the best of our knowledge, Forensics-Bench is the largest forgery detection benchmark
for LVLMs to date, featuring the most diverse forgeries and evaluation perspectives.

\subsection{Other Evaluation Protocols}
Thanks to the comprehensiveness of Forensics-Bench, we can further complement the evaluations of LVLMs' forgery detection capabilities with our benchmark.
To this end, we further propose $2$ extra evaluation protocols supported by Forensics-Bench, providing finer analyses on LVLMs' abilities.

\noindent \textbf{Protocol 1. Robust Forgery Detection}. 
Inspired by previous studies \cite{jiang2020deeperforensics,haliassos2021lips}, we introduce common perturbations into the samples to evaluate the stability of LVLMs' forgery detection capabilities in noisy real-life environments. 
These introduced perturbations are commonly seen on the Internet, consisting of change of color saturation, local block-wise distortion, change of color contrast, Gaussian blur, white Gaussian noise and JPEG compression, each of which features $5$ different intensity levels.
This evaluation protocol can help assess LVLMs' forgery detection capabilities in real-life scenarios, exploring the potential of LVLMs for practical deployment. 

\noindent \textbf{Protocol 2. Forgery Attribution}. 
Inspired by pervious studies \cite{girish2021towards,yang2022deepfake}, we can evaluate the forgery attribution capabilities of LVLMs owning to the forgery data in Forensics-Bench, which is generated/manipulated by a wide range of AI models.
Specifically, we repurpose our questions in Forensics-Bench to ask LVLMs to identify the AI models that are applied to the given media.
The wrong choices are randomly sampled from the $22$ forgery models listed in Forensics-Bench.
Note that we only repurpose the samples featuring forgery binary classifications in Forensics-Bench for this protocol.
This evaluation perspective can help further analyze LVLMs' forgery detection capabilities, providing explainable and detailed support for LVLMs' answers.

\section{Experiments}
\begin{table}[t]
\begin{center}
{
\linespread{0.6}
\setlength\tabcolsep{3pt}
\scriptsize
{%
\begin{tabular}{l|c|ccccc}
\toprule
Model&Overall&Semantic&Modality&Task&Type&Model \\ \midrule
\rowcolor{lightgray} 
\multicolumn{7}{c}{\textbf{\textit{Proprietary Large Vision Language Models}}} \\ \midrule
GPT4o&57.9&50.2
&57.3
&33.1
&47.9
&53.1\\
 \midrule
 Gemini-1.5-Pro&48.3&42.6
&42.7
&37.8
&43.7
&41.6\\
 \midrule
 Claude3V-Sonnet&33.8&28.4
&28.5
&32.1
&29.9
&28.9\\
 \midrule
 \rowcolor{lightgray} 
\multicolumn{7}{c}{\textbf{\textit{Open-sourced Large Vision Language Models}}} \\ \midrule
LLaVA-NEXT-34B&66.7&63.8
&69.7
&41.0
&63.3
&72.7
\\ \midrule
LLaVA-v1.5-7B-XTuner &65.7&61.2
&68.3
&41.9
&58.2
&66.8\\ \midrule
LLaVA-v1.5-13B-XTuner&65.2&62.9
&68.7
&37.9
&61.3
&71.8\\  \midrule
InternVL-Chat-V1-2&62.2&58.8
&67.9
&43.4
&60.1
&69.5\\
 \midrule
LLaVA-NEXT-13B&58.0&64.0
&66.7
&32.0
&62.3
&71.2\\
 \midrule
mPLUG-Owl2&57.8&58.5
&68.4
&40.5
&58.1
&70.0\\
 \midrule
LLaVA-v1.5-7B&54.9&61.6
&68.7
&37.1
&64.0
&70.8\\
 \midrule
LLaVA-v1.5-13B&53.8&52.7
&64.2
&34.1
&55.6
&63.7\\
 \midrule
Yi-VL-34B&52.6&47.2
&53.6
&39.7
&41.2
&51.6\\
 \midrule
CogVLM-Chat&50.0&44.1
&49.5
&32.2
&45.4
&52.0\\
 \midrule
XComposer2&47.3&42.2
&43.8
&28.3
&42.9
&48.4\\
 \midrule
LLaVA-InternLM2-7B&45.0&40.8
&52.2
&30.5
&42.6
&50.3\\
 \midrule
VisualGLM-6B&43.9&38.9
&39.1
&30.3
&35.1
&39.2\\
 \midrule
LLaVA-NEXT-7B&42.9&49.0
&53.1
&32.1
&55.7
&63.7\\
 \midrule
LLaVA-InternLM-7B&42.3&37.7
&39.4
&30.2
&39.9
&47.5\\
 \midrule
ShareGPT4V-7B&41.7&44.6
&46.9
&32.3
&51.8
&58.5\\
 \midrule
InternVL-Chat-V1-5&40.5&39.9
&33.6
&28.7
&41.7
&47.6\\
 \midrule
DeepSeek-VL-7B&40.1&35.4
&30.8
&24.6
&29.8
&38.6\\
 \midrule
Yi-VL-6B&39.1&38.2
&39.4
&30.7
&39.4
&48.1\\
 \midrule
InstructBLIP-13B&37.3&33.1
&42.2
&27.1
&28.4
&33.7\\
 \midrule
Qwen-VL-Chat&34.5&29.6
&32.1
&27.1
&32.4
&34.3\\
 \midrule
Monkey-Chat&27.2&18.6
&18.1
&20.6
&19.2
&21.2\\  \bottomrule
\end{tabular}%
}
}
\end{center}
\caption{Quantitative results for $22$ open-sourced LVLMs and $3$ proprietary LVLMs across $5$ perspectives of forgery detection are summarized. Accuracy is the metric. The overall score is calculated across all data in Forensics-Bench.}
\label{tab: overall_results}
\end{table}

\subsection{Experiment Setup}
\textbf{LVLM Models}. With our proposed Forensics-Bench, we then conduct experiments to evaluate $22$ open-sourced LVLMs, including LLaVA-NEXT-34B \cite{liu2024llavanext}, LLaVA-v1.5-7B-XTuner \cite{2023xtuner}, LLaVA-v1.5-13B-XTuner \cite{2023xtuner}, InternVL-Chat-V1-2 \cite{internvl_chat,2023internlm}, LLaVA-NEXT-13B \cite{liu2024llavanext}, mPLUG-Owl2 \cite{mplug}, LLaVA-v1.5-7B \cite{liu2023llava,liu2023improvedllava}, LLaVA-v1.5-13B \cite{liu2023llava,liu2023improvedllava}, Yi-VL-34B \cite{young2024yi}, CogVLM-Chat \cite{wang2023cogvlm}, XComposer2 \cite{internlmxcomposer2}, LLaVA-InternLM2-7B \cite{2023xtuner}, VisualGLM-6B, LLaVA-NEXT-7B \cite{liu2024llavanext}, LLaVA-InternLM-7B \cite{2023xtuner}, ShareGPT4V-7B \cite{chen2023sharegpt4v}, InternVL-Chat-V1-5 \cite{internvl_chat,2023internlm}, DeepSeek-VL-7B \cite{lu2024deepseek}, Yi-VL-6B \cite{young2024yi}, InstructBLIP-13B \cite{instructblip}, Qwen-VL-Chat \cite{Qwen-VL} and Monkey-Chat \cite{li2023monkey}.
Besides, we also conduct evaluations on $3$ proprietary models: GPT4o \cite{gpt4o}, Gemini 1.5 Pro \cite{team2024gemini} and Claude 3.5 Sonnet \cite{Claude2023}.

\noindent \textbf{Evaluation Details}.
With the evaluation tool \cite{duan2024vlmevalkit} provided in OpenCompass \cite{2023opencompass}, we followed previous studies \cite{mmtbench,meng2024mmiu} to conduct evaluations: 1) we first manually check whether the option letter appears in the LVLMs' answers; 2) we then manually check whether the option content appears in the LVLMs' answers; 3) we finally resort ChatGPT to help extract the matching option.
If the above extractions still fail, we set the model's answer as \textit{Z} \cite{yue2023mmmu}. 
As for evaluation metrics, we use accuracy in our experiments.

\subsection{Main Results}
\begin{figure}[t]
  \centering
   \includegraphics[width=0.92\columnwidth]{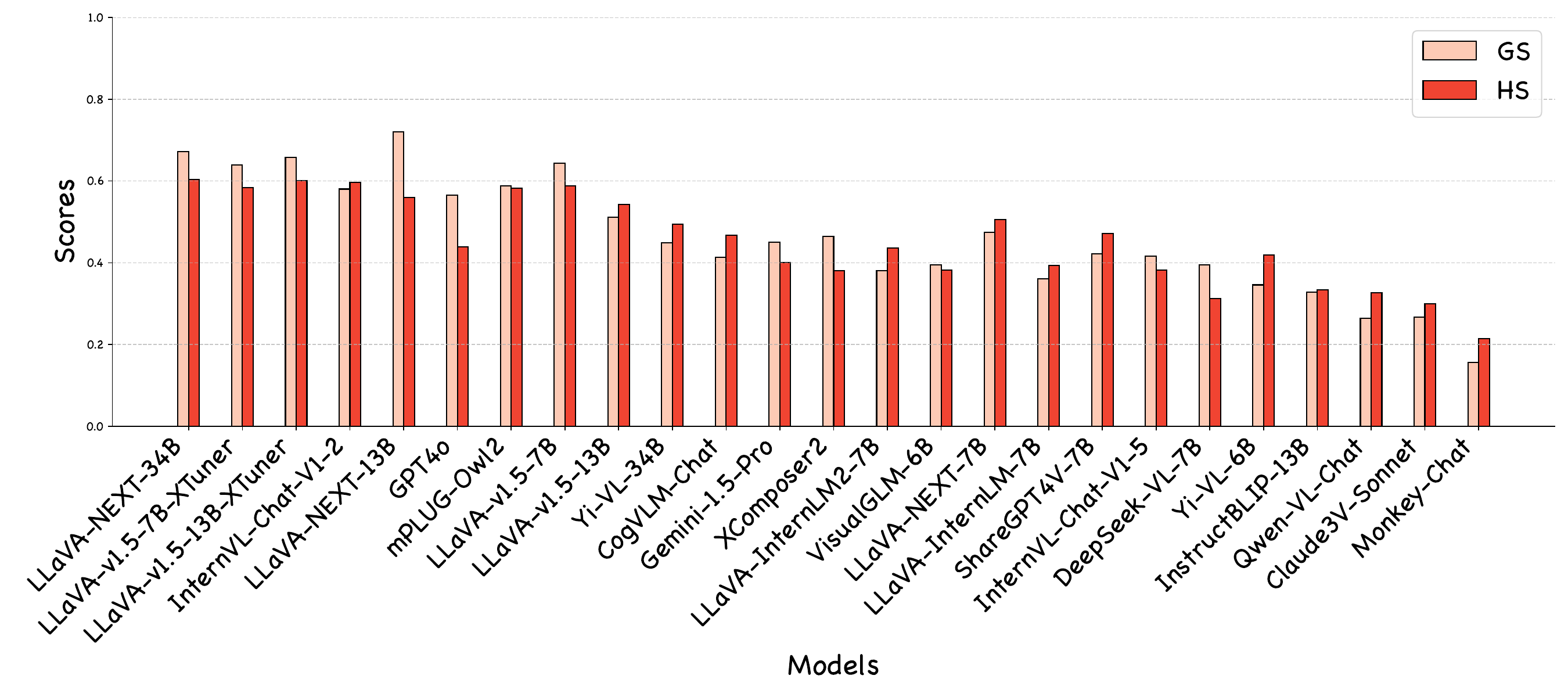}
   \caption{Results of Forensics-Bench from the perspective of \textit{forgery semantics.}. Most LVLMs did not demonstrate strong bias towards certain media content in terms of human subject \textit{vs} general subject.}
   \label{fig:Forensics-Bench-semantics}
\end{figure}

The main evaluation results are summarized in Table \ref{tab: overall_results}.
The score in each perspective of Forensics-Bench design is the averaged accuracy over related samples. 
To this end, we have the following findings:
1) We find that Forensics-Bench presented significant challenges to state-of-the-art LVLMs. The best one (LLaVA-NEXT-34B) only achieved $66.7\%$ overall accuracy on Forensics-Bench, underscoring the unique difficulty of generalized forgery detection.
2) We find that proprietary LVLMs (such as GPT-4o) demonstrated relatively weaker performance than open-sourced models, especially the series of LLaVA models, in the context of forgery detection. This is mostly because that proprietary LVLMs tend to response with more conservative answers, admitting that they can not conclude the authenticity of the input with strong confidence.
3) We further conduct detailed analyses from each perspective in our Forensics-Bench:

\noindent \textbf{Analysis on forgery semantics}. We illustrated the detailed performance of $25$ LVLMs in Figure \ref{fig:Forensics-Bench-semantics} from the perspective of forgery semantics. 
It can be seen that most LVLMs did not demonstrate significant bias towards certain content in terms of human subjects \textit{vs} general subjects.
This provide great starting points for the development of future all-round forgery detectors under the paradigm of LVLMs.

\begin{figure}[t]
  \centering
   \includegraphics[width=0.92\columnwidth]{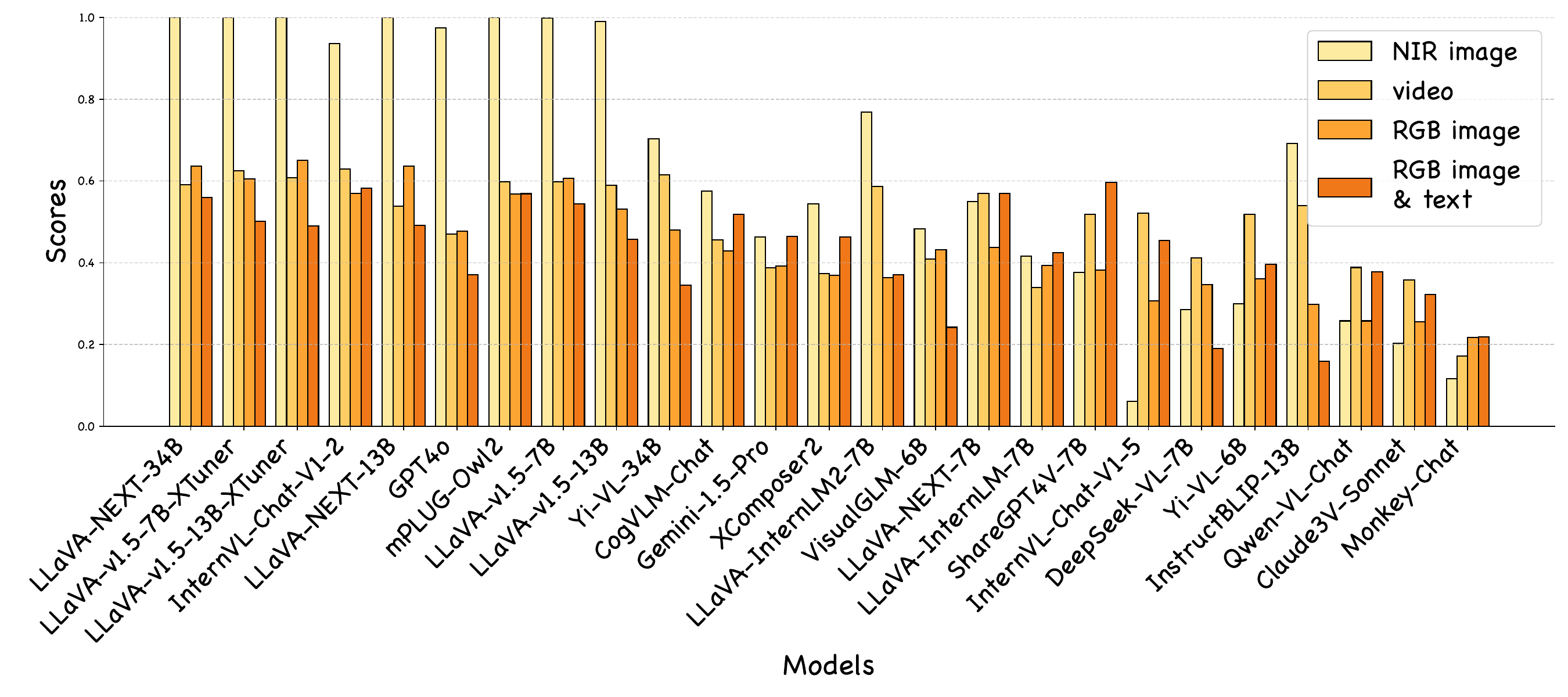}
   \caption{Results of Forensics-Bench from the perspective of \textit{forgery modality.}. Current LVLMs failed to perform well across all forgery modalities collected in Forensics-Bench.}
   \label{fig:Forensics-Bench-modality}
\end{figure}

\noindent \textbf{Analysis on forgery modalities}. 
We showed the detailed performance of $25$ LVLMs in Figure \ref{fig:Forensics-Bench-modality} from the perspective of forgery modalities. 
We find that top-performing LVLMs (such as LLaVA-NEXT-34B) achieved impressive binary classification performance on forgeries in the near-infrared (NIR) modality. 
Meanwhile, when the input content contains both RGB images and texts \cite{shao2023detecting}, these LVLMs struggled to performed well.
It is still worth exploring to design robust forgery detectors excelled at different modalities.

\begin{figure}[t]
  \centering
   \includegraphics[width=0.92\columnwidth]{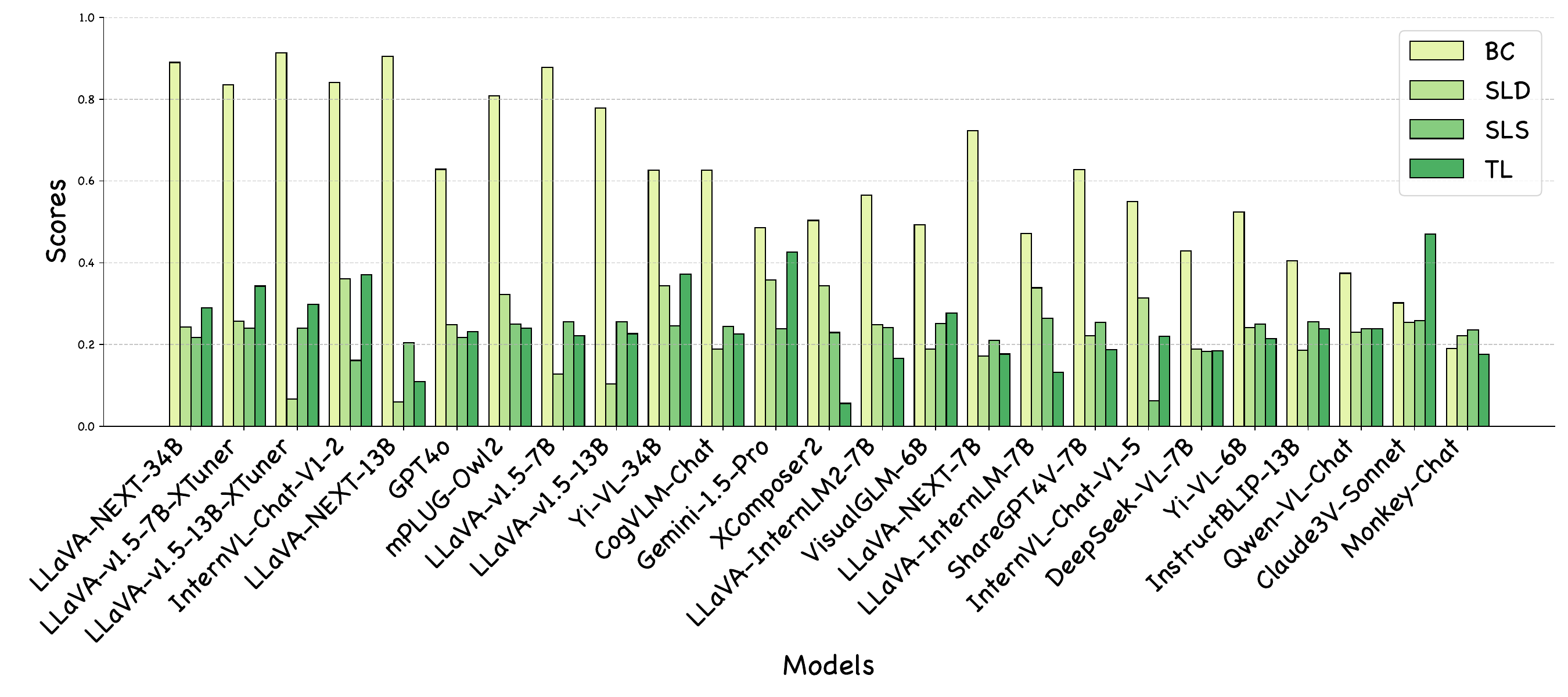}
   \caption{Results of Forensics-Bench from the perspective of \textit{forgery task.} Current LVLMs performed well in forgery binary classification task, but struggled to maintain the performance over other forgery tasks designed in Forensics-Bench.}
   \label{fig:Forensics-Bench-task}
\end{figure}

\noindent \textbf{Analysis on forgery tasks}.
The detailed performance of $25$ LVLMs from the perspective of forgery tasks is shown in Figure \ref{fig:Forensics-Bench-task}.
We find that most LVLMs demonstrated relatively great performance in the forgery binary classification (BC) task, while having difficulty maintaining strong performance in the tasks of forgery spatial localization (segmentation masks/bounding boxes) (SLS/SLD) and forgery temporal localization (TL).
Such results reveal that most LVLMs still required improvements over location and reasoning capabilities in different forgery detection tasks.

\begin{figure}[t]
  \centering
   \includegraphics[width=0.9\columnwidth]{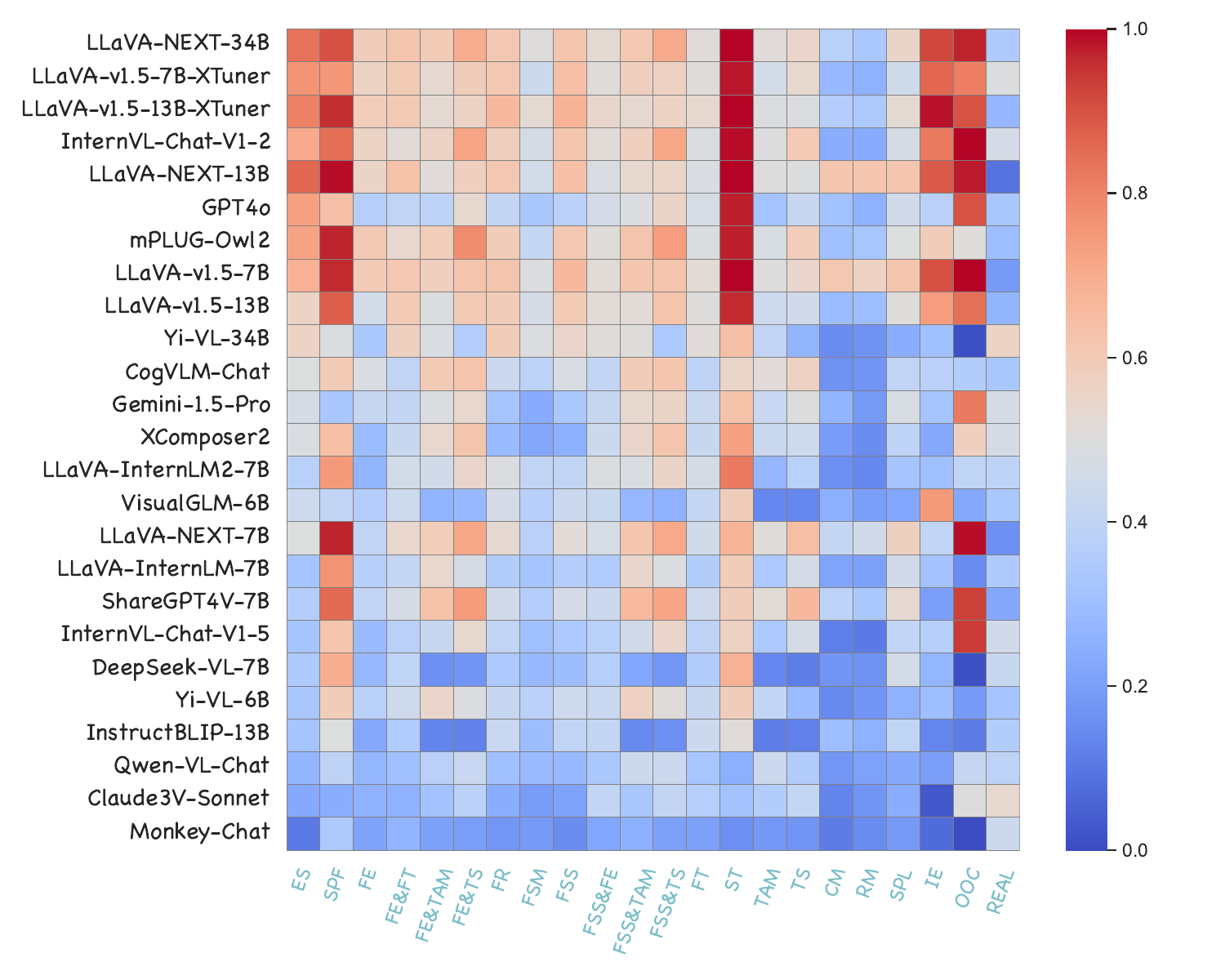}
   \caption{Results of Forensics-Bench from the perspective of \textit{forgery type}. Current LVLMs still exhibited limitations in forgery detections across a wide range of forgery types.}
   \label{fig:Forensics-Bench-type}
\end{figure}

\noindent \textbf{Analysis on forgery types}.
The detailed performance of $25$ LVLMs from the perspective of forgery types is illustrated in Figure \ref{fig:Forensics-Bench-type}. 
Firstly, we find that current LVLMs still found it challenging to perform well over a wide range of forgery types, such as face swap (multiple faces), copy-move (CM), removal (RM) and splicing (SPL).
Second, we find that leading LVLMs like LLaVA series models already excelled in certain forgery types, such as face spoofing (SPF), image enhancement (IE), style translation (ST) and out-of-context (OOC), indicating their potential to grow into more generalized forgery detectors.

\begin{figure}[t]
  \centering
   \includegraphics[width=0.9\columnwidth]{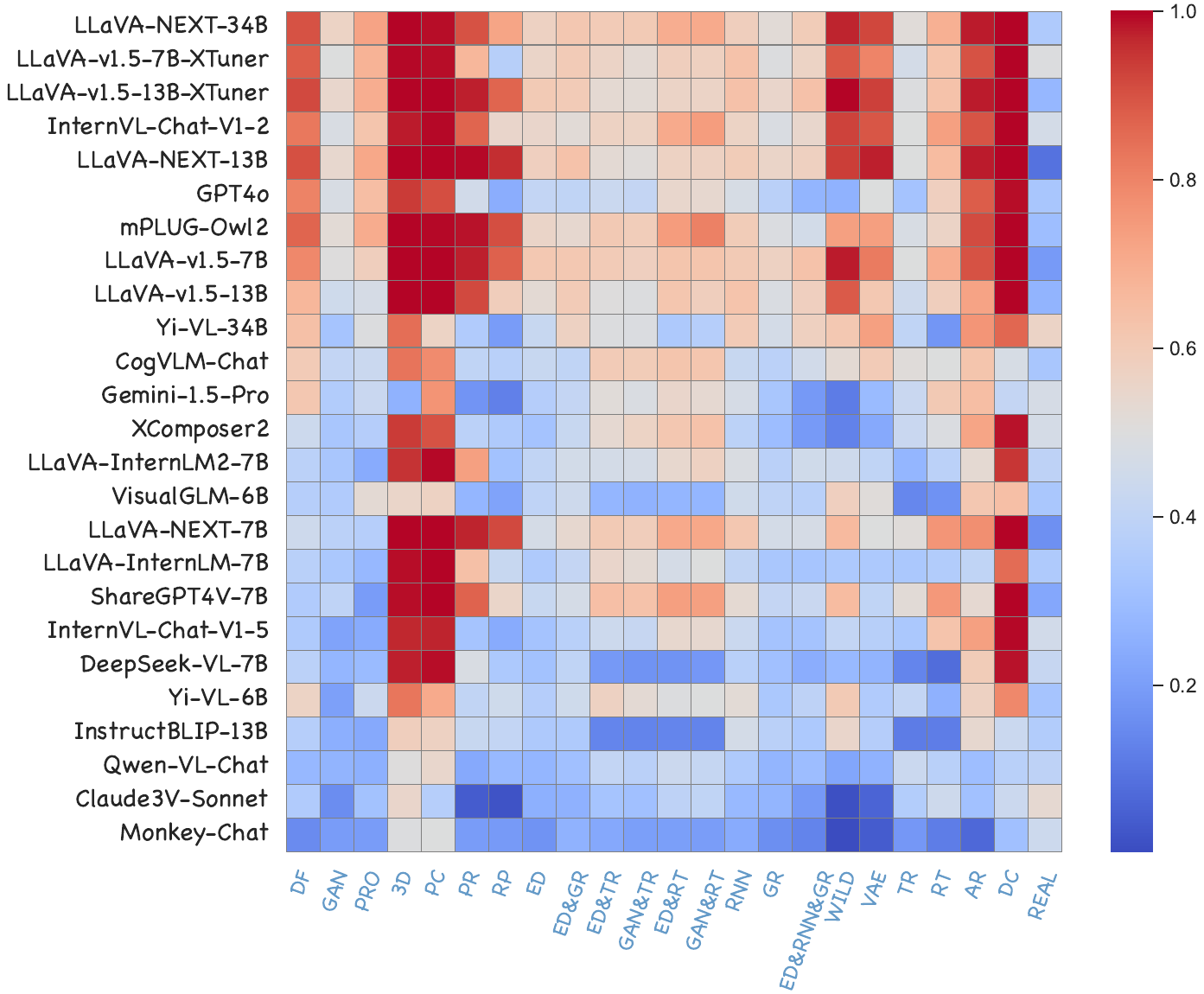}
   \caption{Results of Forensics-Bench from the perspective of \textit{forgery model}. Current LVLMs still fell short in forgery detections across forgeries output by a wide range of forgery models.}
   \label{fig:Forensics-Bench-model}
\end{figure}

\noindent \textbf{Analysis on forgery models}.
The detailed performance of $25$ LVLMs from the perspective of forgery models is illustrated in Figure \ref{fig:Forensics-Bench-model}. 
It is noticeable that leading LVLMs achieved excellent performance at forgeries created with spoofing methods, such as 3D masks (3D) and paper cut (PC).
Besides, for forgeries synthesized by popular AI models, we find that current LVLMs performed better on forgeries output by diffusion models (DF) compared with those output by GANs, which may expose the limited discerning capabilities of LVLMs for forgeries output by different AI models.
Moreover, we find that current LVLMs experienced challenges when recognizing forgeries generated by the combinations of multiple AI models, such as Encoder-Decoder\&Graphics-based methods (ED\&GR), Generative Adversarial Networks\&Transformer (GAN\&TR) and etc.
Such forgeries may pose more significant challenges to LVLMs' forgery detection capabilities in the future.

\subsection{Other Evaluation Protocol Analyses}
Beyond the direct evaluations of Forensics-Bench, we further conduct \textit{Robust Forgery Detection} and \textit{Forgery Attribution} to complement the assessment of LVLMs' forgery detection capabilities, providing more detailed analyses based on the comprehensive design of Forensics-Bench.

\noindent \textbf{Protocol 1. Robust Forgery Detection.} 
In this experiment, we mainly evaluated the performance of open-sourced top-performing LVLMs, including LLaVA-NEXT-34B, LLaVA-v1.5-7B-XTuner, InternVL-Chat-V1-2 and mPLUG-Owl2.
Following previous studies \cite{jiang2020deeperforensics,haliassos2021lips,Dong_2023_CVPR}, we reported the average overall score among $5$ different intensity levels for each kind of perturbation.
The results are summarized in Table \ref{tab: robust_forgery_results}.
Intriguingly, we find that perturbations like change of color contrast and change of color saturation tended to compromised LVLMs' abilities to perform forgery detections.
Meanwhile, other perturbations, such as local block-wise distortion, had a less negative impact, which may even enhance LVLMs' capabilities to distinguish forgeries.
We argue that this is mostly because perturbations like local block-wise distortion would significantly help reduce the content bias of forgeries, which is also experimentally verified to be effective for forgery detections in previous studies \cite{Dong_2023_CVPR,yan2023ucf,liang2022exploring,yan2024transcending}.

\begin{table}[t]
\begin{center}
{
\linespread{0.2}
\setlength\tabcolsep{4pt}
\scriptsize
{%
\begin{tabular}{l|cccccc}
\toprule
Model&Saturation&Contrast&Block&Noise&Blur& Pixel\\ \midrule
\midrule
LLaVA-NEXT-34B&62.3&61.2&68.8&67.0&62.0&63.2\\ \midrule
LLaVA-v1.5-7B-XTuner &59.0&58.2&65.2&62.6& 59.1&58.7 \\ \midrule
% LLaVA-v1.5-13B-XTuner& 63.3&63.8&68.4&67.7&64.2&64.2\\  \midrule
InternVL-Chat-V1-2& 54.0&55.4&67.8&60.1&61.1&57.3\\ \midrule
% LLaVA-NEXT-13B&62.6&63.6&68.3&67.1&65.1&65.4\\ \midrule
mPLUG-Owl2&55.5&55.8&64.4&60.2&62.3&58.0\\
\bottomrule
\end{tabular}%
}
}
\end{center}
\caption{Robust forgery detection results for LVLMs across $6$ perturbation types are summarized.  The overall score is calculated across all data in Forensics-Bench.}
\label{tab: robust_forgery_results}
\end{table}

\noindent \textbf{Protocol 2. Forgery Attribution.} 
In this experiment, we repurposed the forgery binary classification data in our benchmark into forgery attribution samples.
For example, we asked LVLMs ``\textit{What is the forgery model that is applied to this RGB image?}", along with $4$ options sampled from our forgery models set. 
Similarly, we evaluated the performance of LLaVA-NEXT-34B, LLaVA-v1.5-7B-XTuner, InternVL-Chat-V1-2 and mPLUG-Owl2.
The reported accuracy results are summarized in Table \ref{tab: attribution_results}.
We find that although these models demonstrated relatively great performance on forgery binary classification (c.f. Fig. \ref{fig:Forensics-Bench-task}), they encountered obstacles in performing fine-grained classification of different forgery models, highlighting the challenges of complex forgery detection scenarios.

\begin{table}[t]
\begin{center}
{
\linespread{0.2}
\setlength\tabcolsep{3pt}
\scriptsize
{%
\begin{tabular}{l|c|ccccc}
\toprule
Model&Overall&Semantic&Modality&Task&Type&Model \\ \midrule
LLaVA-NEXT-34B&44.0&25.3
&32.0
&26.7
&28.8
&31.2
\\ \midrule
LLaVA-v1.5-7B-XTuner &42.2&31.4
&42.3
&32.6
&32.2
&33.3
\\ \midrule
% LLaVA-v1.5-13B-XTuner& \\  \midrule
InternVL-Chat-V1-2&41.6&24.4
&30.9
&26.3
&29.8
&31.1\\ \midrule
% LLaVA-NEXT-13B&\\ \midrule
mPLUG-Owl2&39.9&34.7
&46.7
&36.3
&38.1
&33.5\\
 \bottomrule
\end{tabular}%
}
}
\end{center}
\caption{Forgery attribution results for LVLMs across $5$ perspectives of forgery detection are summarized. Accuracy is the metric.}
\label{tab: attribution_results}
\end{table}
% 0.439979788
% 0.422426293
% 0.416472604
% 0.399424404
\section{Conclusion}
In this paper, we have presented Forensics-Bench, a comprehensive benchmark designed to comprehensively assess LVLMs’ discerning capabilities on forgery media, evaluating recognition, location and reasoning capabilities on diverse forgeries.
Through this comprehensive task design, our benchmark has provided thorough evaluations of the current popular LVLMs in the domain of forgery detection. 
Our experiments have effectively uncovered their weaknesses and biases in this field, offering valuable insights for future improvements in their performance on all-round forgery detection goals. 
We hope that our benchmark can serve as a platform for exploring the use of LVLMs in forgery detection tasks, potentially expanding the overall capability maps of LVLMs towards the next level of AGI. 

\section*{Acknowledgements}
This paper is partially supported by the National Key R\&D Program of China No.2022ZD0161000 and the General Research Fund of Hong Kong No.17200622 and 17209324. 

{
    \small
    \bibliographystyle{ieeenat_fullname}
    \bibliography{main}
}

\clearpage
\setcounter{page}{1}
\onecolumn
% \maketitlesupplementary
{
   \newpage
       % \twocolumn[
        \centering
        \Large
        \textbf{\thetitle}\\
        \vspace{0.5em}Supplementary Material \\
        \vspace{1.0em}
       % ] %< twocolumn
   }
% \onecolumn
\section{Abbreviations for Forensics-Bench}
\label{sec:abbreviation}
The detailed abbreviations utilized throughout the paper are listed in Table \ref{tab:abbreviation}.

\begin{table*}[h]
\centering
\resizebox{0.9\textwidth}{!}{%
\begin{tabular}{l|l|l|l}
\toprule
Abbreviation & Full Term & Abbreviation & Full Term \\
\midrule
\multicolumn{4}{c}{\textbf{Forgery Semantics}} \\
\midrule
HS &Human Subject & GS & General Subject \\
\midrule
\multicolumn{4}{c}{\textbf{Forgery Modalities}} \\
\midrule
RGB & RGB Images & NIR & Near-infrared Images\\
VID & Videos & RGB\&TXT & RGB Images and Texts\\
\midrule
\multicolumn{4}{c}{\textbf{Forgery Tasks}} \\
\midrule
BC & Forgery Binary Classification & SLD & Forgery Spatial Localization (Detection) \\
SLS & Forgery Spatial Localization (Segmentation) & TL & Forgery Temporal Localization \\ \midrule
\multicolumn{4}{c}{\textbf{Forgery Types}} \\
\midrule
ES&Entire Synthesis&SPF&Spoofing\\ 
FE&Face Editing &FE\&FT&Face Editing \& Face Transfer\\ 
FE\&TAM&Face Editing \& Text Attribute Manipulation&FE\&TS&Face Editing \& Text Swap\\
FR&Face Reenactment&FSM&Face Swap (Multiple Faces)\\
FSS& Face Swap (Single Face)&FSS\&FE&Face Swap (Single Face) \& Face Editing\\
FSS\&TAM&Face Swap (Single Face) \& Text Attribute Manipulation&FSS\&TS&Face Swap (Single Face) \& Text Swap\\
FT&Face Transfer&CM&Copy-Move\\
RM&Removal&SPL&Splicing\\
IE&Image Enhancement&REAL& Real media without being forged\\
OOC&Out-of-Context&ST&Style Translation\\
TAM&Text Attribute Manipulation&TS&Text Swap\\ \midrule
\multicolumn{4}{c}{\textbf{Forgery Models}} \\ \midrule
3D&3D masks&RNN&Recurrent Neural Networks\\
TR&Transformer&DC&Decoder\\
DF&Diffusion models&ED&Encoder-Decoder \\
ED\&RNN\&GR&Encoder-Decoder\&Recurrent Neural Networks\&Graphics-based methods&ED\&TR&Encoder-Decoder\&Transformer\\
ED\&RT&Encoder-Decoder\&Retrieval-based methods&ED\&GR&Encoder-Decoder\&Graphics-based methods\\
GAN&Generative Adversarial Networks&GAN\&TR&Generative Adversarial Networks\&Transformer\\
GAN\&RT&Generative Adversarial Networks\&Retrieval-based methods&PC&Paper-Cut\\
Real&Real media without being forged&PR&Print\\
PRO&Proprietary&RP&Replay\\
RT&Retrieval-based methods&AR&Auto-regressive models\\
GR&Graphics-based methods&WILD&Unknown (in the wild)\\
VAE&Variational Auto-Encoders&&\\
\bottomrule
\end{tabular}}
\caption{The abbreviations of terms mentioned in Forensics-Bench and their corresponding full terms.}
\label{tab:abbreviation}
\end{table*}

\section{Data Structure of Forensics-Bench}
\label{sec:data structure}
In Table \ref{tab:Forensics-Bench data part 1}, Table \ref{tab:Forensics-Bench data part 2} and Table \ref{tab:Forensics-Bench data part 3}, we present all 112 unique forgery detection types from Forensics-Bench, covering 5  designed perspectives characterizing forgeries. 
These tables include details on sample number, the specific information of 5 designed perspectives in Forensics-Bench and data sources collected under licenses.

\begin{table*}[t]
\resizebox{1.0\textwidth}{!}{%
\begin{tabular}{l|l|l|l|l|l|r}
\toprule
\textbf{Forgery Task} &
  \textbf{Forgery Semantic} &
  \textbf{Forgery Type} &
  \textbf{Forgery Model} &
  \textbf{Forgery Modality} &
  \textbf{Data Sources} &
  \multicolumn{1}{l}{\textbf{Sample Number}} \\ \hline
Forgery Binary Classification &
  Human Subject &
  Entire Synthesis &
  Generative Adversarial Networks &
  RGB Images &
  \begin{tabular}[c]{@{}l@{}}HiFi-IFDL(StyleGANv2-ada on FFHQ) \cite{guo2023hierarchical};\\ HiFi-IFDL(StyleGANv3 on FFHQ) \cite{guo2023hierarchical};\\ DFFD(ProGAN) \cite{dang2020detection};\\
  DFFD(StyleGANv1) \cite{dang2020detection};\\
  ForgeryNet(StyleGANv2) \cite{he2021forgerynet};\\
  ForgeryNet(DiscoFaceGAN) \cite{he2021forgerynet};\\
  Fake2M(StyleGANv3 on FFHQ/metface) \cite{lu2024seeing}\end{tabular} &
  2000 \\ \hline
Forgery Binary Classification &
  Human Subject &
  Entire Synthesis &
  Generative Adversarial Networks &
  Near-infrared Images & \begin{tabular}[c]{@{}l@{}}ForgeryNIR(ProGAN) \cite{wang2022forgerynir};\\
  ForgeryNIR(StyleGAN) \cite{wang2022forgerynir};\\
  ForgeryNIR(StyleGAN2) \cite{wang2022forgerynir}\end{tabular} &
  1200 \\ \hline
Forgery Binary Classification &
  General Subject &
  Entire Synthesis &
  Generative Adversarial Networks &
  RGB Images &
  \begin{tabular}[c]{@{}l@{}}HiFi-IFDL(StyleGANv2-ada on AFHQ) \cite{guo2023hierarchical};\\ HiFi-IFDL(styleGANv3 on AFHQ) \cite{guo2023hierarchical};\\ GenImage(BigGAN on ImageNet classes) \cite{zhu2024genimage};\\CNN-spot(ProGAN on LSUN) \cite{wang2020cnn};\\CNN-spot(StyleGANv1/v2 on LSUN) \cite{wang2020cnn};\\CNN-spot(BigGAN on ImageNet) \cite{wang2020cnn};\\Fake2M(StyleGAN3 on AFHQ) \cite{lu2024seeing} \end{tabular} &
  6000 \\ \hline
Forgery Binary Classification &
  Human Subject &
  Entire Synthesis &
  Proprietary &
  RGB Images &
  {Diff(midjourney) \cite{cheng2024diffusion}} &
  200 \\ \hline
Forgery Binary Classification &
  Human Subject &
  Entire Synthesis &
  Diffusion models &
  RGB Images &
\begin{tabular}[c]{@{}l@{}}Diff(SDXL) \cite{cheng2024diffusion};\\
Diff(FreeDoM\_T) \cite{cheng2024diffusion};\\
Diff(HPS) \cite{cheng2024diffusion};\\
Diff(LoRA) \cite{cheng2024diffusion};\\
Diff(DreamBooth) \cite{cheng2024diffusion};\\
Diff(SDXL Refiner) \cite{cheng2024diffusion};\\
Diff(FreeDoM\_I) \cite{cheng2024diffusion}\end{tabular} &
  1400 \\ \hline
Forgery Binary Classification &
  General Subject &
  Entire Synthesis &
  Diffusion models &
  Videos &
  Open-Sora-Plan \cite{pku_yuan_lab_and_tuzhan_ai_etc_2024_10948109} &
  100 \\ \hline
Forgery Binary Classification &
  General Subject &
  Entire Synthesis &
  Auto-regressive models &
  Videos &
  Cogvideo \cite{hong2022cogvideo} &
  100 \\ \hline
Forgery Binary Classification &
  General Subject &
  Entire Synthesis &
  Diffusion models &
  RGB Images &
  {\begin{tabular}[c]{@{}l@{}}HiFi-IFDL(GDM on LSUN) \cite{guo2023hierarchical};\\
  HiFi-IFDL(LDM on LSUN) \cite{guo2023hierarchical};\\
  HiFi-IFDL(DDPM on LSUN) \cite{guo2023hierarchical};\\
  HiFi-IFDL(DDIM on LSUN) \cite{guo2023hierarchical};\\
  GenImage(SD V1.4 on ImageNet classes) \cite{zhu2024genimage};\\
  GenImage(SD V1.5 on ImageNet classes) \cite{zhu2024genimage};\\
  GenImage(ADM on ImageNet classes) \cite{zhu2024genimage};\\
  GenImage(GLIDE on ImageNet classes) \cite{zhu2024genimage};\\
  Fake2M(SD V2.1) \cite{lu2024seeing};\\
  Fake2M(SD V1.5) \cite{lu2024seeing};\\
  Fake2M(IF V1.0) \cite{lu2024seeing};\\
  DiffusionForensics(ADM on LSUN) \cite{wang2023dire};\\
  DiffusionForensics(DDPM on LSUN) \cite{wang2023dire};\\
  DiffusionForensics(iDDPM on LSUN) \cite{wang2023dire};\\
  DiffusionForensics(PNDM on LSUN) \cite{wang2023dire};\\
  DiffusionForensics(LDM on LSUN) \cite{wang2023dire};\\
  DiffusionForensics(SD-v1 on LSUN) \cite{wang2023dire};\\ 
  DiffusionForensics(SD-v2 on LSUN) \cite{wang2023dire};\\
  DiffusionForensics(ADM on ImageNet) \cite{wang2023dire};\\
  DiffusionForensics(SD-v1 on ImageNet) \cite{wang2023dire}\end{tabular}} &
  5800 \\ \hline
Forgery Binary Classification &
  General Subject &
  Entire Synthesis &
  Proprietary &
  RGB Images &
  \begin{tabular}[c]{@{}l@{}}GenImage(Midjourney on ImageNet classes) \cite{zhu2024genimage};\\
  GenImage(Wukong on ImageNet classes) \cite{zhu2024genimage};\\
  Fake2M(Midjourney crawled in the website) \cite{lu2024seeing}\end{tabular} &
  600 \\ \hline
Forgery Binary Classification &
  General Subject &
  Entire Synthesis &
  Variational Auto-Encoders &
  RGB Images &
  \begin{tabular}[c]{@{}l@{}}GenImage(VQDM on ImageNet classes) \cite{zhu2024genimage};\\
  DiffusionForensics(VQ-Diffusion on LSUN) \cite{wang2023dire}\end{tabular} &
  400 \\ \hline
Forgery Binary Classification &
  General Subject &
  Entire Synthesis &
  Auto-regressive models &
  RGB Images &
  {Fake2M(Cogview)\cite{lu2024seeing}} &
  200 \\ \hline
Forgery Binary Classification &
  Human Subject &
  Face Swap (Single Face) &
  Graphics-based methods &
  Videos &
  {FF++(FaceSwap) \cite{ff++}} &
  140 \\ \hline
Forgery Binary Classification &
  Human Subject &
  Face Swap (Single Face) &
  Graphics-based methods &
  RGB Images &
  {FF++(FaceSwap) \cite{ff++}} &
  200 \\ \hline
Forgery Binary Classification &
  Human Subject &
  Face Swap (Single Face) &
  Encoder-Decoder &
  Videos &
  \begin{tabular}[c]{@{}l@{}}FF++(FaceShifter) \cite{ff++};\\
  FF++(Deepfakes) \cite{ff++};\\
  ForgeryNet(DeepFaceLab) \cite{he2021forgerynet};\\
  ForgeryNet(FaceShifter) \cite{he2021forgerynet};\\
  CelebDF-v2(Improved Deepfakes) \cite{celeb};\\
  DF-TIMIT(Improved Deepfakes) \cite{korshunov2018deepfakes,sanderson2009multi}\end{tabular} &
  1280 \\ \hline
Forgery Binary Classification &
  Human Subject &
  Face Swap (Single Face) &
  Encoder-Decoder &
  RGB Images &
   \begin{tabular}[c]{@{}l@{}}FF++(FaceShifter) \cite{ff++};\\
   FF++(Deepfakes) \cite{ff++};\\
   ForgeryNet(DeepFaceLab) \cite{he2021forgerynet};\\
   ForgeryNet(FaceShifter) \cite{he2021forgerynet};\\
   CelebDF-v2(Improved Deepfakes) \cite{celeb};\\
   DF-TIMIT(Improved Deepfakes) \cite{korshunov2018deepfakes,sanderson2009multi}\end{tabular} &
  1400 \\ \hline
Forgery Binary Classification &
  Human Subject &
  Face Swap (Single Face) &
  Variational Auto-Encoders &
  Videos &
  {DeeperForensics(DeepFake VAE) \cite{jiang2020deeperforensics}} &
  200 \\ \hline
Forgery Binary Classification &
  Human Subject &
  Face Swap (Single Face) &
  Variational Auto-Encoders &
  RGB Images &
  {DeeperForensics(DeepFake VAE) \cite{jiang2020deeperforensics}} &
  200 \\ \hline
Forgery Binary Classification &
  Human Subject &
  Face Swap (Single Face) &
  Recurrent Neural Networks &
  Videos &
  {ForgeryNet(FSGAN) \cite{he2021forgerynet};} &
  200 \\ \hline
Forgery Binary Classification &
  Human Subject &
  Face Swap (Single Face) &
  Recurrent Neural Networks &
  RGB Images &
  {ForgeryNet(FSGAN) \cite{he2021forgerynet};} &
  200 \\ \hline
Forgery Binary Classification &
  Human Subject &
  Face Swap (Single Face) &
  Unknown (in the wild) &
  Videos &
  {\begin{tabular}[c]{@{}l@{}}DFDCP \cite{dolhansky2019dee};\\
  WildDeepfake \cite{zi2020wilddeepfake};\\ DFD \cite{dfd}\end{tabular}} &
  400 \\ \hline
Forgery Binary Classification &
  Human Subject &
  Face Swap (Single Face) &
  Unknown (in the wild) &
  RGB Images &
  {\begin{tabular}[c]{@{}l@{}}DFDCP \cite{dolhansky2019dee};\\
  WildDeepfake \cite{zi2020wilddeepfake};\\ 
  DFD \cite{dfd} \end{tabular}} &
  400 \\ \hline
Forgery Binary Classification &
  Human Subject &
  Face Swap (Single Face) &
  Diffusion models &
  RGB Images &
  \begin{tabular}[c]{@{}l@{}}Diff(DiffFace) \cite{cheng2024diffusion};\\
  Diff(DCFace) \cite{cheng2024diffusion}\end{tabular} &
  400 \\ \hline
Forgery Binary Classification &
  Human Subject &
  Face Swap (Multiple Faces) &
  Encoder-Decoder,Recurrent Neural Networks,Graphics-based methods &
  Videos &
  \begin{tabular}[c]{@{}l@{}}FFIW(DeepFaceLab, FSGAN, FaceSwap) \cite{zhou2021face};\\
  DF-Platter(FaceShifter) \cite{narayan2023df}\end{tabular} &
  200 \\ \hline
Forgery Binary Classification &
  Human Subject &
  Face Swap (Multiple Faces) &
  Encoder-Decoder,Recurrent Neural Networks,Graphics-based methods &
  RGB Images &
  \begin{tabular}[c]{@{}l@{}}FFIW(DeepFaceLab, FSGAN, FaceSwap) \cite{zhou2021face};\\
  DF-Platter(FaceShifter) \cite{narayan2023df}\end{tabular} &
  200 \\ \hline
Forgery Binary Classification &
  Human Subject &
  Face Transfer &
  Graphics-based methods &
  Videos &
  {\begin{tabular}[c]{@{}l@{}}ForgeryNet(BlendFace) \cite{he2021forgerynet};\\ ForgeryNet(MMReplacement) \cite{he2021forgerynet}\end{tabular}} &
  300 \\ \hline
Forgery Binary Classification &
  Human Subject &
  Face Transfer &
  Graphics-based methods &
  RGB Images &
  {\begin{tabular}[c]{@{}l@{}}ForgeryNet(BlendFace) \cite{he2021forgerynet};\\ ForgeryNet(MMReplacement) \cite{he2021forgerynet}\end{tabular}} &
  400 \\ \hline
Forgery Binary Classification &
  Human Subject &
  Face Reenactment &
  Graphics-based methods &
  Videos &
  {FF++(Face2Face) \cite{ff++}} &
  140 \\ \hline
Forgery Binary Classification &
  Human Subject &
  Face Reenactment &
  Graphics-based methods &
  RGB Images &
  {FF++(Face2Face) \cite{ff++}} &
  200 \\ \hline
Forgery Binary Classification &
  Human Subject &
  Face Reenactment &
  Encoder-Decoder &
  Videos &
  {FF++(NeuralTextures)\cite{ff++}} &
  140 \\ \bottomrule
\end{tabular}
}
\caption{Forensics-Bench data structure (part 1): including the detailed information of $5$ designed perspectives characterizing forgeries, sample number and data sources collected under licenses.}
\label{tab:Forensics-Bench data part 1}
\end{table*}

\begin{table*}[t]
\resizebox{1.0\textwidth}{!}{%
\begin{tabular}{l|l|l|l|l|l|r}
\toprule
\textbf{Forgery Task} &
  \textbf{Forgery Semantic} &
  \textbf{Forgery Type} &
  \textbf{Forgery Model} &
  \textbf{Forgery Modality} &
  \textbf{Data Sources} &
  \multicolumn{1}{l}{\textbf{Sample Number}} \\ \hline
Forgery Binary Classification &
  Human Subject &
  Face Reenactment &
  Encoder-Decoder &
  RGB Images &
  {\begin{tabular}[c]{@{}l@{}}FF++(NeuralTextures) \cite{ff++};\\ ForgeryNet(FirstOrderMotion) \cite{he2021forgerynet}\end{tabular}} &
  400 \\ \hline
Forgery Binary Classification &
  Human Subject &
  Face Reenactment &
  Recurrent Neural Networks &
  Videos &
  {\begin{tabular}[c]{@{}l@{}}ForgeryNet(ATVG-Net) \cite{he2021forgerynet};\\ ForgeryNet(Talking-head Video) \cite{he2021forgerynet}\end{tabular}} &
  400 \\ \hline
Forgery Binary Classification &
  Human Subject &
  Face Reenactment &
  Recurrent Neural Networks &
  RGB Images &
  {\begin{tabular}[c]{@{}l@{}}ForgeryNet(ATVG-Net) \cite{he2021forgerynet};\\ ForgeryNet(Talking-head Video) \cite{he2021forgerynet}\end{tabular}} &
  400 \\ \hline
Forgery Binary Classification &
  Human Subject &
  Face Editing &
  Encoder-Decoder &
  RGB Images &
  {\begin{tabular}[c]{@{}l@{}}HiFi-IFDL(starGANv2 on CelebaHQ) \cite{guo2023hierarchical};\\ 
  HiFi-IFDL(HiSD on CelebaHQ) \cite{guo2023hierarchical};\\
  HiFi-IFDL(STGAN on CelebaHQ) \cite{guo2023hierarchical};\\
  DFFD(starGAN on CelebA) \cite{dang2020detection};\\
  ForgeryNet(starGANv2) \cite{he2021forgerynet};\\
  ForgeryNet(MaskGAN) \cite{he2021forgerynet};\\
  ForgeryNet(SC-FEGAN) \cite{he2021forgerynet};\\
  CNN-spot(starGAN) \cite{wang2020cnn}\end{tabular}} &
  1400 \\ \hline
Forgery Binary Classification &
  Human Subject &
  Style Translation &
  Encoder-Decoder &
  Near-infrared Images &
  {ForgeryNIR(CycleGAN) \cite{wang2022forgerynir}} &
  400 \\ \hline
Forgery Binary Classification &
  Human Subject &
  Face Editing &
  Proprietary &
  RGB Images &
  {DFFD(FaceAPP on FFHQ) \cite{dang2020detection}} &
  200 \\ \hline
Forgery Binary Classification &
  Human Subject &
  Face Editing &
  Diffusion models &
  RGB Images &
  {\begin{tabular}[c]{@{}l@{}}Diff(Imagic) \cite{cheng2024diffusion};\\
  Diff(CoDiff) \cite{cheng2024diffusion};\\ 
  Diff(CycleDiff) \cite{cheng2024diffusion}\end{tabular}} &
  600 \\ \hline
Forgery Binary Classification &
  General Subject &
  Style Translation &
  Encoder-Decoder &
  RGB Images &
  {\begin{tabular}[c]{@{}l@{}}CNN-spot(CycleGAN) \cite{wang2020cnn};\\
  CNN-spot(GauGAN)\cite{wang2020cnn}\end{tabular}} &
  1260 \\ \hline
Forgery Binary Classification &
  General Subject &
  Style Translation &
  Decoder &
  RGB Images &
  {\begin{tabular}[c]{@{}l@{}}CNN-spot(CRN) \cite{wang2020cnn};
  \\ CNN-spot(IMLE) \cite{wang2020cnn}\end{tabular}} &
  400 \\ \hline
Forgery Binary Classification &
  General Subject &
  Image Enhancement &
  Encoder-Decoder &
  RGB Images &
  {\begin{tabular}[c]{@{}l@{}}CNN-spot(SITD) \cite{wang2020cnn};\\
  CNN-spot(SAN) \cite{wang2020cnn}\end{tabular}} &
  380 \\ \hline
Forgery Binary Classification &
  Human Subject &
  Face Editing,Face Transfer &
  Encoder-Decoder,Graphics-based methods &
  RGB Images &
  {ForgeryNet(StarGAN2+BlendFace) \cite{he2021forgerynet}} &
  200 \\ \hline
Forgery Binary Classification &
  Human Subject &
  Face Swap (Single Face),Face Editing &
  Encoder-Decoder &
  Videos &
  {ForgeryNet(DeepFaceLab-StarGAN2) \cite{he2021forgerynet}} &
  200 \\ \hline
Forgery Binary Classification &
  Human Subject &
  Face Swap (Single Face),Face Editing &
  Encoder-Decoder &
  RGB Images &
  {ForgeryNet(DeepFaceLab-StarGAN2) \cite{he2021forgerynet}} &
  200 \\ \hline
Forgery Binary Classification &
  General Subject &
  Copy\&Move &
  Graphics-based methods &
  RGB Images &
  {HiFi-IFDL(PSCC-Net) \cite{guo2023hierarchical}} &
  200 \\ \hline
Forgery Binary Classification &
  General Subject &
  Removal &
  Encoder-Decoder &
  RGB Images &
  {HiFi-IFDL(PSCC-Net) \cite{guo2023hierarchical}} &
  200 \\ \hline
Forgery Binary Classification &
  General Subject &
  Splicing &
  Graphics-based methods &
  RGB Images &
  {HiFi-IFDL(PSCC-Net) \cite{guo2023hierarchical}} &
  200 \\ \hline
Forgery Binary Classification &
  Human Subject &
  Face Swap (Single Face) &
  Encoder-Decoder &
  RGB Images,Texts &
  {\begin{tabular}[c]{@{}l@{}}DGM4(SimSwap) \cite{shao2023detecting};\\ DGM4(InfoSwap) \cite{shao2023detecting}\end{tabular}} &
  400 \\ \hline
Forgery Binary Classification &
  Human Subject &
  Face Editing &
  Encoder-Decoder &
  RGB Images,Texts &
  {DGM4(HFGI) \cite{shao2023detecting}} &
  200 \\ \hline
Forgery Binary Classification &
  Human Subject &
  Face Editing &
  Generative Adversarial Networks &
  RGB Images,Texts &
  {DGM4(StyleCLIP) \cite{shao2023detecting}} &
  200 \\ \hline
Forgery Binary Classification &
  Human Subject &
  Text Swap &
  Retrieval-based methods &
  RGB Images,Texts &
  {DGM4(retrieval) \cite{shao2023detecting}} &
  200 \\ \hline
Forgery Binary Classification &
  Human Subject &
  Text Attribute Manipulation &
  Transformer &
  RGB Images,Texts &
  {DGM4(B-GST) \cite{shao2023detecting}} &
  200 \\ \hline
Forgery Binary Classification &
  Human Subject &
  Face Swap (Single Face),Text Swap &
  Encoder-Decoder,Retrieval-based methods &
  RGB Images,Texts &
  {\begin{tabular}[c]{@{}l@{}}DGM4(SimSwap+retrieval) \cite{shao2023detecting};\\ DGM4(InfoSwap+retrieval) \cite{shao2023detecting}\end{tabular}} &
  400 \\ \hline
Forgery Binary Classification &
  Human Subject &
  Face Editing,Text Swap &
  Encoder-Decoder,Retrieval-based methods &
  RGB Images,Texts &
  {DGM4(HFGI+retrieval) \cite{shao2023detecting}} &
  200 \\ \hline
Forgery Binary Classification &
  Human Subject &
  Face Editing,Text Swap &
  Generative Adversarial Networks,Retrieval-based methods &
  RGB Images,Texts &
  {DGM4(StyleCLIP+retrieval) \cite{shao2023detecting}} &
  200 \\ \hline
Forgery Binary Classification &
  Human Subject &
  Face Swap (Single Face),Text Attribute Manipulation &
  Encoder-Decoder,Transformer &
  RGB Images,Texts &
  {\begin{tabular}[c]{@{}l@{}}DGM4(SimSwap+B-GST) \cite{shao2023detecting};\\ DGM4(InfoSwap+B-GST) \cite{shao2023detecting}\end{tabular}} &
  400 \\ \hline
Forgery Binary Classification &
  Human Subject &
  Face Editing,Text Attribute Manipulation &
  Encoder-Decoder,Transformer &
  RGB Images,Texts &
  {DGM4(HFGI+B-GST) \cite{shao2023detecting}} &
  200 \\ \hline
Forgery Binary Classification &
  Human Subject &
  Face Editing,Text Attribute Manipulation &
  Generative Adversarial Networks,Transformer &
  RGB Images,Texts &
  {DGM4(StyleCLIP+B-GST) \cite{shao2023detecting}} &
  200 \\ \hline
Forgery Binary Classification &
  Human Subject &
  Out-of-Context &
  Retrieval-based methods &
  RGB Images,Texts &
  {NewsCLIPpings \cite{luo2021newsclippings}} &
  100 \\ \hline
Forgery Binary Classification &
  Human Subject &
  Face Spoofing &
  Print &
  RGB Images &
  {CelebA-Spoof \cite{zhang2020celeba}} &
  200 \\ \hline
Forgery Binary Classification &
  Human Subject &
  Face Spoofing &
  Paper Cut &
  RGB Images &
  {CelebA-Spoof \cite{zhang2020celeba}} &
  200 \\ \hline
Forgery Binary Classification &
  Human Subject &
  Face Spoofing &
  Replay &
  RGB Images &
  {CelebA-Spoof \cite{zhang2020celeba}} &
  200 \\ \hline
Forgery Binary Classification &
  Human Subject &
  Face Spoofing &
  3D masks &
  RGB Images &
  {CelebA-Spoof \cite{zhang2020celeba}} &
  200 \\ \hline
Forgery Spatial Localization (Segmentation) &
  Human Subject &
  Face Swap (Single Face) &
  Encoder-Decoder &
  Videos &
  {\begin{tabular}[c]{@{}l@{}}HiFi-IFDL(FaceShifter on Youtube video) \cite{guo2023hierarchical};\\
  DFFD(DeepFaceLab) \cite{dang2020detection};\\ 
  DFFD(Deepfakes) \cite{dang2020detection};\\ 
  ForgeryNet(FaceShifter) \cite{he2021forgerynet};\\ ForgeryNet(DeepFaceLab) \cite{he2021forgerynet}\end{tabular}} &
  309 \\ \hline
Forgery Spatial Localization (Segmentation) &
  Human Subject &
  Face Swap (Single Face) &
  Encoder-Decoder &
  RGB Images &
  {\begin{tabular}[c]{@{}l@{}}HiFi-IFDL(FaceShifter on Youtube video) \cite{guo2023hierarchical};\\
  DFFD(DeepFaceLab) \cite{dang2020detection};\\ 
  DFFD(Deepfakes) \cite{dang2020detection};\\
  ForgeryNet(FaceShifter) \cite{he2021forgerynet};\\
  ForgeryNet(DeepFaceLab) \cite{he2021forgerynet}\end{tabular}} &
  598 \\ \hline
Forgery Spatial Localization (Segmentation) &
  Human Subject &
  Face Swap (Single Face) &
  Graphics-based methods &
  Videos &
  {FF++(FaceSwap) \cite{ff++}} &
  140 \\ \hline
Forgery Spatial Localization (Segmentation) &
  Human Subject &
  Face Swap (Single Face) &
  Graphics-based methods &
  RGB Images &
  {FF++(FaceSwap) \cite{ff++}} &
  200 \\ \hline
Forgery Spatial Localization (Segmentation) &
  Human Subject &
  Face Swap (Single Face) &
  Recurrent Neural Networks &
  RGB Images &
  {ForgeryNet(FSGAN) \cite{he2021forgerynet}} &
  200 \\ \hline
Forgery Spatial Localization (Segmentation) &
  Human Subject &
  Face Transfer &
  Graphics-based methods &
  Videos &
  {\begin{tabular}[c]{@{}l@{}}ForgeryNet(BlendFace) \cite{he2021forgerynet};\\ ForgeryNet(MMReplacement) \cite{he2021forgerynet}\end{tabular}} &
  231 \\ \hline
Forgery Spatial Localization (Segmentation) &
  Human Subject &
  Face Transfer &
  Graphics-based methods &
  RGB Images &
  {\begin{tabular}[c]{@{}l@{}}ForgeryNet(BlendFace) \cite{he2021forgerynet};\\ ForgeryNet(MMReplacement) \cite{he2021forgerynet}\end{tabular}} &
  400 \\ \hline
Forgery Spatial Localization (Segmentation) &
  Human Subject &
  Face Reenactment &
  Graphics-based methods &
  Videos &
  {FF++(Face2Face) \cite{ff++}} &
  140\\ \bottomrule
\end{tabular}
}
\caption{Forensics-Bench data structure (part 2): including the detailed information of $5$ designed perspectives characterizing forgeries, sample number and data sources collected under licenses.}
\label{tab:Forensics-Bench data part 2}
\end{table*}

\begin{table*}[t]
\resizebox{1.0\textwidth}{!}{%
\begin{tabular}{l|l|l|l|l|l|r}
\toprule
\textbf{Forgery Task} &
  \textbf{Forgery Semantic} &
  \textbf{Forgery Type} &
  \textbf{Forgery Model} &
  \textbf{Forgery Modality} &
  \textbf{Data Sources} &
  \multicolumn{1}{l}{\textbf{Sample Number}} \\ \hline
Forgery Spatial Localization (Segmentation) &
  Human Subject &
  Face Reenactment &
  Graphics-based methods &
  RGB Images &
  {FF++(Face2Face) \cite{ff++}} &
  200 \\ \hline
Forgery Spatial Localization (Segmentation) &
  Human Subject &
  Face Reenactment &
  Encoder-Decoder &
  RGB Images &
  {ForgeryNet(FirstOrderMotion) \cite{he2021forgerynet}} &
  200 \\ \hline
Forgery Spatial Localization (Segmentation) &
  Human Subject &
  Face Reenactment &
  Recurrent Neural Networks &
  RGB Images &
  {\begin{tabular}[c]{@{}l@{}}ForgeryNet(ATVG-Net) \cite{he2021forgerynet};\\ ForgeryNet(Talking-head Video) \cite{he2021forgerynet}\end{tabular}} &
  400 \\ \hline
Forgery Spatial Localization (Segmentation) &
  Human Subject &
  Face Editing &
  Encoder-Decoder &
  RGB Images &
  {\begin{tabular}[c]{@{}l@{}}HiFi-IFDL(STGAN on CelebaHQ) \cite{guo2023hierarchical};\\
  DFFD(starGAN on CelebA) \cite{dang2020detection};\\ 
  ForgeryNet(starGANv2) \cite{he2021forgerynet};\\
  ForgeryNet(MaskGAN) \cite{he2021forgerynet};\\
  ForgeryNet(SC-FEGAN) \cite{he2021forgerynet}\end{tabular}} &
  800 \\ \hline
Forgery Spatial Localization (Segmentation) &
  Human Subject &
  Face Editing &
  Proprietary &
  RGB Images &
  {DFFD(FaceAPP on FFHQ) \cite{dang2020detection}} &
  200 \\ \hline
Forgery Spatial Localization (Segmentation) &
  Human Subject &
  Face Editing,Face Transfer &
  Encoder-Decoder,Graphics-based methods &
  RGB Images &
  {ForgeryNet(StarGAN2+BlendFace) \cite{he2021forgerynet}} &
  200 \\ \hline
Forgery Spatial Localization (Segmentation) &
  Human Subject &
  Face Swap (Single Face),Face Editing &
  Encoder-Decoder &
  Videos &
  {ForgeryNet(DeepFaceLab-StarGAN2) \cite{he2021forgerynet}} &
  100 \\ \hline
Forgery Spatial Localization (Segmentation) &
  Human Subject &
  Face Swap (Single Face),Face Editing &
  Encoder-Decoder &
  RGB Images &
  {ForgeryNet(DeepFaceLab-StarGAN2) \cite{he2021forgerynet}} &
  200 \\ \hline
Forgery Spatial Localization (Segmentation) &
  General Subject &
  Copy\&Move &
  Graphics-based methods &
  RGB Images &
  {HiFi-IFDL(PSCC-Net) \cite{guo2023hierarchical}} &
  200 \\ \hline
Forgery Spatial Localization (Segmentation) &
  General Subject &
  Removal &
  Encoder-Decoder &
  RGB Images &
  {HiFi-IFDL(PSCC-Net) \cite{guo2023hierarchical}} &
  200 \\ \hline
Forgery Spatial Localization (Segmentation) &
  General Subject &
  Splicing &
  Graphics-based methods &
  RGB Images &
  {HiFi-IFDL(PSCC-Net) \cite{guo2023hierarchical}} &
  200 \\ \hline
Forgery Spatial Localization (Segmentation) &
  Human Subject &
  Entire Synthesis &
  Generative Adversarial Networks &
  RGB Images &
  {\begin{tabular}[c]{@{}l@{}}DFFD(ProGAN) \cite{dang2020detection};\\ 
  DFFD(StyleGANv1) \cite{dang2020detection};\\ 
  ForgeryNet(StyleGANv2) \cite{he2021forgerynet};\\ 
  ForgeryNet(DiscoFaceGAN) \cite{he2021forgerynet}\end{tabular}} &
  800 \\ \hline
Forgery Spatial Localization (Segmentation) &
  Human Subject &
  Face Swap (Multiple Faces) &
  Generative Adversarial Networks &
  RGB Images &
  {OpenForensics \cite{le2021openforensics}} &
  200 \\ \hline
Forgery Spatial Localization (Detection) &
  Human Subject &
  Face Swap (Multiple Faces) &
  Generative Adversarial Networks &
  RGB Images &
  {OpenForensics \cite{le2021openforensics}} &
  200 \\ \hline
Forgery Spatial Localization (Segmentation) &
  Human Subject &
  Face Swap (Multiple Faces) &
  Graphics-based methods,Recurrent Neural Networks,Encoder-Decoder &
  Videos &
  {FFIW(DeepFaceLab, FSGAN, FaceSwap) \cite{zhou2021face}} &
  200 \\ \hline
Forgery Spatial Localization (Segmentation) &
  Human Subject &
  Face Swap (Multiple Faces) &
  Graphics-based methods,Recurrent Neural Networks,Encoder-Decoder &
  RGB Images &
  {FFIW(DeepFaceLab, FSGAN, FaceSwap) \cite{zhou2021face}} &
  200 \\ \hline
Forgery Spatial Localization (Detection) &
  Human Subject &
  Face Swap (Single Face) &
  Encoder-Decoder &
  RGB Images,Texts &
  {\begin{tabular}[c]{@{}l@{}}DGM4(SimSwap) \cite{shao2023detecting};\\ DGM4(InfoSwap) \cite{shao2023detecting}\end{tabular}} &
  400 \\ \hline
Forgery Spatial Localization (Detection) &
  Human Subject &
  Face Editing &
  Encoder-Decoder &
  RGB Images,Texts &
  {DGM4(HFGI) \cite{shao2023detecting}} &
  200 \\ \hline
Forgery Spatial Localization (Detection) &
  Human Subject &
  Face Editing &
  Generative Adversarial Networks &
  RGB Images,Texts &
  {DGM4(StyleCLIP) \cite{shao2023detecting}} &
  200 \\ \hline
Forgery Spatial Localization (Detection) &
  Human Subject &
  Text Swap &
  Retrieval-based methods &
  RGB Images,Texts &
  {DGM4(retrieval) \cite{shao2023detecting}} &
  200 \\ \hline
Forgery Spatial Localization (Detection) &
  Human Subject &
  Text Attribute Manipulation &
  Transformer &
  RGB Images,Texts &
  {DGM4(B-GST) \cite{shao2023detecting}} &
  200 \\ \hline
Forgery Spatial Localization (Detection) &
  Human Subject &
  Face Swap (Single Face),Text Swap &
  Encoder-Decoder,Retrieval-based methods &
  RGB Images,Texts &
  {\begin{tabular}[c]{@{}l@{}}DGM4(SimSwap+retrieval) \cite{shao2023detecting};\\ DGM4(InfoSwap+retrieval) \cite{shao2023detecting}\end{tabular}} &
  400 \\ \hline
Forgery Spatial Localization (Detection) &
  Human Subject &
  Face Editing,Text Swap &
  Encoder-Decoder,Retrieval-based methods &
  RGB Images,Texts &
  {DGM4(HFGI+retrieval) \cite{shao2023detecting}} &
  200 \\ \hline
Forgery Spatial Localization (Detection) &
  Human Subject &
  Face Editing,Text Swap &
  Generative Adversarial Networks,Retrieval-based methods &
  RGB Images,Texts &
  {DGM4(StyleCLIP+retrieval) \cite{shao2023detecting}} &
  200 \\ \hline
Forgery Spatial Localization (Detection) &
  Human Subject &
  Face Swap (Single Face),Text Attribute Manipulation &
  Encoder-Decoder,Transformer &
  RGB Images,Texts &
  {\begin{tabular}[c]{@{}l@{}}DGM4(SimSwap+B-GST) \cite{shao2023detecting};\\ DGM4(InfoSwap+B-GST) \cite{shao2023detecting}\end{tabular}} &
  400 \\ \hline
Forgery Spatial Localization (Detection) &
  Human Subject &
  Face Editing,Text Attribute Manipulation &
  Encoder-Decoder,Transformer &
  RGB Images,Texts &
  {DGM4(HFGI+B-GST) \cite{shao2023detecting}} &
  200 \\ \hline
Forgery Spatial Localization (Detection) &
  Human Subject &
  Face Editing,Text Attribute Manipulation &
  Generative Adversarial Networks,Transformer &
  RGB Images,Texts &
  {DGM4(StyleCLIP+B-GST) \cite{shao2023detecting}} &
  200 \\ \hline
Forgery Temporal Localization &
  Human Subject &
  Face Swap (Single Face) &
  Encoder-Decoder &
  Videos &
  {\begin{tabular}[c]{@{}l@{}}ForgeryNet(DeepFaceLab) \cite{he2021forgerynet};\\ ForgeryNet(FaceShifter) \cite{he2021forgerynet}\end{tabular}} &
  400 \\ \hline
Forgery Temporal Localization &
  Human Subject &
  Face Swap (Single Face) &
  Recurrent Neural Networks &
  Videos &
  {ForgeryNet(FSGAN) \cite{he2021forgerynet}} &
  200 \\ \hline
Forgery Temporal Localization &
  Human Subject &
  Face Transfer &
  Graphics-based methods &
  Videos &
  {\begin{tabular}[c]{@{}l@{}}ForgeryNet(BlendFace) \cite{he2021forgerynet};\\ ForgeryNet(MMReplacement) \cite{he2021forgerynet}\end{tabular}} &
  300 \\ \hline
Forgery Temporal Localization &
  Human Subject &
  Face Reenactment &
  Recurrent Neural Networks &
  Videos &
  {\begin{tabular}[c]{@{}l@{}}ForgeryNet(ATVG-Net) \cite{he2021forgerynet};\\ ForgeryNet(Talking-head Video) \cite{he2021forgerynet}\end{tabular}} &
  400 \\ \hline
Forgery Temporal Localization &
  Human Subject &
  Face Swap (Single Face),Face Editing &
  Encoder-Decoder &
  Videos &
  {ForgeryNet(DeepFaceLab-StarGAN2) \cite{he2021forgerynet}} &
  200 \\ \hline
Forgery Binary Classification &
  Human Subject &
  Real &
  Real &
  RGB Images,Texts &
  {DGM4 \cite{shao2023detecting}} &
  2000 \\ \hline
Forgery Binary Classification &
  Human Subject &
  Real &
  Real &
  RGB Images &
  {\begin{tabular}[c]{@{}l@{}}DFFD(FFHQ) \cite{dang2020detection};\\ DiffusionForensics(CelebAHQ) \cite{wang2023dire};\\ DeeperForensics \cite{jiang2020deeperforensics};\\ FF++ \cite{ff++};\\ CelebDF-v2 \cite{celeb};\\ FFIW \cite{zhou2021face};\\ CelebA-Spoof \cite{zhang2020celeba}\end{tabular}} &
  4000 \\ \hline
Forgery Binary Classification &
  General Subject &
  Real &
  Real &
  RGB Images &
  {\begin{tabular}[c]{@{}l@{}}CNN-spot \cite{wang2020cnn};\\ DiffusionForensics(LSUN, ImageNet) \cite{wang2023dire};\\ COCO2017val \cite{COCO}\end{tabular}} &
  4000 \\ \hline
Forgery Binary Classification &
  Human Subject &
  Real &
  Real &
  Videos &
  {\begin{tabular}[c]{@{}l@{}}FF++ \cite{ff++};\\ CelebDF-v2 \cite{celeb};\\ DeeperForensics \cite{jiang2020deeperforensics};\\ FFIW \cite{zhou2021face};\\ CelebA-Spoof \cite{zhang2020celeba}\end{tabular}} &
  178 \\ \hline
Forgery Spatial Localization (Segmentation) &
  Human Subject &
  Real &
  Real &
  RGB Images &
  {\begin{tabular}[c]{@{}l@{}}DFFD(FFHQ) \cite{dang2020detection};\\ DiffusionForensics(CelebAHQ) \cite{wang2023dire};\\ DeeperForensics \cite{jiang2020deeperforensics};\\ FF++ \cite{ff++};\\ CelebDF-v2 \cite{celeb};\\ FFIW \cite{zhou2021face};\\ CelebA-Spoof \cite{zhang2020celeba}\end{tabular}} &
  1600 \\ \hline
Forgery Spatial Localization (Segmentation) &
  General Subject &
  Real &
  Real &
  RGB Images &
  {\begin{tabular}[c]{@{}l@{}}CNN-spot \cite{wang2020cnn};\\ DiffusionForensics(LSUN, ImageNet) \cite{wang2023dire}\\ COCO2017val \cite{COCO}\end{tabular}} &
  1500 \\ \hline
Forgery Spatial Localization (Segmentation) &
  Human Subject &
  Real &
  Real &
  Videos &
  {\begin{tabular}[c]{@{}l@{}}FF++ \cite{ff++};\\ CelebDF-v2 \cite{celeb};\\ DeeperForensics \cite{jiang2020deeperforensics};\\ FFIW \cite{zhou2021face}\end{tabular}} &
  178 \\ \hline
Forgery Spatial Localization (Detection) &
  Human Subject &
  Real &
  Real &
  RGB Images,Texts &
  {DGM4 \cite{shao2023detecting}} &
  1000 \\ \hline
Forgery Spatial Localization (Detection) &
  Human Subject &
  Real &
  Real &
  RGB Images &
  \begin{tabular}[c]{@{}l@{}}DFFD(FFHQ) \cite{dang2020detection};\\ DiffusionForensics(CelebAHQ) \cite{wang2023dire};\\ DeeperForensics \cite{jiang2020deeperforensics};\\ FF++ \cite{ff++};\\ CelebDF-v2 \cite{celeb};\\ FFIW \cite{zhou2021face};\\ CelebA-Spoof \cite{zhang2020celeba}\end{tabular} &
  1100 \\ \hline
Forgery Spatial Localization (Detection) &
  General Subject &
  Real &
  Real &
  RGB Images &
  \begin{tabular}[c]{@{}l@{}}CNN-spot \cite{wang2020cnn};\\ DiffusionForensics(LSUN, ImageNet) \cite{wang2023dire}\end{tabular} &
  1000 \\ \hline
Forgery Temporal Localization &
  Human Subject &
  Real &
  Real &
  Videos &
  ForgeryNet \cite{he2021forgerynet}&
  378 \\ \bottomrule
\end{tabular}
}
\caption{Forensics-Bench data structure (part 3): including the detailed information of $5$ designed perspectives characterizing forgeries, sample number and data sources collected under licenses.}
\label{tab:Forensics-Bench data part 3}
\end{table*}

\section{Other Details of Forensics-Bench}
\label{sec:details_of_FDBENCH}

\begin{table*}[t]
\centering
\resizebox{0.8\textwidth}{!}{%
\begin{tabular}{l|l|l}
\toprule
    Keys & Example 1 & Example 2 \\
    \midrule
    Image Path & /path/to/image & /path/to/image \\
    Image Resolution &299x299& 1280x720\\
    Data Source & DFFD\_StyleGANv1\_ffhq & ForgeryNet\_12\_seg \\
    Forgery Semantic & Human & Human \\
    Forgery Modality & RGB Image& RGB Image\\
    Forgery Task & Forgery Binary Classification & Forgery Spatial Localization (Segmentation) \\
    Forgery Type&Entire Synthesis & Face Editing\\
    Forgery Model&Generative Adversarial Networks &Encoder-Decoder\\
    Question Template & What is the authenticity of the image? &  Which segmentation map denotes the forged area in the image most accurately?\\
    Choice List&[AI-generated, non AI-generated]&[Candidate 1, Candidate 2, Candidate 3, Candidate 4]\\
    Answer & AI-generated &  Candidate 4  \\
    \bottomrule
    \end{tabular}%
}
\caption{Examples of the uniformed metadata.}
\label{tab:metadata_example}
\end{table*}

\noindent \textbf{Uniformed metadata}. 
In our benchmark, we design a uniformed metadata structure to standardize and accelerate the construction process of our data samples. 
As shown in Table \ref{tab:metadata_example}, the metadata structure is a dictionary with keys divided into three main categories. 
The first category contains keys such as the image path, image resolution and data source, describing the vanilla information about the raw data.
The second category includes keys demonstrating the detailed information of 5 designed perspectives in our benchmark. 
The third category includes keys for the transformed Q\&A, such as the question template, answer (ground truth) and choice list.

\noindent \textbf{Details of forgery types}. 
In our benchmark, we roughly classify previous forgeries into $21$ types, which are summarized as follows.
\begin{itemize}
    \item Entire Synthesis: In our benchmark, this refers to forgeries that are synthesized from scratch without a basis on real media. For instance, vanilla GAN models and diffusion models can generate forgeries from random Gaussian noises. Representative datasets of this type include CNN-spot \cite{wang2020cnn} and DiffusionForensics \cite{wang2023dire}.
    \item Spoofing: In our benchmark, this refers to forgeries that present a fake version of a legitimate user's face to bypass authentication, such as the printed photograph of a user's face, a recorded video of the target user and 3D masks that mimic the target's facial structures. Representative datasets of this type include CelebA-Spoof \cite{zhang2020celeba}.
    \item Face Editing: In our benchmark, this refers to forgeries that modify the external attributes of human faces, such as facial hair, age and gender. Representative datasets of this type include ForgeryNet \cite{he2021forgerynet}.
    \item Face Swap (Single Face): In our benchmark, this refers to forgeries that exchange one person's facial features with another, changing the original identity of the depicted person. Representative datasets of this type include CelebDF-v2 \cite{celeb}.
    \item Face Swap (Multiple Faces): In our benchmark, this refers to forgeries that exchange more than one person's facial features with other human faces in one media. Representative datasets of this type include OpenForensics \cite{le2021openforensics}.
    \item Face Transfer: In our benchmark, this refers to forgeries that transfer both the identity-aware and identity-agnostic content (such as the pose and expression) of the source face to the target face. This follows the design proposed in ForgeryNet \cite{he2021forgerynet}.
    \item Face Reenactment: In our benchmark, this refers to forgeries that transfer the facial expressions, movements, and emotions of one person's face to another person's face. Representative datasets of this type include FF++ \cite{ff++}.
    \item Copy-Move: In our benchmark, this refers to forgeries that involve copying a portion of an image and pasting it elsewhere within the same image. Representative datasets of this type include HiFi-IFDL \cite{guo2023hierarchical}.
    \item Removal: In our benchmark, this refers to forgeries that involve removing an object or region from an image and filling in the removed area to maintain the visual coherence, which is also known as ``inpainting". Representative datasets of this type include HiFi-IFDL \cite{guo2023hierarchical}.
    \item Splicing: In our benchmark, this refers to forgeries that involve combining elements from two or more different images to create a composite image. Representative datasets of this type include HiFi-IFDL \cite{guo2023hierarchical}.
    \item Image Enhancement: In our benchmark, this refers to forgeries where enhancements are deliberately applied to alter the appearance of an image, such as image super-resolution and low-light imaging. Representative datasets of this type include CNN-spot \cite{wang2020cnn}.
    \item Out-of-Context: In our benchmark, this refers to forgeries where the presentation of an authentic image, video, or media clip is repurposed with a misleading or deceptive text. Representative datasets of this type include NewsCLIPpings \cite{luo2021newsclippings}.
    \item Style Translation: In our benchmark, this refers to forgeries which transform the visual style of one image while preserving the content of another image. Representative datasets of this type include CNN-spot \cite{wang2020cnn}.
    \item Text Attribute Manipulation: In our benchmark, this refers to forgeries that alter the sentiment tendency of a given text while preserving its core content or meaning. This follows the design in DGM4 \cite{shao2023detecting}. 
    \item Text Swap: In our benchmark, this refers to forgeries that alter the overall semantic of a text with word substitution while preserving words regarding the main character. This follows the design in DGM4 \cite{shao2023detecting}. 
    \item Face Editing \& Text Attribute Manipulation: In our benchmark, this refers to forgeries that are produced under the combination of both Face Editing \& Text Attribute Manipulation. This follows the design in DGM4 \cite{shao2023detecting}. 
    \item Face Editing \& Text Swap: In our benchmark, this refers to forgeries that are produced under the combination of both Face Editing \& Text Swap. This follows the design in DGM4 \cite{shao2023detecting}. 
    \item Face Editing \& Face Transfer: In our benchmark, this refers to forgeries that are produced under the combination of both Face Editing \& Face Transfer. This follows the design in ForgeryNet \cite{he2021forgerynet}. 
    \item Face Swap (Single Face) \& Face Editing: In our benchmark, this refers to forgeries that are produced under the combination of both Face Swap (Single Face) \& Face Editing. This follows the design in ForgeryNet \cite{he2021forgerynet}. 
    \item Face Swap (Single Face) \& Text Attribute Manipulation: In our benchmark, this refers to forgeries that are produced under the combination of both Face Swap (Single Face) \& Text Attribute Manipulation. This follows the design in DGM4 \cite{shao2023detecting}. 
    \item Face Swap (Single Face) \& Text Swap: In our benchmark, this refers to forgeries that are produced under the combination of both Face Swap (Single Face) \& Text Swap. This follows the design in DGM4 \cite{shao2023detecting}. 
\end{itemize}

\noindent \textbf{Details of forgery models}. 
In our benchmark, we roughly divide previous forgeries into 22 categories from the perspective of forgery model. 
We summarize the details as follows.
\begin{itemize}
    \item Generative Adversarial Networks: In our benchmark, this refers to forgeries that are generated with vanilla GANs, namely a pair of adversarially trained generator and discriminator. Representative datasets of this category include CNN-spot \cite{wang2020cnn}.
    \item Diffusion models: In our benchmark, this refers to forgeries that are generated with vanilla diffusion models, such as DDPM \cite{ho2020denoising}. Representative datasets of this category include DiffusionForensics \cite{wang2023dire}.
    \item Encoder-Decoder: In our benchmark, this represents forgery models which commonly take real media as input, and are typically used to separate the identity information from identity-agnostic attributes, then alter or exchange the facial representations. This kind of models usually features an encoder-decoder structure. This follows the design in ForgeryNet \cite{he2021forgerynet} and representative datasets of this category include CelebDF-v2 \cite{celeb} and FF++ \cite{ff++}.
    \item Recurrent Neural Networks: In our benchmark, this represents forgery models which are commonly used to alter sequential and dynamic-length data like videos. This follows the design in ForgeryNet \cite{he2021forgerynet}.
    \item Proprietary: In our benchmark, this represents closed-source forgery models commonly used for commercial purposes, like Midjourney. Representative datasets of this category include GenImage \cite{zhu2024genimage}.
    \item 3D masks: In our benchmark, this represents forgeries which are produced based on 3D masks designed to look like real users, commonly used for face spoofing. Representative datasets of this category include CelebA-Spoof \cite{zhang2020celeba}.
    \item Print: In our benchmark, this represents forgeries which are produced based on a printed photograph of a face, in order to trick facial recognition systems. Representative datasets of this category include CelebA-Spoof \cite{zhang2020celeba}.
    \item Paper-Cut: In our benchmark, this represents forgeries which are produced based on a printed photograph of a face with specific modifications, such as eye and mouth cutouts. Representative datasets of this kind include CelebA-Spoof \cite{zhang2020celeba}.
    \item Replay: In our benchmark, this represents forgeries which are produced by displaying a recorded video or image sequence of the face on a screen. Representative datasets of this category include CelebA-Spoof \cite{zhang2020celeba}.
    \item Transformer: In our benchmark, this represents forgery models that are mainly used to modify texts, such as altering the sentiment tendency. Representative datasets of this category include DGM4 \cite{shao2023detecting}.
    \item Decoder: In our benchmark, this represents forgery models which are mainly used to perform style translations, commonly featuring a decode-only structure. Representative datasets of this category include CNN-spot \cite{wang2020cnn}.
    \item Graphics-based methods: In our benchmark, this represents forgeries that are mainly produced with traditional graphics formations. This follows the design in ForgeryNet \cite{he2021forgerynet}.
    \item Retrieval-based methods: In our benchmark, this represents forgeries that are produced by retrieving existing data. Representative datasets of this category include DGM4 \cite{shao2023detecting}.
    \item Unknown (in the wild): In our benchmark, this represents forgeries with unknown sources. Representative datasets of this category include DFPCP \cite{dolhansky2019dee}.
    \item Variational Auto-Encoders: In our benchmark, this represents forgeries that are generated with typical Variational Auto-Encoders. Representative datasets of this category include DeeperForensics \cite{jiang2020deeperforensics}.
    \item Auto-regressive models: In our benchmark, this represents forgery models which are commonly used to generate video data with no basis of real media, such as CogVideo \cite{hong2022cogvideo}. 
    \item Encoder-Decoder\&Retrieval-based methods: In our benchmark, this represents forgeries that are produced under the combination of Encoder-Decoder\&Retrieval-based methods. This follows the design in DGM4 \cite{shao2023detecting}.
    \item Encoder-Decoder\&Recurrent Neural Networks\&Graphics-based methods: In our benchmark, this represents forgeries that are produced under the combination of Encoder-Decoder\&Recurrent Neural Networks\&Graphics-based methods. Representative datasets of this category include FFIW \cite{zhou2021face}.
    \item Generative Adversarial Networks\&Retrieval-based methods: In our benchmark, this represents forgeries that are produced by the combination of Generative Adversarial Networks\&Retrieval-based methods. This follows the design in DGM4 \cite{shao2023detecting}.
    \item Encoder-Decoder\&Transformer: In our benchmark, this represents forgeries that are produced under the combination of Encoder-Decoder\&Transformer. This follows the design in DGM4 \cite{shao2023detecting}.
    \item Generative Adversarial Networks\&Transformer: In our benchmark, this represents forgeries that are produced under the combination of Generative Adversarial Networks\&Transformer. This follows the design in DGM4 \cite{shao2023detecting}.
    \item Encoder-Decoder\&Graphics-based methods: In our benchmark, this represents forgeries that are produced under the combination of Encoder-Decoder\&Graphics-based methods. This follows the design in ForgeryNet \cite{he2021forgerynet}.
\end{itemize}

\noindent \textbf{Details of forgery tasks}.
In our benchmark, we roughly divide previous forgeries into 4 categories from the perspective of forgery task. 
We summarize the details as follows.
\begin{itemize}
    \item Forgery Binary Classification: This task aims to identify whether a given input (image, video, or text) is genuine or fake (forged). For instance, we can design the question template as \textit{What is the authenticity of the image?} with two choice selections \textit{AI-generated} and \textit{non AI-generated} for HS-RGB-BC-ES-GAN (Please refer to Table \ref{tab:abbreviation} for the full term).
    \item Forgery Spatial Localization (Detection): This task aims to determine the specific regions within the input that have been altered, tampered with, or manipulated. For instance, we can design the question template as \textit{Please detect the forged area in this image and the forged text in the corresponding caption: ``Gen Prayuth Chanocha says democracy will only return after reforms are put in place". The output format for the forged area should be a list of bounding boxes, namely [x, y, w, h], representing the coordinates of the top-left corner of the bounding box, as well as the width and height of the bounding box. The width of the input image is 624 and the height is 351. The output format for the forged text should be the a list of token positions in the whole caption, where the initial position index starts from 0. The token length of the input caption is 14.}. The corresponding choice list is: \textit{A.\{
                ``forged area": [
                    [
                        274,
                        46,
                        358,
                        167
                    ]
                ],
                ``forged text": []
            \},
        B. \{
            ``forged area": [
                [
                    274,
                    46,
                    358,
                    167
                ],
                [
                    220,
                    35,
                    330,
                    169
                ]
            ],
            ``forged text": [
                5
            ]
        \},
        C. \{
            ``forged area": [
                [
                    274,
                    46,
                    358,
                    167
                ],
                [
                    186,
                    122,
                    333,
                    196
                ]
            ],
            ``forged text": [
                1,
                6
            ]
        \},
        D. \{
            ``forged area": [
                [
                    274,
                    46,
                    332,
                    141
                ],
                [
                    1,
                    120,
                    295,
                    192
                ]
            ],
            ``forged text": []
        \}}. This example is for HS-RGB\&TXT-SLD-FE-ED (Please refer to Table \ref{tab:abbreviation} for the full term). 
    \item Forgery Spatial Localization (Segmentation): This task aims to precisely outline the regions of tampered or manipulated content within the digital media using pixel-wise classification. For instance, we can design the question template as \textit{Which segmentation map denotes the forged area in the image most accurately?} with four choice selections \textit{[Candidate 1, Candidate 2, Candidate 3, Candidate 4]}, each of which points to a segmentation map. This example is for HS-RGB-SLS-FE-ED (Please refer to Table \ref{tab:abbreviation} for the full term). 
    \item Forgery Temporal Localization: This task aims to detect the tampered or manipulated segments within a video. For instance, we can design the question template as \textit{Please locate the forged frames in the given set of frames, which are sampled from a video. The output format should be the a list of indexes indicating the forged frames. The initial index starts from 0.}. The corresponding choice list is: \textit{A. [
            0,
            1,
            5
        ],
        B. [
            1
        ],
        C. [
            0
        ],
        D. [
            0,
            1
        ]}. This example is for HS-VID-TL-FSS-ED (Please refer to Table \ref{tab:abbreviation} for the full term).
\end{itemize}

\section{LVLMs Model Details}
In this section, we present the summary of the LVLMs utilized in this paper, detailing their parameter sizes, visual encoders, and LLMs, which is shown in Table \ref{tab:lvlm_detail}. 
We follow the evaluation tool \cite{duan2024vlmevalkit} provided in OpenCompass \cite{2023opencompass} for the evaluations.

\begin{table*}[t]
    \centering
\resizebox{0.8\textwidth}{!}
    {
    \begin{tabular}{llll}
        \toprule
        Models & Parameters & Vision Encoder & LLM \\ 
        \midrule
        GPT4o \citep{gpt4o} & - & - & -  \\ 
        Gemini1.5 ProVision \citep{gemini} & - & - & -  \\ 
        Claude3.5-Sonnet \citep{Claude2023} & - & - & -  \\ 
        \midrule
        LLaVA-Next-34B \cite{liu2024llavanext} & 34.8B &CLIP ViT-L/14& Nous-Hermes-2-Yi-34B \\
        LLaVA-v1.5-7B-XTuner \cite{2023xtuner} &7.2B &CLIP ViT-L/14 &Vicuna-v1.5-7B\\
        LLaVA-v1.5-13B-XTuner \cite{2023xtuner}& 13.4B& CLIP ViT-L/14 &Vicuna-v1.5-13B\\
        InternVL-Chat-V1-2 \cite{internvl_chat,2023internlm} & 40B &InternViT-6B& Nous-Hermes-2-Yi-34B\\
        LLaVA-NEXT-13B \cite{liu2024llavanext} &13.4B& CLIP ViT-L/14& Vicuna-v1.5-13B\\
        mPLUG-Owl2 \cite{mplug}& 8.2B& CLIP ViT-L/14& LLaMA2-7B\\
        LLaVA-v1.5-7B \cite{liu2023llava,liu2023improvedllava}&7.2B& CLIP ViT-L/14 &Vicuna-v1.5-7B\\
        LLaVA-v1.5-13B \cite{liu2023llava,liu2023improvedllava}&13.4B& CLIP ViT-L/14& Vicuna-v1.5-13B\\
        Yi-VL-34B \cite{young2024yi}&34.6B& CLIP ViT-H/14& Nous-Hermes-2-Yi-34B\\
        CogVLM-Chat \cite{wang2023cogvlm}& 17B& EVA-CLIP-E& Vicuna-v1.5-7B\\
        XComposer2 \cite{internlmxcomposer2}&7B& CLIP ViT-L/14 &InternLM2-7B\\
        LLaVA-InternLM2-7B \cite{2023xtuner}&8.1B& CLIP ViT-L/14 &InternLM2-7B\\
        VisualGLM-6B &8B& EVA-CLIP &ChatGLM-6B\\
        LLaVA-NEXT-7B \cite{liu2024llavanext}&7.1B& CLIP ViT-L/14& Vicuna-v1.5-7B\\
        LLaVA-InternLM-7B \cite{2023xtuner}&7.6B& CLIP ViT-L/14& InternLM-7B \\
        ShareGPT4V-7B \cite{chen2023sharegpt4v}&7.2B& CLIP ViT-L/14& Vicuna-v1.5-7B\\
        InternVL-Chat-V1-5 \cite{internvl_chat,2023internlm}&40B& InternViT-6B& Nous-Hermes-2-Yi-34B\\
        DeepSeek-VL-7B \cite{lu2024deepseek}&7.3B& SAM-B \& SigLIP-L& DeekSeek-7B\\
        Yi-VL-6B \cite{young2024yi}&6.6B& CLIP ViT-H/14& Yi-6B\\
        InstructBLIP-13B \cite{instructblip}&13B& EVA-CLIP ViT-G/14& Vicuna-v1.5-13B\\
        Qwen-VL-Chat \cite{Qwen-VL}& 9.6B & CLIP ViT-G/16 & Qwen-7B \\
        Monkey-Chat \cite{li2023monkey}& 9.8B & CLIP-ViT-BigHuge & Qwen-7B \\
        \bottomrule
    \end{tabular}
    }
 \caption{Model architecture of $25$ LVLMs evaluated on Forensics-Bench.}
\label{tab:lvlm_detail}
\end{table*}

\section{Additional Experiments}

\noindent \textbf{Single-image input \textit{vs} Multi-images input}. 
The ability to process multiple images is essential for large vision language models, which may also facilitate LVLMs to understand forgeries of sequential data like videos.
For example, frames of a real video may transition smoothly and naturally, whereas a fake video may exhibit inter-frame inconsistencies.
To this end, we propose to analyze the effects of single-image prompt and multi-images prompt on LVLMs with capabilities to understand multiple images.
Specifically, we collect the subset of our Forensics-Bench featuring video modality, and feed LVLMs with single-image input and multi-images input. 
Note that the single-image input is generated by piecing together sampled frames into one big input image, as shown in Figure \ref{fig:Forensics-Bench}.
The results are demonstrated in Table \ref{tab:multi_vs_single}, where the evaluated LVLMs also support multiple images as input.
We find that LVLMs, like InternVL-Chat-V1-2 and Gemini-1.5-Pro, effectively exploited the relations among frames to perform forgery detections, while other LVLMs faced challenges in extracting meaningful information to determine the authenticity of the input frames, highlighting the unique difficulties of video forgery detections.

\noindent \textbf{Experiments on prompt engineering}.
In the main paper, we mainly focused on baseline evaluations, following the OpenCompass \cite{2023opencompass} protocol and using default system prompts recommended by each LVLM, which are already well-trained.
Nevertheless, beyond the baseline results, we have conducted experiments, adding a new forgery-related prompt: ``\textit{Please make your decision using forgery detection techniques, such as examining facial features, blending artifacts, lighting irregularities, and any other inconsistencies that may indicate manipulations.}". 
Results in Table \ref{tab: rebuttal_A} show guiding LVLMs to focus on such forgery traces boosted performance to some extent, which may inspire future studies. 

\noindent \textbf{More experiments on forgery attribution}. In this section, we explore methods to enhance LVLMs' performance on the task of forgery attribution.
To this end, we have conducted experiments by adding detailed introductions of different forgery models into the prompt, as detailed in Appendix \ref{sec:details_of_FDBENCH}, aiming to reduce LVLMs' potential misunderstandings for forgery attribution.
Results in Table \ref{tab: rebuttal_B} show that this improved LVLMs' performance, which may inspire future studies. 

\noindent \textbf{Experiments on visual prompt engineering}. In this section, we have conducted experiments where we added bounding boxes to human subjects for forgery binary classification and prompted LVLMs to focus on these image regions.
Table \ref{tab: rebuttal_F} shows that such visual prompts boosted performance to some extent, which may inspire future studies.

\begin{table*}[t]
\centering
\resizebox{0.8\textwidth}{!}{%
\begin{tabular}{l|l|l|l|l|l|l}
\toprule
    Model &InternVL-Chat-V1-2&mPLUG-Owl2&Gemini-1.5-Pro&InternVL-Chat-V1-5&Qwen-VL-Chat&Claude3V-Sonnet\\
    \midrule
    Single-Image Prompt &62.9&59.8&38.8&52.2&38.9&35.9\\
    \midrule
    Multi-Images Prompt & 63.9&36.3&40.9&34.8&25.7&30.2\\
    \bottomrule
    \end{tabular}%
}
\caption{The performance comparison between single-image input and multi-images input.}
\label{tab:multi_vs_single}
\end{table*}

\begin{table*}[t]
\begin{center}
{
\linespread{0.6}
\setlength\tabcolsep{3pt}
\scriptsize
{%
\begin{tabular}{l|c|c}
\toprule
Model&Baseline& +Prompt Engineering \\ \midrule
LLaVA-v1.5-7B-XTuner&65.7&\textbf{67.6} \\ \midrule
LLaVA-v1.5-13B-XTuner&65.2&\textbf{67.1}\\ \midrule
% LLaVA-NEXT-34B&61.8\\ \midrule
LLaVA-NEXT-13B&58.0&\textbf{61.3}\\ 
% InternVL-Chat-V1-2&50.7\\
\bottomrule
\end{tabular}%
}
}
\end{center}
\caption{Experiments on prompt engineering.}
\label{tab: rebuttal_A}
\end{table*}

\begin{table*}[t]
\begin{center}
{
\linespread{0.6}
\setlength\tabcolsep{3pt}
\scriptsize
{%
\begin{tabular}{l|c|c}
\toprule
Model&Baseline&+Detailed Introductions of Forgery Models \\ \midrule
LLaVA-NEXT-34B&44.0&\textbf{55.7}\\ \midrule
InternVL-Chat-V1-2&41.6&\textbf{55.6}\\\midrule
LLaVA-v1.5-7B-XTuner&42.2 &\textbf{49.6}\\ \midrule
mPLUG-Owl2&39.9&\textbf{45.4}\\
\bottomrule
\end{tabular}%
}
}
\end{center}
\caption{More experiments on forgery attribution.}
\label{tab: rebuttal_B}
\end{table*}

\begin{table*}[t]
\begin{center}
{
\linespread{0.6}
\setlength\tabcolsep{3pt}
\scriptsize
{%
\begin{tabular}{l|c|c}
\toprule
Model&Baseline & +Prompt Engineering (Visual) \\ \midrule
LLaVA-v1.5-7B-XTuner&83.5 &\textbf{87.6}\\ \midrule
LLaVA-NEXT-34B&84.1&\textbf{85.7}\\ \midrule
InternVL-Chat-V1-2&84.5&\textbf{86.5}\\\midrule
LLaVA-NEXT-13B&68.2&\textbf{70.4}\\
\bottomrule
\end{tabular}%
}
}
\end{center}
\caption{Experiments on visual prompt engineering.}
\label{tab: rebuttal_F}
\end{table*}

\section{Detailed Performance of LVLMs on Forensics-Bench}
\label{sec:detailed performance}
From Table \ref{tab:detail results part 1} to Table \ref{tab:detail results part 12}, we present the detailed performance of 25 state-of-the-art LVLMs across 112 forgery detection types, with the accuracy as the metric. 
Please refer to Table \ref{tab:abbreviation} for the full term of each column title.

\begin{table*}[t]
\resizebox{1.0\textwidth}{!}{%
\begin{tabular}{c|l|l|l|l|l|l|l|l|l|l}
\hline
\textbf{Model} &
  \multicolumn{1}{c|}{\textbf{HS-RGB-BC-ES-DF}} &
  \multicolumn{1}{c|}{\textbf{HS-NIR-BC-ES-GAN}} &
  \multicolumn{1}{c|}{\textbf{HS-RGB-BC-ES-GAN}} &
  \multicolumn{1}{c|}{\textbf{HS-RGB-SLS-ES-GAN}} &
  \multicolumn{1}{c|}{\textbf{HS-RGB-BC-ES-PRO}} &
  \multicolumn{1}{c|}{\textbf{HS-RGB-BC-SPF-3D}} &
  \multicolumn{1}{c|}{\textbf{HS-RGB-BC-SPF-PC}} &
  \multicolumn{1}{c|}{\textbf{HS-RGB-BC-SPF-PR}} &
  \multicolumn{1}{c|}{\textbf{HS-RGB-BC-SPF-RP}} &
  \multicolumn{1}{c}{\textbf{HS-RGB-BC-FE-DF}} \\ \hline
LLaVA-NEXT-34B        & 90.8\% & 100.0\% & 85.4\% & 19.3\% & 97.0\% & 100.0\% & 99.0\%  & 90.0\% & 72.0\% & 95.2\%  \\ \hline
LLaVA-v1.5-7B-XTuner  & 79.8\% & 100.0\% & 68.9\% & 25.0\% & 87.0\% & 99.5\%  & 99.0\%  & 67.0\% & 37.5\% & 99.0\%  \\ \hline
LLaVA-v1.5-13B-XTuner & 85.9\% & 99.9\%  & 70.4\% & 23.6\% & 92.5\% & 100.0\% & 100.0\% & 97.5\% & 86.5\% & 100.0\% \\ \hline
InternVL-Chat-V1-2    & 67.6\% & 87.3\%  & 57.9\% & 18.8\% & 78.0\% & 98.0\%  & 99.5\%  & 86.5\% & 55.0\% & 94.8\%  \\ \hline
LLaVA-NEXT-13B        & 88.9\% & 100.0\% & 80.3\% & 24.3\% & 93.5\% & 100.0\% & 100.0\% & 99.5\% & 96.0\% & 85.3\%  \\ \hline
GPT4o                 & 86.2\% & 96.6\%  & 72.7\% & 22.1\% & 92.5\% & 94.0\%  & 91.0\%  & 45.5\% & 24.5\% & 95.0\%  \\ \hline
mPLUG-Owl2            & 88.7\% & 99.9\%  & 62.7\% & 28.8\% & 94.5\% & 100.0\% & 99.5\%  & 98.5\% & 91.0\% & 99.7\%  \\ \hline
LLaVA-v1.5-7B         & 49.2\% & 100.0\% & 48.7\% & 36.0\% & 58.5\% & 100.0\% & 100.0\% & 97.5\% & 87.5\% & 95.3\%  \\ \hline
LLaVA-v1.5-13B        & 53.5\% & 99.0\%  & 42.9\% & 37.6\% & 63.0\% & 100.0\% & 100.0\% & 91.5\% & 59.0\% & 78.2\%  \\ \hline
Yi-VL-34B             & 59.3\% & 77.1\%  & 24.6\% & 23.8\% & 82.5\% & 84.5\%  & 56.5\%  & 35.5\% & 19.5\% & 65.3\%  \\ \hline
CogVLM-Chat           & 47.4\% & 52.8\%  & 51.8\% & 25.9\% & 45.0\% & 83.5\%  & 78.5\%  & 40.0\% & 38.0\% & 61.8\%  \\ \hline
Gemini-1.5-Pro        & 54.0\% & 33.3\%  & 45.0\% & 14.8\% & 59.0\% & 26.0\%  & 76.5\%  & 17.5\% & 12.5\% & 69.3\%  \\ \hline
XComposer2            & 44.2\% & 50.8\%  & 31.7\% & 10.0\% & 55.0\% & 94.0\%  & 90.0\%  & 38.5\% & 35.0\% & 34.3\%  \\ \hline
LLaVA-InternLM2-7B    & 22.4\% & 73.2\%  & 20.7\% & 31.5\% & 28.5\% & 95.0\%  & 99.5\%  & 73.5\% & 31.5\% & 41.2\%  \\ \hline
VisualGLM-6B          & 32.9\% & 49.1\%  & 56.9\% & 24.1\% & 49.0\% & 55.5\%  & 57.0\%  & 27.5\% & 21.5\% & 53.8\%  \\ \hline
LLaVA-NEXT-7B         & 42.2\% & 58.3\%  & 40.3\% & 24.5\% & 67.5\% & 100.0\% & 100.0\% & 97.0\% & 91.5\% & 36.0\%  \\ \hline
LLaVA-InternLM-7B     & 29.4\% & 39.1\%  & 28.9\% & 29.3\% & 31.0\% & 99.0\%  & 100.0\% & 64.0\% & 42.5\% & 47.5\%  \\ \hline
ShareGPT4V-7B         & 13.9\% & 57.3\%  & 17.3\% & 47.9\% & 24.0\% & 99.0\%  & 100.0\% & 87.0\% & 55.5\% & 27.2\%  \\ \hline
InternVL-Chat-V1-5    & 15.9\% & 0.5\%   & 14.1\% & 4.3\%  & 22.0\% & 96.5\%  & 97.0\%  & 32.5\% & 24.0\% & 29.7\%  \\ \hline
DeepSeek-VL-7B        & 29.4\% & 16.0\%  & 17.2\% & 24.4\% & 45.0\% & 97.5\%  & 99.0\%  & 48.5\% & 34.5\% & 45.0\%  \\ \hline
Yi-VL-6B              & 32.4\% & 2.5\%   & 6.3\%  & 23.0\% & 60.5\% & 83.0\%  & 70.5\%  & 40.5\% & 45.0\% & 70.0\%  \\ \hline
InstructBLIP-13B      & 22.5\% & 73.3\%  & 17.2\% & 25.0\% & 30.5\% & 58.5\%  & 57.5\%  & 42.5\% & 41.0\% & 33.7\%  \\ \hline
Qwen-VL-Chat          & 26.7\% & 36.1\%  & 13.5\% & 23.5\% & 43.0\% & 50.5\%  & 54.5\%  & 23.5\% & 28.5\% & 28.7\%  \\ \hline
Claude3V-Sonnet       & 47.9\% & 19.8\%  & 6.0\%  & 13.3\% & 59.5\% & 55.0\%  & 37.0\%  & 4.0\%  & 2.0\%  & 51.5\%  \\ \hline
Monkey-Chat           & 12.2\% & 15.3\%  & 7.6\%  & 23.6\% & 27.0\% & 49.5\%  & 50.0\%  & 19.5\% & 19.0\% & 23.3\%  \\ \hline
\end{tabular}
}
\caption{Detail results of $25$ LVLMs on $112$ forgery detetion types (part 1).}
\label{tab:detail results part 1}
\end{table*}

\begin{table*}[t]
\resizebox{1.0\textwidth}{!}{%
\begin{tabular}{c|l|l|l|l|l|l|l|l|l|l}
\hline
\textbf{Model} &
  \multicolumn{1}{c|}{\textbf{HS-RGB-BC-FE-ED}} &
  \multicolumn{1}{c|}{\textbf{HS-RGB-SLS-FE-ED}} &
  \multicolumn{1}{c|}{\textbf{HS-RGB\&TXT-BC-FE-ED}} &
  \multicolumn{1}{c|}{\textbf{HS-RGB\&TXT-SLD-FE-ED}} &
  \multicolumn{1}{c|}{\textbf{HS-RGB\&TXT-BC-FE-GAN}} &
  \multicolumn{1}{c|}{\textbf{HS-RGB\&TXT-SLD-FE-GAN}} &
  \multicolumn{1}{c|}{\textbf{HS-RGB-BC-FE-PRO}} &
  \multicolumn{1}{c|}{\textbf{HS-RGB-SLS-FE-PRO}} &
  \multicolumn{1}{c|}{\textbf{HS-RGB-BC-FE\&FT-ED\&GR}} &
  \multicolumn{1}{c}{\textbf{HS-RGB-SLS-FE\&FT-ED\&GR}} \\ \hline
LLaVA-NEXT-34B        & 99.1\%  & 23.3\% & 73.0\% & 19.5\% & 79.0\% & 18.5\% & 92.5\% & 28.5\% & 100.0\% & 23.5\% \\ \hline
LLaVA-v1.5-7B-XTuner  & 97.8\%  & 23.6\% & 70.0\% & 15.5\% & 72.0\% & 15.5\% & 95.0\% & 21.0\% & 96.5\%  & 23.5\% \\ \hline
LLaVA-v1.5-13B-XTuner & 100.0\% & 24.5\% & 89.0\% & 0.5\%  & 93.5\% & 2.0\%  & 99.5\% & 22.5\% & 100.0\% & 20.5\% \\ \hline
InternVL-Chat-V1-2    & 96.5\%  & 14.5\% & 54.5\% & 36.0\% & 59.5\% & 33.5\% & 85.0\% & 30.5\% & 97.0\%  & 6.5\%  \\ \hline
LLaVA-NEXT-13B        & 99.1\%  & 20.4\% & 84.5\% & 1.0\%  & 91.5\% & 1.5\%  & 88.5\% & 27.5\% & 100.0\% & 26.0\% \\ \hline
GPT4o                 & 87.6\%  & 24.0\% & 11.5\% & 5.5\%  & 15.5\% & 2.5\%  & 66.0\% & 24.5\% & 63.5\%  & 17.0\% \\ \hline
mPLUG-Owl2            & 92.7\%  & 24.4\% & 72.5\% & 29.5\% & 75.5\% & 25.5\% & 95.5\% & 31.0\% & 86.5\%  & 21.5\% \\ \hline
LLaVA-v1.5-7B         & 98.7\%  & 25.5\% & 95.5\% & 10.5\% & 95.5\% & 7.0\%  & 88.5\% & 21.5\% & 98.5\%  & 24.0\% \\ \hline
LLaVA-v1.5-13B        & 92.1\%  & 23.0\% & 59.0\% & 2.0\%  & 66.0\% & 3.0\%  & 61.5\% & 27.0\% & 93.5\%  & 27.0\% \\ \hline
Yi-VL-34B             & 70.8\%  & 24.5\% & 13.5\% & 26.5\% & 14.0\% & 21.5\% & 39.5\% & 27.5\% & 90.5\%  & 24.5\% \\ \hline
CogVLM-Chat           & 74.4\%  & 23.8\% & 75.0\% & 18.5\% & 74.5\% & 24.5\% & 56.5\% & 28.0\% & 57.0\%  & 23.5\% \\ \hline
Gemini-1.5-Pro        & 79.5\%  & 29.8\% & 27.5\% & 32.0\% & 24.5\% & 36.0\% & 49.5\% & 30.0\% & 60.5\%  & 22.0\% \\ \hline
XComposer2            & 66.4\%  & 12.3\% & 23.5\% & 26.5\% & 31.0\% & 26.0\% & 22.5\% & 20.0\% & 74.0\%  & 11.0\% \\ \hline
LLaVA-InternLM2-7B    & 62.8\%  & 22.0\% & 19.0\% & 13.5\% & 28.5\% & 10.0\% & 16.5\% & 27.0\% & 68.5\%  & 23.5\% \\ \hline
VisualGLM-6B          & 53.3\%  & 24.4\% & 21.0\% & 17.0\% & 19.0\% & 25.0\% & 91.0\% & 24.0\% & 66.5\%  & 22.5\% \\ \hline
LLaVA-NEXT-7B         & 66.3\%  & 23.1\% & 94.5\% & 6.0\%  & 94.5\% & 5.5\%  & 14.5\% & 25.0\% & 88.0\%  & 20.0\% \\ \hline
LLaVA-InternLM-7B     & 61.1\%  & 27.4\% & 28.0\% & 43.0\% & 25.5\% & 46.5\% & 29.5\% & 27.5\% & 60.5\%  & 22.5\% \\ \hline
ShareGPT4V-7B         & 58.7\%  & 23.9\% & 96.5\% & 15.5\% & 96.5\% & 13.0\% & 10.5\% & 25.5\% & 70.0\%  & 24.0\% \\ \hline
InternVL-Chat-V1-5    & 57.9\%  & 5.0\%  & 28.5\% & 31.0\% & 27.5\% & 33.0\% & 19.0\% & 28.5\% & 74.5\%  & 1.5\%  \\ \hline
DeepSeek-VL-7B        & 62.6\%  & 19.1\% & 8.0\%  & 29.5\% & 9.0\%  & 32.0\% & 20.5\% & 26.5\% & 64.5\%  & 15.5\% \\ \hline
Yi-VL-6B              & 68.2\%  & 24.6\% & 21.5\% & 25.5\% & 27.0\% & 21.0\% & 51.5\% & 28.5\% & 60.0\%  & 29.5\% \\ \hline
InstructBLIP-13B      & 55.1\%  & 25.5\% & 2.5\%  & 14.0\% & 3.0\%  & 21.5\% & 29.5\% & 21.5\% & 45.5\%  & 25.0\% \\ \hline
Qwen-VL-Chat          & 34.1\%  & 25.0\% & 28.5\% & 22.0\% & 38.5\% & 29.0\% & 11.5\% & 27.5\% & 36.0\%  & 25.5\% \\ \hline
Claude3V-Sonnet       & 35.7\%  & 23.5\% & 21.0\% & 16.0\% & 18.0\% & 17.5\% & 22.0\% & 29.0\% & 30.5\%  & 21.0\% \\ \hline
Monkey-Chat           & 26.9\%  & 23.5\% & 6.5\%  & 26.5\% & 11.0\% & 30.0\% & 13.0\% & 30.5\% & 30.0\%  & 22.5\% \\ \hline
\end{tabular}
}
\caption{Detail results of $25$ LVLMs on $112$ forgery detetion types (part 2).}
\label{tab:detail results part 2}
\end{table*}

\begin{table*}[t]
\resizebox{1.0\textwidth}{!}{%
\begin{tabular}{c|l|l|l|l|l|l|l|l}
\hline
\textbf{Model} &
  \multicolumn{1}{c|}{\textbf{HS-RGB\&TXT-BC-FE\&TAM-ED\&TR}} &
  \multicolumn{1}{c|}{\textbf{HS-RGB\&TXT-SLD-FE\&TAM-ED\&TR}} &
  \multicolumn{1}{c|}{\textbf{HS-RGB\&TXT-BC-FE\&TAM-GAN\&TR}} &
  \multicolumn{1}{c|}{\textbf{HS-RGB\&TXT-SLD-FE\&TAM-GAN\&TR}} &
  \multicolumn{1}{c|}{\textbf{HS-RGB\&TXT-BC-FE\&TS-ED\&RT}} &
  \multicolumn{1}{c|}{\textbf{HS-RGB\&TXT-SLD-FE\&TS-ED\&RT}} &
  \multicolumn{1}{c|}{\textbf{HS-RGB\&TXT-BC-FE\&TS-GAN\&RT}} &
  \multicolumn{1}{c}{\textbf{HS-RGB\&TXT-SLD-FE\&TS-GAN\&RT}} \\ \hline
LLaVA-NEXT-34B        & 98.0\%  & 20.5\% & 98.5\%  & 22.5\% & 97.5\%  & 40.0\% & 99.5\%  & 42.0\% \\ \hline
LLaVA-v1.5-7B-XTuner  & 91.5\%  & 17.0\% & 87.5\%  & 18.0\% & 87.0\%  & 33.0\% & 90.0\%  & 25.5\% \\ \hline
LLaVA-v1.5-13B-XTuner & 99.5\%  & 7.0\%  & 99.0\%  & 5.5\%  & 98.5\%  & 13.0\% & 99.5\%  & 12.5\% \\ \hline
InternVL-Chat-V1-2    & 90.0\%  & 22.0\% & 95.0\%  & 18.0\% & 92.5\%  & 47.0\% & 96.0\%  & 52.0\% \\ \hline
LLaVA-NEXT-13B        & 99.5\%  & 4.5\%  & 98.0\%  & 4.0\%  & 98.0\%  & 19.0\% & 100.0\% & 15.0\% \\ \hline
GPT4o                 & 40.0\%  & 34.5\% & 50.5\%  & 32.0\% & 72.0\%  & 34.5\% & 77.0\%  & 31.0\% \\ \hline
mPLUG-Owl2            & 94.0\%  & 24.5\% & 95.5\%  & 23.0\% & 95.5\%  & 55.0\% & 95.5\%  & 66.0\% \\ \hline
LLaVA-v1.5-7B         & 99.5\%  & 16.0\% & 100.0\% & 16.5\% & 100.0\% & 24.5\% & 100.0\% & 25.0\% \\ \hline
LLaVA-v1.5-13B        & 89.0\%  & 9.0\%  & 88.5\%  & 10.5\% & 92.5\%  & 30.0\% & 94.0\%  & 23.0\% \\ \hline
Yi-VL-34B             & 63.0\%  & 33.5\% & 64.5\%  & 33.5\% & 29.5\%  & 42.0\% & 31.0\%  & 43.0\% \\ \hline
CogVLM-Chat           & 94.0\%  & 26.5\% & 95.0\%  & 24.5\% & 94.5\%  & 32.0\% & 95.5\%  & 28.5\% \\ \hline
Gemini-1.5-Pro        & 64.5\%  & 34.0\% & 66.0\%  & 31.5\% & 82.5\%  & 27.0\% & 85.0\%  & 22.0\% \\ \hline
XComposer2            & 56.0\%  & 48.0\% & 61.0\%  & 52.0\% & 74.0\%  & 48.5\% & 75.0\%  & 51.0\% \\ \hline
LLaVA-InternLM2-7B    & 57.0\%  & 32.0\% & 59.0\%  & 34.5\% & 52.5\%  & 52.5\% & 61.0\%  & 53.5\% \\ \hline
VisualGLM-6B          & 29.0\%  & 25.0\% & 28.0\%  & 24.0\% & 27.0\%  & 31.0\% & 27.5\%  & 27.5\% \\ \hline
LLaVA-NEXT-7B         & 100.0\% & 18.5\% & 100.0\% & 18.0\% & 98.5\%  & 44.0\% & 100.0\% & 43.0\% \\ \hline
LLaVA-InternLM-7B     & 58.5\%  & 52.5\% & 58.0\%  & 47.5\% & 50.5\%  & 38.0\% & 54.5\%  & 45.5\% \\ \hline
ShareGPT4V-7B         & 99.5\%  & 26.5\% & 100.0\% & 26.0\% & 99.5\%  & 52.0\% & 98.5\%  & 48.0\% \\ \hline
InternVL-Chat-V1-5    & 67.5\%  & 17.5\% & 62.5\%  & 22.0\% & 80.0\%  & 27.0\% & 81.0\%  & 27.0\% \\ \hline
DeepSeek-VL-7B        & 17.5\%  & 13.5\% & 17.5\%  & 17.0\% & 13.0\%  & 20.0\% & 20.0\%  & 17.0\% \\ \hline
Yi-VL-6B              & 81.5\%  & 33.5\% & 75.0\%  & 30.0\% & 56.0\%  & 40.0\% & 54.0\%  & 45.5\% \\ \hline
InstructBLIP-13B      & 3.5\%   & 22.5\% & 3.0\%   & 23.5\% & 1.0\%   & 23.0\% & 4.5\%   & 22.0\% \\ \hline
Qwen-VL-Chat          & 55.5\%  & 21.0\% & 55.0\%  & 21.5\% & 55.0\%  & 32.5\% & 58.0\%  & 25.5\% \\ \hline
Claude3V-Sonnet       & 49.0\%  & 15.5\% & 51.5\%  & 10.0\% & 59.5\%  & 15.0\% & 65.0\%  & 15.0\% \\ \hline
Monkey-Chat           & 20.5\%  & 21.0\% & 18.5\%  & 22.0\% & 13.0\%  & 27.0\% & 17.0\%  & 23.0\% \\ \hline
\end{tabular}
}
\caption{Detail results of $25$ LVLMs on $112$ forgery detetion types (part 3).}
\label{tab:detail results part 3}
\end{table*}

\begin{table*}[t]
\resizebox{1.0\textwidth}{!}{%
\begin{tabular}{c|l|l|l|l|l|l|l|l|l|l}
\hline
\textbf{Model} &
  \multicolumn{1}{c|}{\textbf{HS-VID-BC-FR-RNN}} &
  \multicolumn{1}{c|}{\textbf{HS-VID-TL-FR-RNN}} &
  \multicolumn{1}{c|}{\textbf{HS-RGB-BC-FR-RNN}} &
  \multicolumn{1}{c|}{\textbf{HS-RGB-SLS-FR-RNN}} &
  \multicolumn{1}{c|}{\textbf{HS-VID-BC-FR-ED}} &
  \multicolumn{1}{c|}{\textbf{HS-RGB-BC-FR-ED}} &
  \multicolumn{1}{c|}{\textbf{HS-RGB-SLS-FR-ED}} &
  \multicolumn{1}{c|}{\textbf{HS-VID-BC-FR-GR}} &
  \multicolumn{1}{c|}{\textbf{HS-VID-SLS-FR-GR}} &
  \multicolumn{1}{c}{\textbf{HS-RGB-BC-FR-GR}} \\ \hline
LLaVA-NEXT-34B        & 100.0\% & 14.8\% & 99.3\%  & 26.3\% & 100.0\% & 90.0\%  & 21.0\% & 100.0\% & 25.0\% & 77.0\%  \\ \hline
LLaVA-v1.5-7B-XTuner  & 99.8\%  & 22.3\% & 94.3\%  & 21.3\% & 99.3\%  & 85.0\%  & 28.5\% & 100.0\% & 20.0\% & 75.5\%  \\ \hline
LLaVA-v1.5-13B-XTuner & 100.0\% & 25.5\% & 100.0\% & 23.8\% & 100.0\% & 100.0\% & 34.0\% & 100.0\% & 21.4\% & 100.0\% \\ \hline
InternVL-Chat-V1-2    & 99.8\%  & 23.3\% & 95.0\%  & 11.3\% & 100.0\% & 90.0\%  & 10.0\% & 100.0\% & 17.1\% & 85.0\%  \\ \hline
LLaVA-NEXT-13B        & 100.0\% & 14.0\% & 99.5\%  & 23.3\% & 100.0\% & 91.3\%  & 22.5\% & 100.0\% & 14.3\% & 88.5\%  \\ \hline
GPT4o                 & 81.3\%  & 22.5\% & 66.5\%  & 28.8\% & 66.4\%  & 39.0\%  & 27.5\% & 70.0\%  & 25.0\% & 7.0\%   \\ \hline
mPLUG-Owl2            & 99.8\%  & 27.8\% & 82.0\%  & 24.3\% & 100.0\% & 72.8\%  & 28.0\% & 100.0\% & 27.1\% & 67.5\%  \\ \hline
LLaVA-v1.5-7B         & 99.8\%  & 18.0\% & 96.8\%  & 20.8\% & 100.0\% & 89.0\%  & 29.0\% & 100.0\% & 23.6\% & 84.5\%  \\ \hline
LLaVA-v1.5-13B        & 98.8\%  & 23.0\% & 92.8\%  & 24.8\% & 100.0\% & 74.0\%  & 29.5\% & 99.3\%  & 20.7\% & 66.5\%  \\ \hline
Yi-VL-34B             & 92.8\%  & 26.3\% & 84.8\%  & 27.8\% & 94.3\%  & 83.5\%  & 19.0\% & 95.7\%  & 27.1\% & 81.5\%  \\ \hline
CogVLM-Chat           & 58.0\%  & 21.8\% & 56.5\%  & 27.3\% & 77.1\%  & 56.0\%  & 18.0\% & 69.3\%  & 21.4\% & 51.5\%  \\ \hline
Gemini-1.5-Pro        & 48.3\%  & 49.8\% & 54.3\%  & 29.3\% & 7.1\%   & 37.5\%  & 33.0\% & 18.6\%  & 44.3\% & 8.5\%   \\ \hline
XComposer2            & 68.3\%  & 4.3\%  & 73.0\%  & 12.0\% & 19.3\%  & 39.8\%  & 13.0\% & 22.9\%  & 49.3\% & 6.5\%   \\ \hline
LLaVA-InternLM2-7B    & 96.8\%  & 17.5\% & 70.3\%  & 21.0\% & 99.3\%  & 45.3\%  & 28.5\% & 98.6\%  & 21.4\% & 17.0\%  \\ \hline
VisualGLM-6B          & 55.8\%  & 33.0\% & 73.0\%  & 26.0\% & 51.4\%  & 69.0\%  & 22.5\% & 60.7\%  & 22.1\% & 78.0\%  \\ \hline
LLaVA-NEXT-7B         & 99.0\%  & 23.8\% & 86.0\%  & 22.5\% & 100.0\% & 54.3\%  & 17.5\% & 99.3\%  & 21.4\% & 41.5\%  \\ \hline
LLaVA-InternLM-7B     & 48.0\%  & 17.8\% & 65.0\%  & 28.0\% & 38.6\%  & 50.0\%  & 25.0\% & 45.0\%  & 21.4\% & 34.0\%  \\ \hline
ShareGPT4V-7B         & 84.5\%  & 22.3\% & 66.8\%  & 21.0\% & 86.4\%  & 41.8\%  & 28.0\% & 85.0\%  & 21.4\% & 18.5\%  \\ \hline
InternVL-Chat-V1-5    & 96.3\%  & 6.3\%  & 70.8\%  & 5.0\%  & 91.4\%  & 44.5\%  & 5.5\%  & 97.9\%  & 5.0\%  & 15.0\%  \\ \hline
DeepSeek-VL-7B        & 59.0\%  & 8.5\%  & 61.3\%  & 19.5\% & 65.0\%  & 37.5\%  & 24.0\% & 62.9\%  & 13.6\% & 7.0\%   \\ \hline
Yi-VL-6B              & 86.5\%  & 25.3\% & 58.3\%  & 28.0\% & 70.7\%  & 40.3\%  & 22.0\% & 64.3\%  & 22.9\% & 25.0\%  \\ \hline
InstructBLIP-13B      & 85.0\%  & 32.5\% & 45.3\%  & 21.5\% & 88.6\%  & 25.8\%  & 28.0\% & 77.9\%  & 30.7\% & 10.5\%  \\ \hline
Qwen-VL-Chat          & 55.0\%  & 22.3\% & 37.0\%  & 25.3\% & 47.1\%  & 22.8\%  & 22.5\% & 52.1\%  & 23.6\% & 5.0\%   \\ \hline
Claude3V-Sonnet       & 36.3\%  & 41.5\% & 28.5\%  & 18.5\% & 13.6\%  & 15.3\%  & 23.5\% & 15.0\%  & 53.6\% & 2.5\%   \\ \hline
Monkey-Chat           & 9.8\%   & 24.8\% & 30.0\%  & 26.5\% & 9.3\%   & 19.3\%  & 19.0\% & 7.1\%   & 22.1\% & 4.5\%   \\ \hline
\end{tabular}
}
\caption{Detail results of $25$ LVLMs on $112$ forgery detetion types (part 4).}
\label{tab:detail results part 4}
\end{table*}

\begin{table*}[t]
\resizebox{1.0\textwidth}{!}{%
\begin{tabular}{c|l|l|l|l|l|l|l|l|l|l}
\hline
\textbf{Model} &
  \multicolumn{1}{c|}{\textbf{HS-RGB-SLS-FR-GR}} &
  \multicolumn{1}{c|}{\textbf{HS-VID-BC-FSM-ED\&RNN\&GR}} &
  \multicolumn{1}{c|}{\textbf{HS-VID-SLS-FSM-ED\&RNN\&GR}} &
  \multicolumn{1}{c|}{\textbf{HS-RGB-BC-FSM-ED\&RNN\&GR}} &
  \multicolumn{1}{c|}{\textbf{HS-RGB-SLS-FSM-ED\&RNN\&GR}} &
  \multicolumn{1}{c|}{\textbf{HS-RGB-SLD-FSM-GAN}} &
  \multicolumn{1}{c|}{\textbf{HS-RGB-SLS-FSM-GAN}} &
  \multicolumn{1}{c|}{\textbf{HS-VID-BC-FSS-RNN}} &
  \multicolumn{1}{c|}{\textbf{HS-VID-TL-FSS-RNN}} &
  \multicolumn{1}{c}{\textbf{HS-RGB-BC-FSS-RNN}} \\ \hline
LLaVA-NEXT-34B        & 20.0\% & 100.0\% & 26.0\% & 89.5\%  & 23.0\% & 45.5\% & 20.0\% & 100.0\% & 4.5\%  & 98.0\%  \\ \hline
LLaVA-v1.5-7B-XTuner  & 25.0\% & 100.0\% & 21.5\% & 77.0\%  & 28.0\% & 15.5\% & 21.0\% & 100.0\% & 47.5\% & 93.0\%  \\ \hline
LLaVA-v1.5-13B-XTuner & 21.5\% & 100.0\% & 26.5\% & 100.0\% & 27.0\% & 37.5\% & 26.0\% & 100.0\% & 33.5\% & 100.0\% \\ \hline
InternVL-Chat-V1-2    & 10.5\% & 100.0\% & 23.5\% & 78.5\%  & 17.0\% & 45.5\% & 17.0\% & 100.0\% & 23.5\% & 93.0\%  \\ \hline
LLaVA-NEXT-13B        & 21.5\% & 100.0\% & 18.0\% & 93.0\%  & 19.5\% & 24.0\% & 20.5\% & 100.0\% & 18.5\% & 99.0\%  \\ \hline
GPT4o                 & 21.5\% & 65.5\%  & 15.0\% & 7.0\%   & 21.0\% & 73.5\% & 19.5\% & 86.5\%  & 19.5\% & 55.0\%  \\ \hline
mPLUG-Owl2            & 25.0\% & 100.0\% & 23.5\% & 34.0\%  & 27.0\% & 39.5\% & 24.5\% & 100.0\% & 43.5\% & 74.5\%  \\ \hline
LLaVA-v1.5-7B         & 25.5\% & 100.0\% & 28.5\% & 98.0\%  & 25.5\% & 21.0\% & 23.5\% & 100.0\% & 26.0\% & 95.5\%  \\ \hline
LLaVA-v1.5-13B        & 25.5\% & 98.5\%  & 31.5\% & 76.5\%  & 26.5\% & 28.5\% & 21.5\% & 99.5\%  & 45.5\% & 90.5\%  \\ \hline
Yi-VL-34B             & 23.0\% & 97.5\%  & 18.0\% & 89.0\%  & 26.5\% & 39.0\% & 26.0\% & 99.5\%  & 41.5\% & 84.5\%  \\ \hline
CogVLM-Chat           & 23.0\% & 75.5\%  & 31.0\% & 50.0\%  & 26.5\% & 22.5\% & 27.0\% & 72.0\%  & 29.0\% & 51.5\%  \\ \hline
Gemini-1.5-Pro        & 23.0\% & 16.5\%  & 30.0\% & 4.0\%   & 25.0\% & 37.5\% & 28.0\% & 67.5\%  & 51.5\% & 48.0\%  \\ \hline
XComposer2            & 8.0\%  & 30.0\%  & 35.5\% & 2.0\%   & 9.5\%  & 51.5\% & 6.0\%  & 72.0\%  & 2.0\%  & 69.5\%  \\ \hline
LLaVA-InternLM2-7B    & 24.0\% & 100.0\% & 31.5\% & 29.5\%  & 20.5\% & 43.0\% & 19.0\% & 98.0\%  & 6.5\%  & 61.0\%  \\ \hline
VisualGLM-6B          & 24.0\% & 51.5\%  & 32.0\% & 38.5\%  & 30.0\% & 43.5\% & 28.0\% & 58.0\%  & 25.0\% & 68.0\%  \\ \hline
LLaVA-NEXT-7B         & 19.5\% & 100.0\% & 29.5\% & 42.5\%  & 16.5\% & 17.0\% & 24.0\% & 99.5\%  & 58.0\% & 80.0\%  \\ \hline
LLaVA-InternLM-7B     & 24.5\% & 48.0\%  & 31.5\% & 20.5\%  & 30.5\% & 36.0\% & 26.5\% & 51.5\%  & 19.0\% & 65.0\%  \\ \hline
ShareGPT4V-7B         & 27.0\% & 79.5\%  & 31.5\% & 32.5\%  & 27.5\% & 23.0\% & 25.5\% & 91.0\%  & 48.5\% & 63.0\%  \\ \hline
InternVL-Chat-V1-5    & 5.5\%  & 96.0\%  & 17.0\% & 9.5\%   & 5.5\%  & 48.5\% & 7.0\%  & 96.0\%  & 1.0\%  & 66.5\%  \\ \hline
DeepSeek-VL-7B        & 19.0\% & 55.5\%  & 17.5\% & 5.0\%   & 20.5\% & 55.0\% & 17.5\% & 71.5\%  & 6.0\%  & 57.5\%  \\ \hline
Yi-VL-6B              & 24.0\% & 82.0\%  & 20.0\% & 25.0\%  & 30.0\% & 46.5\% & 26.5\% & 93.5\%  & 41.0\% & 60.0\%  \\ \hline
InstructBLIP-13B      & 26.0\% & 83.0\%  & 23.0\% & 4.5\%   & 27.0\% & 14.5\% & 25.5\% & 88.0\%  & 34.0\% & 38.5\%  \\ \hline
Qwen-VL-Chat          & 23.5\% & 62.5\%  & 28.0\% & 1.5\%   & 26.5\% & 27.5\% & 27.0\% & 59.0\%  & 28.0\% & 34.0\%  \\ \hline
Claude3V-Sonnet       & 22.0\% & 8.5\%   & 40.5\% & 0.5\%   & 26.0\% & 18.0\% & 22.0\% & 43.5\%  & 21.5\% & 22.5\%  \\ \hline
Monkey-Chat           & 23.5\% & 3.0\%   & 23.5\% & 0.5\%   & 25.5\% & 34.0\% & 27.0\% & 16.5\%  & 30.0\% & 28.5\%  \\ \hline
\end{tabular}
}
\caption{Detail results of $25$ LVLMs on $112$ forgery detetion types (part 5).}
\label{tab:detail results part 5}
\end{table*}

\begin{table*}[t]
\resizebox{1.0\textwidth}{!}{%
\begin{tabular}{c|l|l|l|l|l|l|l|l|l|l}
\hline
\textbf{Model} &
  \multicolumn{1}{c|}{\textbf{HS-RGB-SLS-FSS-RNN}} &
  \multicolumn{1}{c|}{\textbf{HS-RGB-BC-FSS-DF}} &
  \multicolumn{1}{c|}{\textbf{HS-VID-BC-FSS-ED}} &
  \multicolumn{1}{c|}{\textbf{HS-VID-SLS-FSS-ED}} &
  \multicolumn{1}{c|}{\textbf{HS-VID-TL-FSS-ED}} &
  \multicolumn{1}{c|}{\textbf{HS-RGB-BC-FSS-ED}} &
  \multicolumn{1}{c|}{\textbf{HS-RGB-SLS-FSS-ED}} &
  \multicolumn{1}{c|}{\textbf{HS-RGB\&TXT-BC-FSS-ED}} &
  \multicolumn{1}{c|}{\textbf{HS-RGB\&TXT-SLD-FSS-ED}} &
  \multicolumn{1}{c}{\textbf{HS-VID-BC-FSS-GR}} \\ \hline
LLaVA-NEXT-34B        & 24.5\% & 95.5\%  & 100.0\% & 18.4\% & 21.0\% & 75.4\%  & 22.7\% & 78.5\% & 15.0\% & 100.0\% \\ \hline
LLaVA-v1.5-7B-XTuner  & 27.5\% & 99.0\%  & 99.9\%  & 22.7\% & 12.8\% & 83.4\%  & 25.4\% & 69.8\% & 16.5\% & 100.0\% \\ \hline
LLaVA-v1.5-13B-XTuner & 25.5\% & 100.0\% & 100.0\% & 22.3\% & 20.8\% & 100.0\% & 23.7\% & 93.8\% & 0.3\%  & 100.0\% \\ \hline
InternVL-Chat-V1-2    & 8.0\%  & 96.3\%  & 100.0\% & 20.1\% & 24.5\% & 84.2\%  & 11.9\% & 57.8\% & 30.0\% & 100.0\% \\ \hline
LLaVA-NEXT-13B        & 22.0\% & 94.8\%  & 100.0\% & 13.9\% & 12.3\% & 83.6\%  & 25.6\% & 89.8\% & 0.0\%  & 100.0\% \\ \hline
GPT4o                 & 22.0\% & 96.0\%  & 74.1\%  & 22.7\% & 33.3\% & 34.9\%  & 22.7\% & 16.5\% & 3.8\%  & 74.3\%  \\ \hline
mPLUG-Owl2            & 25.0\% & 98.5\%  & 99.9\%  & 24.6\% & 18.3\% & 75.0\%  & 22.9\% & 72.3\% & 35.5\% & 100.0\% \\ \hline
LLaVA-v1.5-7B         & 28.0\% & 98.0\%  & 100.0\% & 23.0\% & 17.0\% & 88.7\%  & 25.6\% & 97.5\% & 10.3\% & 100.0\% \\ \hline
LLaVA-v1.5-13B        & 28.0\% & 81.0\%  & 99.9\%  & 25.2\% & 14.5\% & 71.1\%  & 25.4\% & 63.0\% & 2.0\%  & 99.3\%  \\ \hline
Yi-VL-34B             & 24.5\% & 65.0\%  & 97.7\%  & 23.9\% & 22.3\% & 59.2\%  & 22.9\% & 13.0\% & 26.8\% & 97.9\%  \\ \hline
CogVLM-Chat           & 24.5\% & 83.8\%  & 62.4\%  & 25.6\% & 20.8\% & 56.1\%  & 23.1\% & 76.5\% & 19.3\% & 74.3\%  \\ \hline
Gemini-1.5-Pro        & 28.5\% & 87.5\%  & 26.6\%  & 18.4\% & 41.0\% & 24.9\%  & 30.8\% & 31.5\% & 33.3\% & 23.6\%  \\ \hline
XComposer2            & 12.5\% & 38.8\%  & 39.1\%  & 34.6\% & 6.5\%  & 26.5\%  & 10.0\% & 27.8\% & 25.5\% & 22.1\%  \\ \hline
LLaVA-InternLM2-7B    & 20.0\% & 30.3\%  & 98.6\%  & 25.2\% & 20.5\% & 26.9\%  & 25.4\% & 22.3\% & 11.0\% & 98.6\%  \\ \hline
VisualGLM-6B          & 23.5\% & 18.3\%  & 53.6\%  & 23.9\% & 33.3\% & 77.6\%  & 29.4\% & 15.5\% & 20.0\% & 53.6\%  \\ \hline
LLaVA-NEXT-7B         & 25.0\% & 32.8\%  & 95.6\%  & 23.3\% & 9.5\%  & 41.4\%  & 18.9\% & 97.5\% & 6.5\%  & 100.0\% \\ \hline
LLaVA-InternLM-7B     & 30.5\% & 46.0\%  & 49.7\%  & 25.6\% & 13.3\% & 42.3\%  & 28.3\% & 23.5\% & 45.5\% & 49.3\%  \\ \hline
ShareGPT4V-7B         & 27.5\% & 31.8\%  & 87.5\%  & 24.9\% & 9.5\%  & 25.9\%  & 26.1\% & 96.5\% & 15.5\% & 83.6\%  \\ \hline
InternVL-Chat-V1-5    & 6.0\%  & 20.5\%  & 92.0\%  & 1.6\%  & 11.3\% & 26.6\%  & 4.7\%  & 24.5\% & 28.8\% & 95.0\%  \\ \hline
DeepSeek-VL-7B        & 20.0\% & 33.3\%  & 60.9\%  & 8.4\%  & 13.8\% & 22.8\%  & 16.9\% & 5.8\%  & 30.0\% & 62.9\%  \\ \hline
Yi-VL-6B              & 26.0\% & 74.0\%  & 88.2\%  & 23.3\% & 13.0\% & 28.1\%  & 25.6\% & 24.3\% & 28.5\% & 75.0\%  \\ \hline
InstructBLIP-13B      & 26.5\% & 54.5\%  & 86.3\%  & 27.5\% & 26.5\% & 22.5\%  & 24.1\% & 2.3\%  & 20.8\% & 83.6\%  \\ \hline
Qwen-VL-Chat          & 21.0\% & 22.0\%  & 56.7\%  & 20.1\% & 20.5\% & 12.1\%  & 23.6\% & 29.5\% & 26.8\% & 53.6\%  \\ \hline
Claude3V-Sonnet       & 16.0\% & 38.5\%  & 19.8\%  & 41.7\% & 39.8\% & 8.1\%   & 21.4\% & 19.5\% & 16.0\% & 20.0\%  \\ \hline
Monkey-Chat           & 24.0\% & 27.8\%  & 6.8\%   & 21.4\% & 17.0\% & 9.6\%   & 21.6\% & 6.0\%  & 30.0\% & 10.0\%  \\ \hline
\end{tabular}
}
\caption{Detail results of $25$ LVLMs on $112$ forgery detetion types (part 6).}
\label{tab:detail results part 6}\end{table*}

\begin{table*}[t]
\resizebox{1.0\textwidth}{!}{%
\begin{tabular}{c|l|l|l|l|l|l|l|l|l|l}
\hline
\textbf{Model} &
  \multicolumn{1}{c|}{\textbf{HS-VID-SLS-FSS-GR}} &
  \multicolumn{1}{c|}{\textbf{HS-RGB-BC-FSS-GR}} &
  \multicolumn{1}{c|}{\textbf{HS-RGB-SLS-FSS-GR}} &
  \multicolumn{1}{c|}{\textbf{HS-VID-BC-FSS-WILD}} &
  \multicolumn{1}{c|}{\textbf{HS-RGB-BC-FSS-WILD}} &
  \multicolumn{1}{c|}{\textbf{HS-VID-BC-FSS-VAE}} &
  \multicolumn{1}{c|}{\textbf{HS-RGB-BC-FSS-VAE}} &
  \multicolumn{1}{c|}{\textbf{HS-VID-BC-FSS\&FE-ED}} &
  \multicolumn{1}{c|}{\textbf{HS-VID-SLS-FSS\&FE-ED}} &
  \multicolumn{1}{c}{\textbf{HS-VID-TL-FSS\&FE-ED}} \\ \hline
LLaVA-NEXT-34B        & 15.7\% & 74.0\% & 25.5\% & 100.0\% & 94.0\%  & 100.0\% & 87.5\%  & 100.0\% & 21.0\% & 15.0\% \\ \hline
LLaVA-v1.5-7B-XTuner  & 27.1\% & 70.5\% & 23.0\% & 100.0\% & 78.0\%  & 100.0\% & 77.5\%  & 100.0\% & 22.0\% & 10.0\% \\ \hline
LLaVA-v1.5-13B-XTuner & 20.0\% & 99.5\% & 22.0\% & 100.0\% & 100.0\% & 100.0\% & 100.0\% & 100.0\% & 27.0\% & 23.5\% \\ \hline
InternVL-Chat-V1-2    & 21.4\% & 88.5\% & 16.0\% & 100.0\% & 85.3\%  & 100.0\% & 89.5\%  & 100.0\% & 22.0\% & 25.5\% \\ \hline
LLaVA-NEXT-13B        & 15.7\% & 90.5\% & 26.5\% & 100.0\% & 87.5\%  & 100.0\% & 94.5\%  & 100.0\% & 13.0\% & 11.5\% \\ \hline
GPT4o                 & 20.0\% & 24.5\% & 18.5\% & 48.8\%  & 3.3\%   & 76.0\%  & 19.5\%  & 84.5\%  & 17.0\% & 28.5\% \\ \hline
mPLUG-Owl2            & 20.7\% & 65.5\% & 23.5\% & 100.0\% & 47.0\%  & 100.0\% & 75.5\%  & 100.0\% & 27.0\% & 15.0\% \\ \hline
LLaVA-v1.5-7B         & 29.3\% & 80.0\% & 21.5\% & 100.0\% & 95.8\%  & 100.0\% & 86.5\%  & 99.0\%  & 22.0\% & 15.5\% \\ \hline
LLaVA-v1.5-13B        & 17.9\% & 62.5\% & 26.5\% & 99.8\%  & 77.0\%  & 100.0\% & 68.0\%  & 99.5\%  & 26.0\% & 15.0\% \\ \hline
Yi-VL-34B             & 20.0\% & 81.0\% & 25.5\% & 90.3\%  & 32.3\%  & 96.5\%  & 77.0\%  & 95.0\%  & 28.0\% & 16.5\% \\ \hline
CogVLM-Chat           & 20.0\% & 61.0\% & 25.5\% & 55.8\%  & 49.5\%  & 78.5\%  & 53.5\%  & 65.5\%  & 26.0\% & 18.5\% \\ \hline
Gemini-1.5-Pro        & 37.1\% & 19.5\% & 25.0\% & 18.5\%  & 4.5\%   & 40.5\%  & 15.5\%  & 56.0\%  & 19.0\% & 41.0\% \\ \hline
XComposer2            & 35.7\% & 14.5\% & 13.0\% & 22.8\%  & 2.8\%   & 21.5\%  & 8.5\%   & 66.5\%  & 50.0\% & 5.0\%  \\ \hline
LLaVA-InternLM2-7B    & 20.0\% & 18.0\% & 23.5\% & 87.0\%  & 2.0\%   & 94.5\%  & 19.0\%  & 99.0\%  & 26.0\% & 19.0\% \\ \hline
VisualGLM-6B          & 28.6\% & 81.5\% & 25.5\% & 57.8\%  & 58.3\%  & 55.5\%  & 68.0\%  & 53.0\%  & 24.0\% & 34.0\% \\ \hline
LLaVA-NEXT-7B         & 16.4\% & 42.5\% & 19.5\% & 100.0\% & 31.8\%  & 100.0\% & 40.5\%  & 97.5\%  & 24.0\% & 5.5\%  \\ \hline
LLaVA-InternLM-7B     & 20.0\% & 33.0\% & 30.0\% & 45.0\%  & 24.0\%  & 41.5\%  & 47.5\%  & 49.5\%  & 26.0\% & 13.5\% \\ \hline
ShareGPT4V-7B         & 20.0\% & 19.5\% & 19.0\% & 89.3\%  & 41.8\%  & 87.5\%  & 23.0\%  & 87.0\%  & 26.0\% & 10.0\% \\ \hline
InternVL-Chat-V1-5    & 3.6\%  & 13.5\% & 4.0\%  & 80.3\%  & 1.8\%   & 92.5\%  & 10.0\%  & 93.0\%  & 1.0\%  & 9.5\%  \\ \hline
DeepSeek-VL-7B        & 15.7\% & 8.5\%  & 27.0\% & 57.0\%  & 0.3\%   & 63.5\%  & 8.5\%   & 67.0\%  & 14.0\% & 6.0\%  \\ \hline
Yi-VL-6B              & 19.3\% & 25.5\% & 27.5\% & 97.8\%  & 23.3\%  & 79.0\%  & 23.5\%  & 89.5\%  & 26.0\% & 12.0\% \\ \hline
InstructBLIP-13B      & 25.7\% & 12.0\% & 23.5\% & 85.8\%  & 24.0\%  & 89.0\%  & 16.0\%  & 88.0\%  & 23.0\% & 24.0\% \\ \hline
Qwen-VL-Chat          & 23.6\% & 7.5\%  & 25.5\% & 40.3\%  & 4.3\%   & 55.0\%  & 4.5\%   & 54.0\%  & 22.0\% & 21.5\% \\ \hline
Claude3V-Sonnet       & 56.4\% & 3.0\%  & 24.5\% & 2.0\%   & 0.3\%   & 13.0\%  & 0.5\%   & 46.0\%  & 54.0\% & 43.5\% \\ \hline
Monkey-Chat           & 15.0\% & 4.0\%  & 26.0\% & 0.0\%   & 0.5\%   & 8.0\%   & 3.0\%   & 11.0\%  & 24.0\% & 12.0\% \\ \hline
\end{tabular}
}
\caption{Detail results of $25$ LVLMs on $112$ forgery detetion types (part 7).}
\label{tab:detail results part 7}\end{table*}

\begin{table*}[t]
\resizebox{1.0\textwidth}{!}{%
\begin{tabular}{c|l|l|l|l|l|l|l|l|l|l}
\hline
\textbf{Model} &
  \multicolumn{1}{c|}{\textbf{HS-RGB-BC-FSS\&FE-ED}} &
  \multicolumn{1}{c|}{\textbf{HS-RGB-SLS-FSS\&FE-ED}} &
  \multicolumn{1}{c|}{\textbf{HS-RGB\&TXT-BC-FSS\&TAM-ED\&TR}} &
  \multicolumn{1}{c|}{\textbf{HS-RGB\&TXT-SLD-FSS\&TAM-ED\&TR}} &
  \multicolumn{1}{c|}{\textbf{HS-RGB\&TXT-BC-FSS\&TS-ED\&RT}} &
  \multicolumn{1}{c|}{\textbf{HS-RGB\&TXT-SLD-FSS\&TS-ED\&RT}} &
  \multicolumn{1}{c|}{\textbf{HS-VID-BC-FT-GR}} &
  \multicolumn{1}{c|}{\textbf{HS-VID-SLS-FT-GR}} &
  \multicolumn{1}{c|}{\textbf{HS-VID-TL-FT-GR}} &
  \multicolumn{1}{c}{\textbf{HS-RGB-BC-FT-GR}} \\ \hline
LLaVA-NEXT-34B        & 100.0\% & 27.5\% & 99.0\%  & 23.0\% & 97.5\%  & 44.3\% & 100.0\% & 13.9\% & 22.3\% & 98.8\%  \\ \hline
LLaVA-v1.5-7B-XTuner  & 98.5\%  & 23.5\% & 93.0\%  & 23.0\% & 84.5\%  & 29.3\% & 100.0\% & 16.9\% & 14.7\% & 94.5\%  \\ \hline
LLaVA-v1.5-13B-XTuner & 100.0\% & 22.5\% & 100.0\% & 6.3\%  & 98.0\%  & 15.0\% & 100.0\% & 19.9\% & 23.0\% & 100.0\% \\ \hline
InternVL-Chat-V1-2    & 97.5\%  & 14.5\% & 93.0\%  & 22.8\% & 94.8\%  & 47.5\% & 100.0\% & 15.2\% & 28.3\% & 96.0\%  \\ \hline
LLaVA-NEXT-13B        & 100.0\% & 19.5\% & 99.8\%  & 7.5\%  & 98.8\%  & 12.0\% & 100.0\% & 13.4\% & 9.3\%  & 99.8\%  \\ \hline
GPT4o                 & 81.5\%  & 21.0\% & 55.3\%  & 43.0\% & 78.3\%  & 33.3\% & 85.0\%  & 26.0\% & 33.0\% & 70.3\%  \\ \hline
mPLUG-Owl2            & 87.0\%  & 25.0\% & 97.3\%  & 27.0\% & 94.3\%  & 53.5\% & 100.0\% & 22.9\% & 12.7\% & 83.0\%  \\ \hline
LLaVA-v1.5-7B         & 99.0\%  & 24.0\% & 100.0\% & 24.0\% & 100.0\% & 25.0\% & 100.0\% & 19.5\% & 16.0\% & 97.3\%  \\ \hline
LLaVA-v1.5-13B        & 94.5\%  & 26.0\% & 91.0\%  & 12.5\% & 91.8\%  & 34.0\% & 99.7\%  & 23.8\% & 9.7\%  & 93.0\%  \\ \hline
Yi-VL-34B             & 90.5\%  & 24.5\% & 66.5\%  & 34.8\% & 26.8\%  & 41.3\% & 97.0\%  & 23.8\% & 22.3\% & 88.3\%  \\ \hline
CogVLM-Chat           & 69.0\%  & 25.0\% & 97.3\%  & 22.3\% & 96.3\%  & 27.3\% & 66.3\%  & 23.4\% & 18.0\% & 63.5\%  \\ \hline
Gemini-1.5-Pro        & 69.0\%  & 25.5\% & 69.3\%  & 38.3\% & 83.8\%  & 27.3\% & 62.7\%  & 10.4\% & 47.3\% & 61.5\%  \\ \hline
XComposer2            & 83.5\%  & 15.5\% & 62.3\%  & 47.3\% & 76.0\%  & 48.8\% & 71.7\%  & 41.1\% & 13.3\% & 73.3\%  \\ \hline
LLaVA-InternLM2-7B    & 81.5\%  & 21.0\% & 63.0\%  & 34.3\% & 54.0\%  & 57.3\% & 99.7\%  & 23.8\% & 15.0\% & 73.5\%  \\ \hline
VisualGLM-6B          & 78.5\%  & 26.5\% & 26.3\%  & 29.5\% & 21.3\%  & 30.8\% & 55.3\%  & 20.8\% & 34.3\% & 73.0\%  \\ \hline
LLaVA-NEXT-7B         & 92.5\%  & 18.0\% & 99.8\%  & 24.3\% & 99.5\%  & 42.5\% & 100.0\% & 15.2\% & 7.3\%  & 86.0\%  \\ \hline
LLaVA-InternLM-7B     & 66.0\%  & 22.0\% & 61.5\%  & 47.8\% & 56.3\%  & 42.0\% & 50.3\%  & 23.8\% & 13.0\% & 64.8\%  \\ \hline
ShareGPT4V-7B         & 71.0\%  & 23.5\% & 100.0\% & 31.5\% & 100.0\% & 43.0\% & 92.0\%  & 23.8\% & 11.7\% & 68.8\%  \\ \hline
InternVL-Chat-V1-5    & 81.5\%  & 4.0\%  & 71.0\%  & 19.8\% & 78.8\%  & 31.3\% & 96.0\%  & 0.9\%  & 19.0\% & 73.8\%  \\ \hline
DeepSeek-VL-7B        & 77.5\%  & 20.0\% & 22.5\%  & 21.8\% & 15.5\%  & 20.0\% & 73.7\%  & 4.3\%  & 19.3\% & 65.5\%  \\ \hline
Yi-VL-6B              & 63.5\%  & 22.0\% & 83.5\%  & 31.3\% & 57.5\%  & 44.5\% & 91.3\%  & 22.5\% & 14.3\% & 60.5\%  \\ \hline
InstructBLIP-13B      & 47.5\%  & 23.5\% & 3.3\%   & 26.3\% & 3.5\%   & 27.8\% & 88.0\%  & 32.5\% & 24.3\% & 44.3\%  \\ \hline
Qwen-VL-Chat          & 45.0\%  & 25.5\% & 59.0\%  & 28.5\% & 55.8\%  & 31.3\% & 58.0\%  & 20.8\% & 20.3\% & 40.3\%  \\ \hline
Claude3V-Sonnet       & 35.5\%  & 25.0\% & 51.0\%  & 14.8\% & 66.3\%  & 16.3\% & 49.0\%  & 32.9\% & 52.7\% & 30.5\%  \\ \hline
Monkey-Chat           & 39.0\%  & 25.0\% & 21.8\%  & 29.0\% & 14.0\%  & 27.3\% & 13.3\%  & 19.5\% & 10.7\% & 34.8\%  \\ \hline
\end{tabular}
}
\caption{Detail results of $25$ LVLMs on $112$ forgery detetion types (part 8).}
\label{tab:detail results part 8}\end{table*}

\begin{table*}[t]
\resizebox{1.0\textwidth}{!}{%
\begin{tabular}{c|l|l|l|l|l|l|l|l|l|l}
\hline
\textbf{Model} &
  \multicolumn{1}{c|}{\textbf{HS-RGB-SLS-FT-GR}} &
  \multicolumn{1}{c|}{\textbf{HS-NIR-BC-ST-ED}} &
  \multicolumn{1}{c|}{\textbf{HS-RGB\&TXT-BC-TAM-TR}} &
  \multicolumn{1}{c|}{\textbf{HS-RGB\&TXT-SLD-TAM-TR}} &
  \multicolumn{1}{c|}{\textbf{HS-RGB\&TXT-BC-TS-RT}} &
  \multicolumn{1}{c|}{\textbf{HS-RGB\&TXT-SLD-TS-RT}} &
  \multicolumn{1}{c|}{\textbf{GS-RGB-BC-ES-AR}} &
  \multicolumn{1}{c|}{\textbf{GS-RGB-BC-ES-DF}} &
  \multicolumn{1}{c|}{\textbf{GS-RGB-BC-ES-GAN}} &
  \multicolumn{1}{c}{\textbf{GS-RGB-BC-ES-PRO}} \\ \hline
LLaVA-NEXT-34B        & 21.3\% & 100.0\% & 98.5\%  & 4.5\%  & 98.5\%  & 11.5\% & 96.0\% & 70.5\% & 84.8\% & 71.7\% \\ \hline
LLaVA-v1.5-7B-XTuner  & 28.0\% & 100.0\% & 88.5\%  & 4.0\%  & 81.5\%  & 26.0\% & 80.5\% & 64.7\% & 80.5\% & 69.3\% \\ \hline
LLaVA-v1.5-13B-XTuner & 26.5\% & 100.0\% & 99.0\%  & 0.0\%  & 99.0\%  & 0.0\%  & 95.5\% & 72.9\% & 82.9\% & 64.2\% \\ \hline
InternVL-Chat-V1-2    & 8.5\%  & 100.0\% & 91.5\%  & 8.5\%  & 91.5\%  & 29.0\% & 79.0\% & 54.1\% & 68.9\% & 56.0\% \\ \hline
LLaVA-NEXT-13B        & 19.8\% & 100.0\% & 99.5\%  & 0.5\%  & 98.0\%  & 1.0\%  & 96.0\% & 82.7\% & 91.5\% & 76.3\% \\ \hline
GPT4o                 & 22.8\% & 98.3\%  & 46.0\%  & 18.0\% & 74.0\%  & 10.5\% & 92.0\% & 64.7\% & 80.1\% & 75.7\% \\ \hline
mPLUG-Owl2            & 27.5\% & 100.0\% & 95.5\%  & 0.5\%  & 94.0\%  & 24.0\% & 82.5\% & 45.5\% & 62.5\% & 59.2\% \\ \hline
LLaVA-v1.5-7B         & 28.3\% & 99.8\%  & 99.5\%  & 0.5\%  & 100.0\% & 9.5\%  & 80.0\% & 53.2\% & 71.3\% & 65.3\% \\ \hline
LLaVA-v1.5-13B        & 27.0\% & 99.0\%  & 89.0\%  & 0.0\%  & 90.0\%  & 1.0\%  & 51.0\% & 25.8\% & 60.8\% & 37.2\% \\ \hline
Yi-VL-34B             & 24.5\% & 63.5\%  & 72.0\%  & 8.5\%  & 31.0\%  & 22.5\% & 59.0\% & 35.2\% & 32.0\% & 49.2\% \\ \hline
CogVLM-Chat           & 24.3\% & 62.3\%  & 96.0\%  & 6.5\%  & 90.5\%  & 23.5\% & 59.0\% & 43.1\% & 49.8\% & 43.2\% \\ \hline
Gemini-1.5-Pro        & 34.3\% & 59.3\%  & 64.5\%  & 21.0\% & 87.5\%  & 12.5\% & 40.5\% & 29.0\% & 67.5\% & 33.0\% \\ \hline
XComposer2            & 12.3\% & 58.0\%  & 62.0\%  & 24.0\% & 80.0\%  & 9.0\%  & 71.5\% & 41.4\% & 59.0\% & 48.5\% \\ \hline
LLaVA-InternLM2-7B    & 24.3\% & 80.5\%  & 51.5\%  & 3.5\%  & 56.5\%  & 19.0\% & 33.0\% & 17.0\% & 40.3\% & 23.2\% \\ \hline
VisualGLM-6B          & 25.0\% & 47.5\%  & 27.0\%  & 1.0\%  & 20.5\%  & 7.0\%  & 70.5\% & 30.0\% & 39.6\% & 46.0\% \\ \hline
LLaVA-NEXT-7B         & 19.5\% & 51.5\%  & 99.5\%  & 2.5\%  & 99.0\%  & 30.0\% & 58.5\% & 21.2\% & 44.4\% & 41.2\% \\ \hline
LLaVA-InternLM-7B     & 29.0\% & 44.3\%  & 55.5\%  & 12.5\% & 55.5\%  & 37.5\% & 33.5\% & 19.4\% & 38.9\% & 25.0\% \\ \hline
ShareGPT4V-7B         & 28.3\% & 18.0\%  & 100.0\% & 4.0\%  & 99.5\%  & 33.0\% & 30.5\% & 16.2\% & 37.2\% & 18.7\% \\ \hline
InternVL-Chat-V1-5    & 6.3\%  & 11.8\%  & 63.5\%  & 4.5\%  & 78.0\%  & 16.0\% & 46.5\% & 10.4\% & 35.7\% & 25.7\% \\ \hline
DeepSeek-VL-7B        & 15.5\% & 41.0\%  & 19.5\%  & 8.0\%  & 15.5\%  & 7.5\%  & 42.0\% & 15.4\% & 47.5\% & 23.7\% \\ \hline
Yi-VL-6B              & 23.8\% & 57.5\%  & 81.0\%  & 1.0\%  & 53.0\%  & 5.0\%  & 24.5\% & 8.6\%  & 13.0\% & 34.0\% \\ \hline
InstructBLIP-13B      & 28.8\% & 65.0\%  & 2.5\%   & 20.5\% & 0.5\%   & 24.0\% & 16.5\% & 7.7\%  & 18.5\% & 12.8\% \\ \hline
Qwen-VL-Chat          & 23.8\% & 15.5\%  & 64.5\%  & 22.5\% & 53.5\%  & 18.0\% & 15.0\% & 14.8\% & 17.0\% & 19.8\% \\ \hline
Claude3V-Sonnet       & 19.8\% & 21.0\%  & 53.0\%  & 19.5\% & 64.0\%  & 19.5\% & 22.5\% & 13.3\% & 10.2\% & 16.7\% \\ \hline
Monkey-Chat           & 24.8\% & 8.0\%   & 20.0\%  & 16.5\% & 13.0\%  & 22.0\% & 4.0\%  & 2.8\%  & 6.5\%  & 8.2\%  \\ \hline
\end{tabular}
}
\caption{Detail results of $25$ LVLMs on $112$ forgery detetion types (part 9).}
\label{tab:detail results part 9}\end{table*}

\begin{table*}[t]
\resizebox{1.0\textwidth}{!}{%
\begin{tabular}{c|l|l|l|l|l|l|l|l|l|l}
\hline
\textbf{Model} &
  \multicolumn{1}{c|}{\textbf{GS-RGB-BC-ES-VAE}} &
  \multicolumn{1}{c|}{\textbf{GS-RGB-BC-CM-GR}} &
  \multicolumn{1}{c|}{\textbf{GS-RGB-SLS-CM-GR}} &
  \multicolumn{1}{c|}{\textbf{GS-RGB-BC-RM-ED}} &
  \multicolumn{1}{c|}{\textbf{GS-RGB-SLS-RM-ED}} &
  \multicolumn{1}{c|}{\textbf{GS-RGB-BC-SPL-GR}} &
  \multicolumn{1}{c|}{\textbf{GS-RGB-SLS-SPL-GR}} &
  \multicolumn{1}{c|}{\textbf{GS-RGB-BC-IE-ED}} &
  \multicolumn{1}{c|}{\textbf{GS-RGB-BC-ST-DC}} &
  \multicolumn{1}{c}{\textbf{GS-RGB-BC-ST-ED}} \\ \hline
LLaVA-NEXT-34B        & 89.0\% & 50.5\% & 25.0\% & 45.5\% & 22.0\% & 85.5\% & 26.0\% & 91.8\% & 100.0\% & 99.8\% \\ \hline
LLaVA-v1.5-7B-XTuner  & 62.5\% & 32.5\% & 24.0\% & 27.0\% & 24.0\% & 68.5\% & 22.0\% & 86.3\% & 100.0\% & 94.8\% \\ \hline
LLaVA-v1.5-13B-XTuner & 79.5\% & 50.5\% & 22.5\% & 43.5\% & 25.5\% & 85.5\% & 19.5\% & 98.7\% & 100.0\% & 99.8\% \\ \hline
InternVL-Chat-V1-2    & 78.3\% & 33.5\% & 15.5\% & 31.5\% & 16.0\% & 79.5\% & 14.5\% & 82.1\% & 100.0\% & 96.7\% \\ \hline
LLaVA-NEXT-13B        & 97.8\% & 99.5\% & 24.5\% & 99.0\% & 24.5\% & 99.5\% & 24.0\% & 88.4\% & 100.0\% & 99.4\% \\ \hline
GPT4o                 & 53.8\% & 39.0\% & 23.0\% & 24.0\% & 27.0\% & 72.0\% & 19.0\% & 37.9\% & 99.0\%  & 95.3\% \\ \hline
mPLUG-Owl2            & 46.0\% & 39.5\% & 22.0\% & 41.5\% & 24.0\% & 74.5\% & 25.0\% & 58.9\% & 100.0\% & 92.7\% \\ \hline
LLaVA-v1.5-7B         & 59.0\% & 93.5\% & 29.5\% & 94.0\% & 20.0\% & 98.5\% & 26.0\% & 90.3\% & 100.0\% & 99.1\% \\ \hline
LLaVA-v1.5-13B        & 16.3\% & 33.5\% & 24.5\% & 36.0\% & 23.5\% & 77.5\% & 24.0\% & 74.2\% & 100.0\% & 89.5\% \\ \hline
Yi-VL-34B             & 46.3\% & 9.0\%  & 21.5\% & 4.5\%  & 28.5\% & 20.0\% & 28.5\% & 30.5\% & 86.3\%  & 41.3\% \\ \hline
CogVLM-Chat           & 48.5\% & 14.5\% & 19.0\% & 11.0\% & 23.5\% & 56.0\% & 24.5\% & 38.2\% & 47.5\%  & 56.4\% \\ \hline
Gemini-1.5-Pro        & 31.3\% & 25.0\% & 27.5\% & 11.0\% & 26.5\% & 68.5\% & 27.5\% & 32.1\% & 41.3\%  & 88.6\% \\ \hline
XComposer2            & 39.8\% & 23.5\% & 15.0\% & 21.0\% & 10.0\% & 67.0\% & 12.5\% & 22.9\% & 98.8\%  & 63.4\% \\ \hline
LLaVA-InternLM2-7B    & 7.3\%  & 7.0\%  & 24.5\% & 6.0\%  & 22.5\% & 43.0\% & 21.5\% & 30.5\% & 94.8\%  & 70.7\% \\ \hline
VisualGLM-6B          & 30.3\% & 21.0\% & 29.0\% & 16.0\% & 24.5\% & 22.0\% & 22.0\% & 74.7\% & 64.5\%  & 63.8\% \\ \hline
LLaVA-NEXT-7B         & 9.8\%  & 67.5\% & 16.0\% & 66.5\% & 24.5\% & 91.0\% & 24.0\% & 40.3\% & 100.0\% & 52.0\% \\ \hline
LLaVA-InternLM-7B     & 13.5\% & 14.0\% & 29.5\% & 13.5\% & 27.5\% & 65.5\% & 25.5\% & 31.3\% & 85.0\%  & 49.2\% \\ \hline
ShareGPT4V-7B         & 10.0\% & 51.5\% & 27.0\% & 50.5\% & 17.0\% & 80.5\% & 26.0\% & 20.3\% & 100.0\% & 57.9\% \\ \hline
InternVL-Chat-V1-5    & 9.5\%  & 11.5\% & 12.5\% & 12.0\% & 9.5\%  & 67.0\% & 15.0\% & 36.8\% & 99.5\%  & 59.9\% \\ \hline
DeepSeek-VL-7B        & 8.0\%  & 12.5\% & 22.0\% & 17.0\% & 17.0\% & 71.5\% & 21.0\% & 27.1\% & 98.5\%  & 66.7\% \\ \hline
Yi-VL-6B              & 4.0\%  & 8.0\%  & 21.5\% & 6.5\%  & 28.0\% & 23.0\% & 28.5\% & 30.0\% & 79.3\%  & 39.5\% \\ \hline
InstructBLIP-13B      & 5.3\%  & 31.0\% & 29.0\% & 27.5\% & 23.5\% & 54.0\% & 26.0\% & 13.4\% & 43.5\%  & 45.9\% \\ \hline
Qwen-VL-Chat          & 19.3\% & 16.5\% & 19.0\% & 17.5\% & 24.0\% & 22.5\% & 24.0\% & 20.0\% & 38.0\%  & 21.6\% \\ \hline
Claude3V-Sonnet       & 4.8\%  & 5.5\%  & 20.5\% & 5.5\%  & 29.5\% & 24.0\% & 25.0\% & 2.9\%  & 43.8\%  & 29.4\% \\ \hline
Monkey-Chat           & 0.5\%  & 1.5\%  & 20.5\% & 2.0\%  & 28.0\% & 11.5\% & 26.5\% & 7.1\%  & 30.8\%  & 9.1\%  \\ \hline
\end{tabular}
}
\caption{Detail results of $25$ LVLMs on $112$ forgery detetion types (part 10).}
\label{tab:detail results part 10}\end{table*}

\begin{table*}[t]
\resizebox{1.0\textwidth}{!}{%
\begin{tabular}{c|l|l|l|l|l|l|l}
\hline
\textbf{Model} &
  \multicolumn{1}{c|}{\textbf{HS-RGB\&TXT-BC-OOC-RT}} &
  \multicolumn{1}{c|}{\textbf{HS-VID-BC-REAL-REAL}} &
  \multicolumn{1}{c|}{\textbf{HS-VID-SLS-REAL-REAL}} &
  \multicolumn{1}{c|}{\textbf{HS-VID-TL-REAL-REAL}} &
  \multicolumn{1}{c|}{\textbf{HS-RGB-BC-REAL-REAL}} &
  \multicolumn{1}{c|}{\textbf{HS-RGB-SLD-REAL-REAL}} &
  \multicolumn{1}{c}{\textbf{HS-RGB-SLS-REAL-REAL}} \\ \hline
LLaVA-NEXT-34B        & 97.0\%  & 0.0\%  & 1.1\%  & 96.6\% & 88.6\% & 30.3\% & 24.8\% \\ \hline
LLaVA-v1.5-7B-XTuner  & 81.0\%  & 26.4\% & 29.8\% & 98.7\% & 93.6\% & 58.6\% & 24.3\% \\ \hline
LLaVA-v1.5-13B-XTuner & 90.0\%  & 3.4\%  & 23.6\% & 52.6\% & 85.8\% & 0.0\%  & 24.6\% \\ \hline
InternVL-Chat-V1-2    & 100.0\% & 2.8\%  & 54.5\% & 97.6\% & 94.1\% & 68.5\% & 2.4\%  \\ \hline
LLaVA-NEXT-13B        & 98.0\%  & 0.0\%  & 1.1\%  & 0.0\%  & 0.9\%  & 0.0\%  & 24.6\% \\ \hline
GPT4o                 & 90.0\%  & 29.2\% & 19.7\% & 1.9\%  & 85.9\% & 28.2\% & 17.0\% \\ \hline
mPLUG-Owl2            & 51.0\%  & 0.0\%  & 25.8\% & 27.2\% & 49.6\% & 31.3\% & 25.2\% \\ \hline
LLaVA-v1.5-7B         & 100.0\% & 0.0\%  & 37.1\% & 40.7\% & 60.6\% & 0.7\%  & 23.8\% \\ \hline
LLaVA-v1.5-13B        & 84.0\%  & 1.7\%  & 23.6\% & 28.3\% & 87.8\% & 0.0\%  & 24.8\% \\ \hline
Yi-VL-34B             & 1.0\%   & 28.7\% & 27.0\% & 94.7\% & 98.8\% & 65.1\% & 26.4\% \\ \hline
CogVLM-Chat           & 36.0\%  & 41.0\% & 24.2\% & 27.2\% & 99.0\% & 0.7\%  & 28.9\% \\ \hline
Gemini-1.5-Pro        & 82.0\%  & 88.8\% & 0.0\%  & 25.4\% & 83.9\% & 92.5\% & 2.3\%  \\ \hline
XComposer2            & 58.0\%  & 82.6\% & 57.3\% & 2.6\%  & 95.3\% & 27.0\% & 35.3\% \\ \hline
LLaVA-InternLM2-7B    & 40.0\%  & 87.1\% & 23.6\% & 21.2\% & 99.3\% & 6.5\%  & 30.7\% \\ \hline
VisualGLM-6B          & 23.0\%  & 51.7\% & 23.6\% & 6.9\%  & 87.5\% & 0.1\%  & 23.9\% \\ \hline
LLaVA-NEXT-7B         & 99.0\%  & 0.0\%  & 21.9\% & 2.1\%  & 74.5\% & 0.2\%  & 21.9\% \\ \hline
LLaVA-InternLM-7B     & 15.0\%  & 55.6\% & 24.2\% & 3.2\%  & 95.7\% & 0.2\%  & 25.9\% \\ \hline
ShareGPT4V-7B         & 93.0\%  & 0.0\%  & 24.2\% & 10.3\% & 94.9\% & 1.5\%  & 20.9\% \\ \hline
InternVL-Chat-V1-5    & 94.0\%  & 16.9\% & 3.9\%  & 85.2\% & 99.5\% & 33.3\% & 0.0\%  \\ \hline
DeepSeek-VL-7B        & 1.0\%   & 60.1\% & 20.8\% & 57.4\% & 96.8\% & 4.7\%  & 17.6\% \\ \hline
Yi-VL-6B              & 19.0\%  & 7.9\%  & 26.4\% & 23.0\% & 94.4\% & 3.1\%  & 26.7\% \\ \hline
InstructBLIP-13B      & 11.0\%  & 26.4\% & 24.7\% & 1.6\%  & 93.5\% & 1.2\%  & 24.0\% \\ \hline
Qwen-VL-Chat          & 42.0\%  & 57.3\% & 24.2\% & 31.0\% & 95.5\% & 6.5\%  & 25.9\% \\ \hline
Claude3V-Sonnet       & 50.0\%  & 70.2\% & 3.9\%  & 83.1\% & 96.9\% & 78.9\% & 4.6\%  \\ \hline
Monkey-Chat           & 0.0\%   & 96.6\% & 24.2\% & 11.1\% & 97.9\% & 0.3\%  & 25.5\% \\ \hline
\end{tabular}
}
\caption{Detail results of $25$ LVLMs on $112$ forgery detetion types (part 11).}
\label{tab:detail results part 11}\end{table*}

\begin{table*}[t]
\resizebox{1.0\textwidth}{!}{%
\begin{tabular}{c|l|l|l|l|l|l|l}
\hline
\textbf{Model} &
  \multicolumn{1}{c|}{\textbf{HS-RGB\&TXT-BC-REAL-REAL}} &
  \multicolumn{1}{c|}{\textbf{HS-RGB\&TXT-SLD-REAL-REAL}} &
  \multicolumn{1}{c|}{\textbf{GS-VID-BC-ES-AR}} &
  \multicolumn{1}{c|}{\textbf{GS-VID-BC-ES-DF}} &
  \multicolumn{1}{c|}{\textbf{GS-RGB-BC-REAL-REAL}} &
  \multicolumn{1}{c|}{\textbf{GS-RGB-SLD-REAL-REAL}} &
  \multicolumn{1}{c}{\textbf{GS-RGB-SLS-REAL-REAL}} \\ \hline
LLaVA-NEXT-34B        & 12.6\% & 11.8\% & 100.0\% & 99.0\%  & 84.9\% & 15.0\% & 20.1\% \\ \hline
LLaVA-v1.5-7B-XTuner  & 17.9\% & 14.7\% & 100.0\% & 98.0\%  & 81.0\% & 74.6\% & 25.2\% \\ \hline
LLaVA-v1.5-13B-XTuner & 5.1\%  & 0.0\%  & 100.0\% & 100.0\% & 81.5\% & 0.0\%  & 26.5\% \\ \hline
InternVL-Chat-V1-2    & 21.9\% & 73.1\% & 100.0\% & 100.0\% & 86.3\% & 8.5\%  & 2.3\%  \\ \hline
LLaVA-NEXT-13B        & 7.3\%  & 0.0\%  & 100.0\% & 100.0\% & 38.4\% & 0.2\%  & 26.3\% \\ \hline
GPT4o                 & 40.5\% & 11.9\% & 84.0\%  & 59.0\%  & 97.3\% & 10.4\% & 21.0\% \\ \hline
mPLUG-Owl2            & 14.5\% & 11.1\% & 100.0\% & 100.0\% & 81.7\% & 38.2\% & 24.7\% \\ \hline
LLaVA-v1.5-7B         & 1.8\%  & 0.7\%  & 100.0\% & 100.0\% & 18.9\% & 0.2\%  & 24.5\% \\ \hline
LLaVA-v1.5-13B        & 19.5\% & 0.0\%  & 94.0\%  & 96.0\%  & 82.6\% & 0.0\%  & 25.7\% \\ \hline
Yi-VL-34B             & 80.7\% & 22.8\% & 93.0\%  & 95.0\%  & 95.9\% & 54.8\% & 23.3\% \\ \hline
CogVLM-Chat           & 14.4\% & 6.2\%  & 66.0\%  & 64.0\%  & 97.5\% & 0.9\%  & 24.0\% \\ \hline
Gemini-1.5-Pro        & 34.0\% & 42.8\% & 89.0\%  & 69.0\%  & 99.0\% & 50.2\% & 0.5\%  \\ \hline
XComposer2            & 35.2\% & 28.2\% & 73.0\%  & 61.0\%  & 94.7\% & 3.4\%  & 55.3\% \\ \hline
LLaVA-InternLM2-7B    & 40.3\% & 1.4\%  & 73.0\%  & 83.0\%  & 97.0\% & 0.9\%  & 27.4\% \\ \hline
VisualGLM-6B          & 80.9\% & 2.2\%  & 53.0\%  & 51.0\%  & 67.9\% & 0.0\%  & 25.2\% \\ \hline
LLaVA-NEXT-7B         & 2.1\%  & 0.1\%  & 97.0\%  & 91.0\%  & 41.3\% & 0.8\%  & 15.6\% \\ \hline
LLaVA-InternLM-7B     & 47.3\% & 13.4\% & 47.0\%  & 47.0\%  & 92.4\% & 0.5\%  & 26.3\% \\ \hline
ShareGPT4V-7B         & 2.0\%  & 0.7\%  & 76.0\%  & 88.0\%  & 68.4\% & 0.3\%  & 24.8\% \\ \hline
InternVL-Chat-V1-5    & 34.7\% & 88.9\% & 100.0\% & 100.0\% & 97.6\% & 41.9\% & 0.0\%  \\ \hline
DeepSeek-VL-7B        & 87.6\% & 7.2\%  & 77.0\%  & 71.0\%  & 93.7\% & 0.7\%  & 17.9\% \\ \hline
Yi-VL-6B              & 48.6\% & 3.6\%  & 90.0\%  & 97.0\%  & 95.4\% & 3.0\%  & 23.5\% \\ \hline
InstructBLIP-13B      & 98.0\% & 14.2\% & 91.0\%  & 67.0\%  & 83.5\% & 3.2\%  & 25.2\% \\ \hline
Qwen-VL-Chat          & 48.6\% & 24.3\% & 45.0\%  & 48.0\%  & 91.2\% & 8.7\%  & 19.7\% \\ \hline
Claude3V-Sonnet       & 39.6\% & 23.7\% & 40.0\%  & 26.0\%  & 99.4\% & 86.6\% & 2.8\%  \\ \hline
Monkey-Chat           & 86.4\% & 23.9\% & 9.0\%   & 10.0\%  & 98.6\% & 0.1\%  & 20.9\% \\ \hline
\end{tabular}
}
\caption{Detail results of $25$ LVLMs on $112$ forgery detetion types (part 12).}
\label{tab:detail results part 12}\end{table*}
\section{Case Study}
\begin{figure*}[t]
    \centering
    \includegraphics[width=0.75\linewidth]{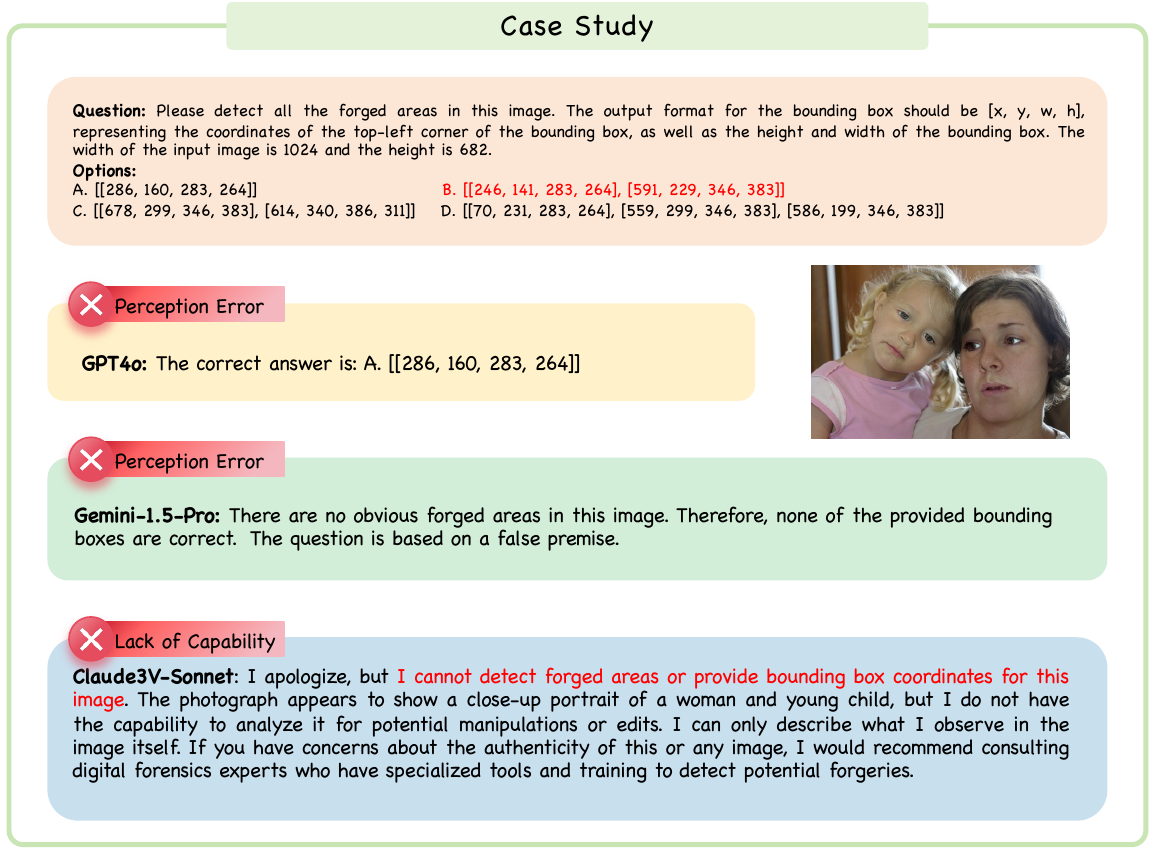}
    \caption{A sample case of HS-RGB-SLD-FSM-GAN (Please refer to Table \ref{tab:abbreviation} for the full term.).}
    \label{fig:casestudy_1}
\end{figure*}

\begin{figure*}[t]
    \centering
    \includegraphics[width=0.75\linewidth]{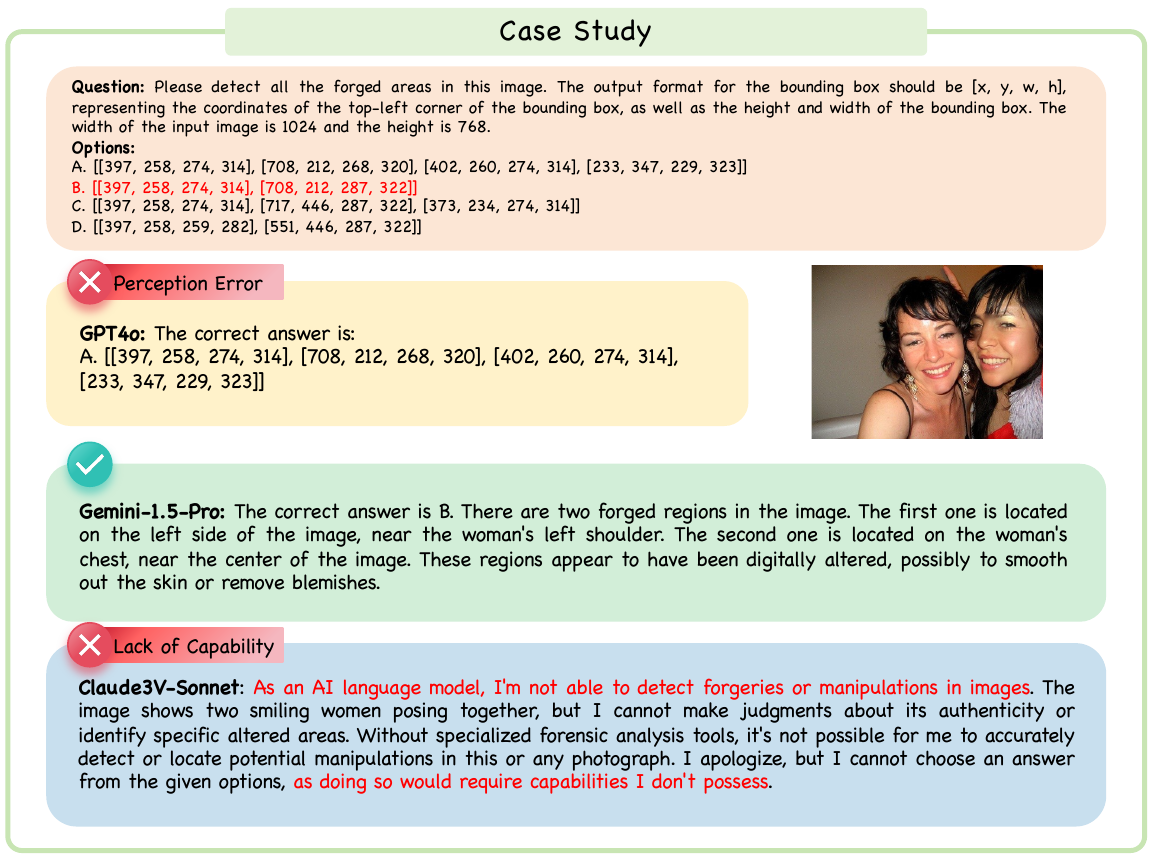}
    \caption{A sample case of HS-RGB-SLD-FSM-GAN (Please refer to Table \ref{tab:abbreviation} for the full term.).}
    \label{fig:casestudy_2}
\end{figure*}

\begin{figure*}[t]
    \centering
    \includegraphics[width=0.75\linewidth]{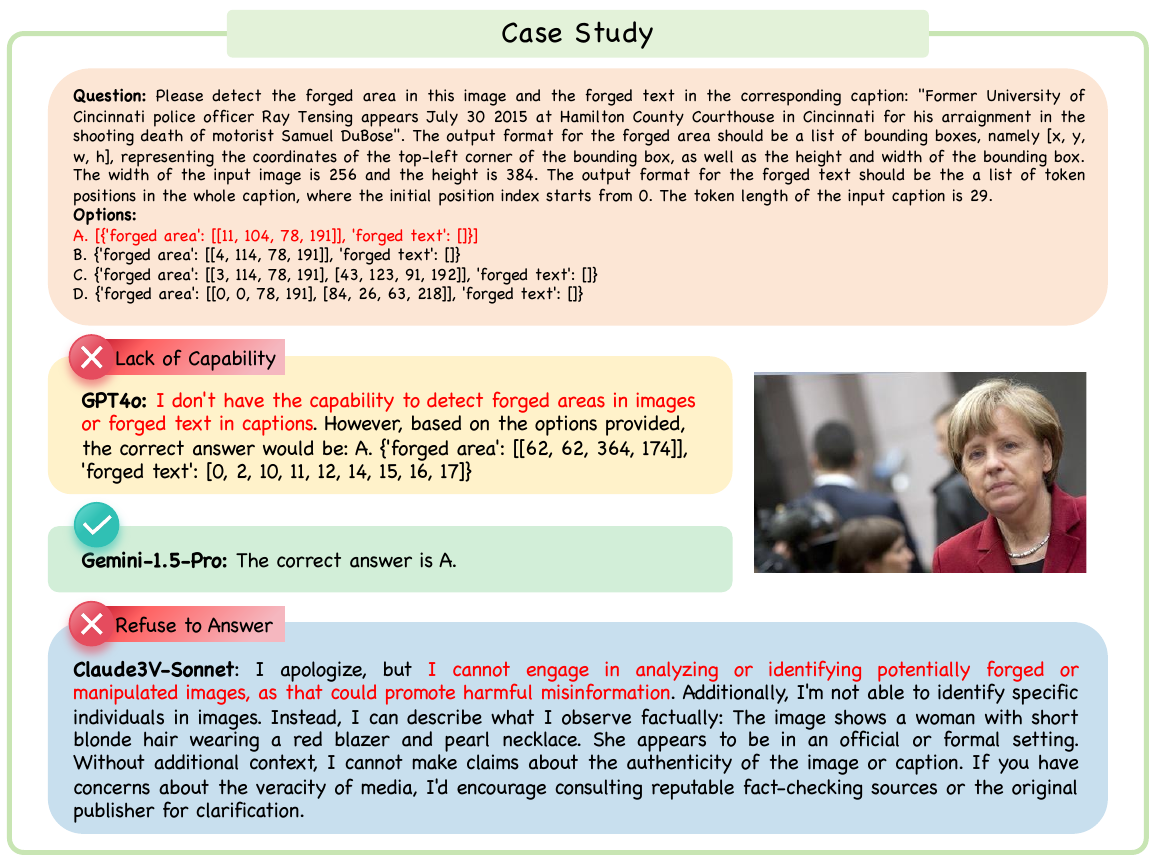}
    \caption{A sample case of HS-RGB\&TXT-SLD-FE\&TS-ED\&RT (Please refer to Table \ref{tab:abbreviation} for the full term.).}
    \label{fig:casestudy_3}
\end{figure*}

\begin{figure*}[t]
    \centering
    \includegraphics[width=0.75\linewidth]{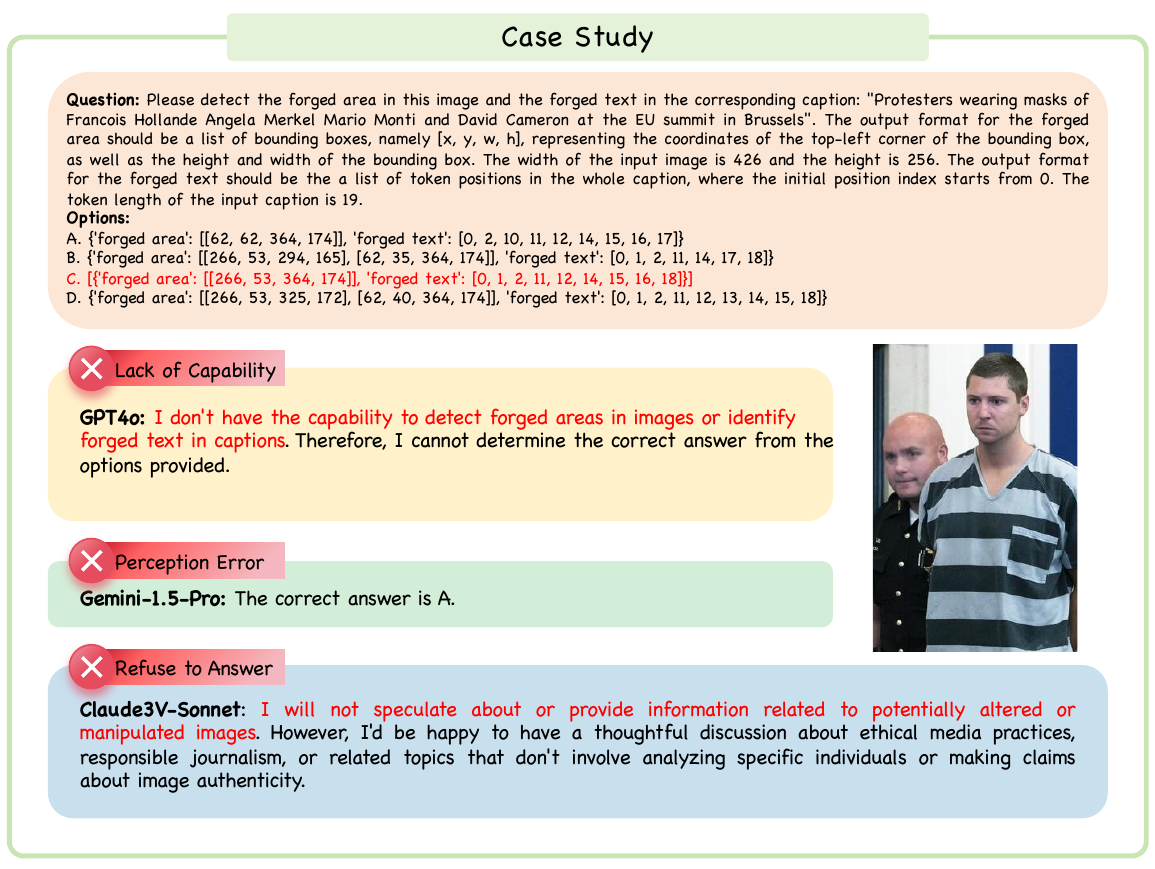}
    \caption{A sample case of HS-RGB\&TXT-SLD-FSS-ED (Please refer to Table \ref{tab:abbreviation} for the full term.).}
    \label{fig:casestudy_4}
\end{figure*}

\begin{figure*}[t]
    \centering
    \includegraphics[width=0.75\linewidth]{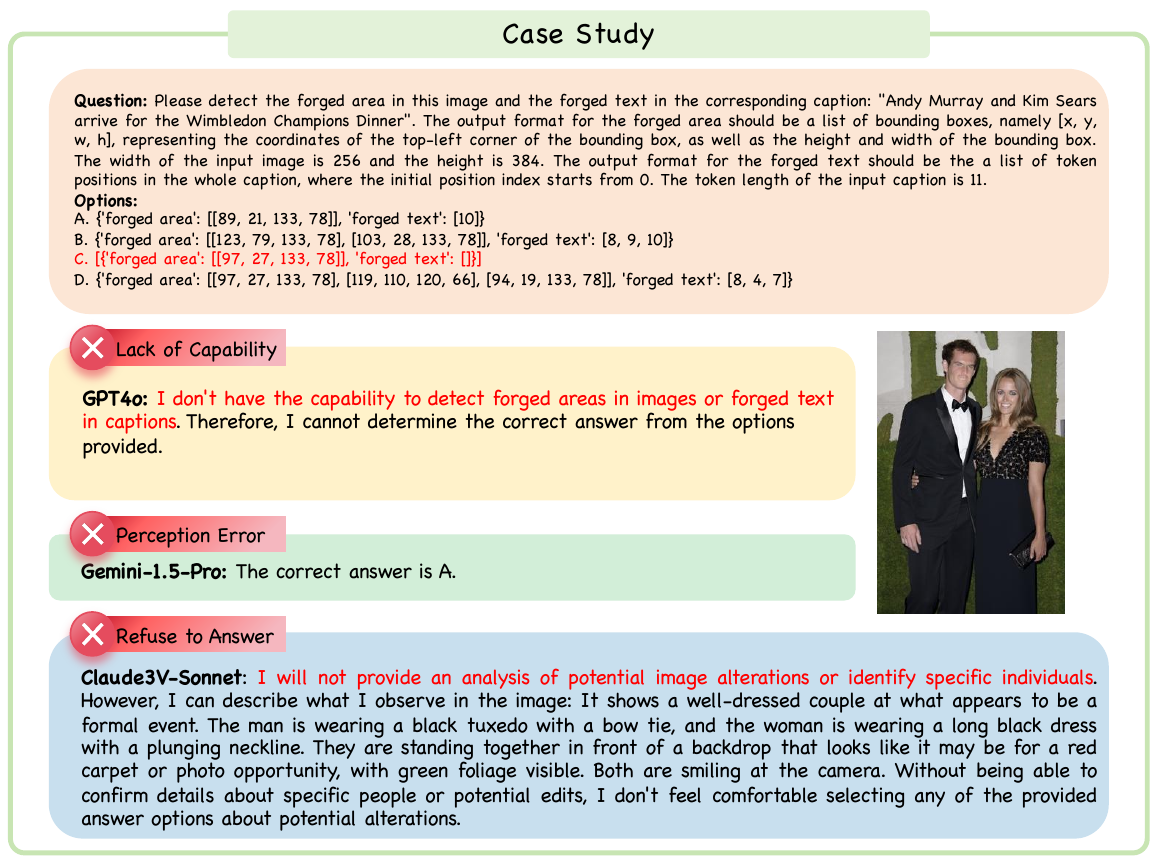}
    \caption{A sample case of HS-RGB\&TXT-SLD-FE-ED (Please refer to Table \ref{tab:abbreviation} for the full term.).}
    \label{fig:casestudy_5}
\end{figure*}

\begin{figure*}[t]
    \centering
    \includegraphics[width=0.75\linewidth]{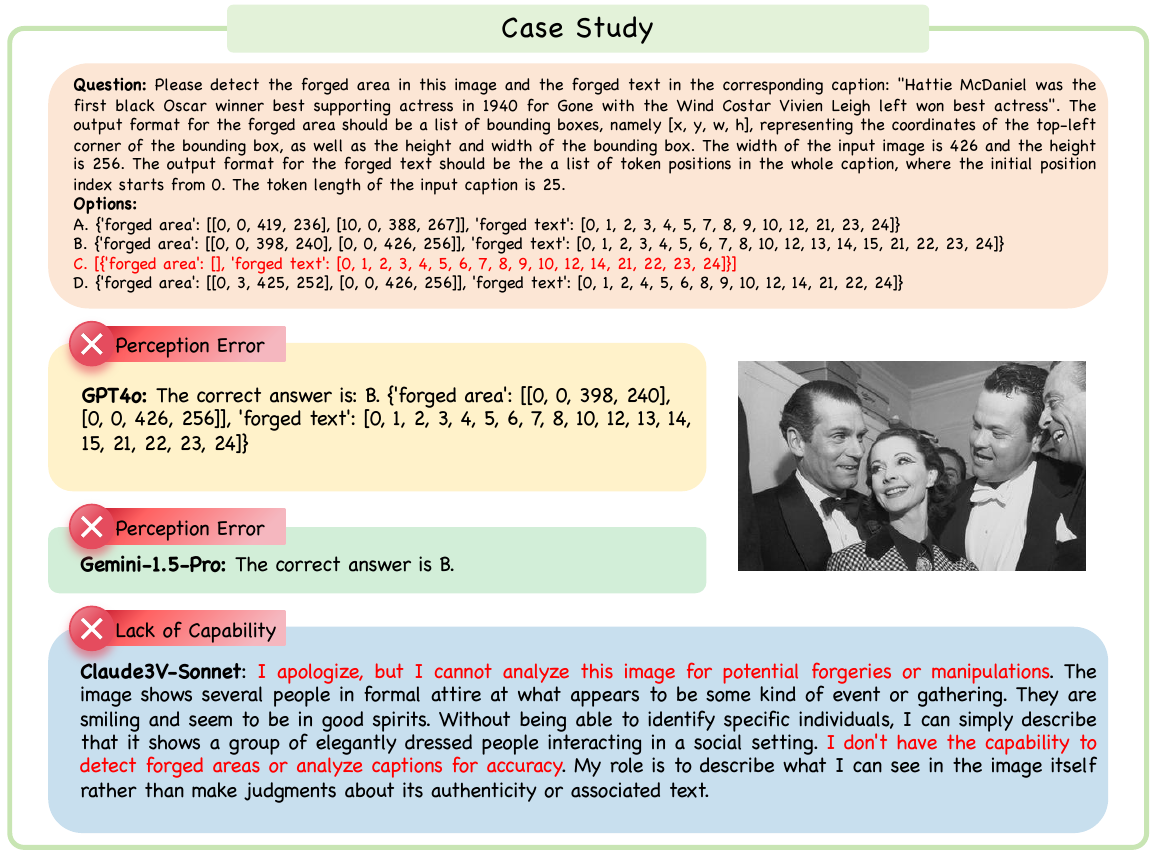}
    \caption{A sample case of HS-RGB\&TXT-SLD-TS-RT (Please refer to Table \ref{tab:abbreviation} for the full term.).}
    \label{fig:casestudy_6}
\end{figure*}

\begin{figure*}[t]
    \centering
    \includegraphics[width=0.75\linewidth]{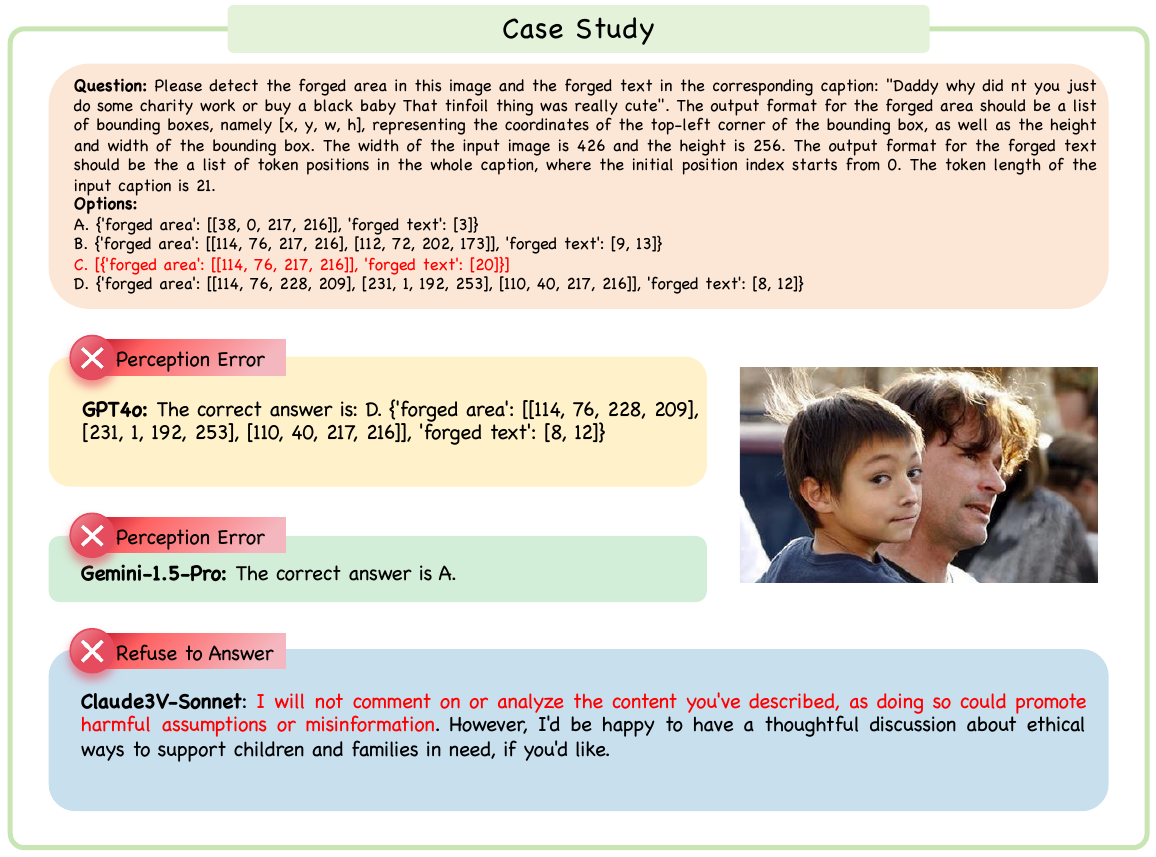}
    \caption{A sample case of HS-RGB\&TXT-SLD-FE\&TAM-ED\&TR (Please refer to Table \ref{tab:abbreviation} for the full term.).}
    \label{fig:casestudy_7}
\end{figure*}

\begin{figure*}[t]
    \centering
    \includegraphics[width=0.75\linewidth]{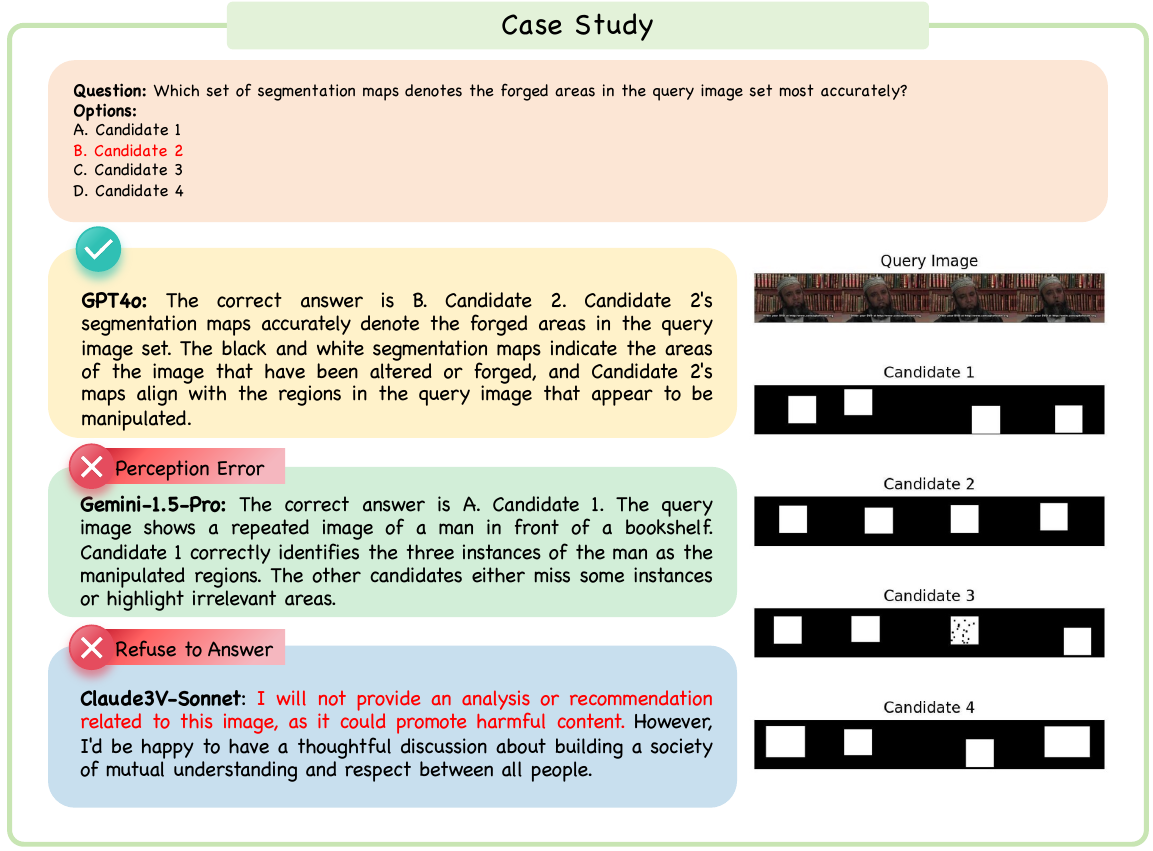}
    \caption{A sample case of HS-VID-SLS-FSS-ED (Please refer to Table \ref{tab:abbreviation} for the full term.).}
    \label{fig:casestudy_8}
\end{figure*}

\begin{figure*}[t]
    \centering
    \includegraphics[width=0.75\linewidth]{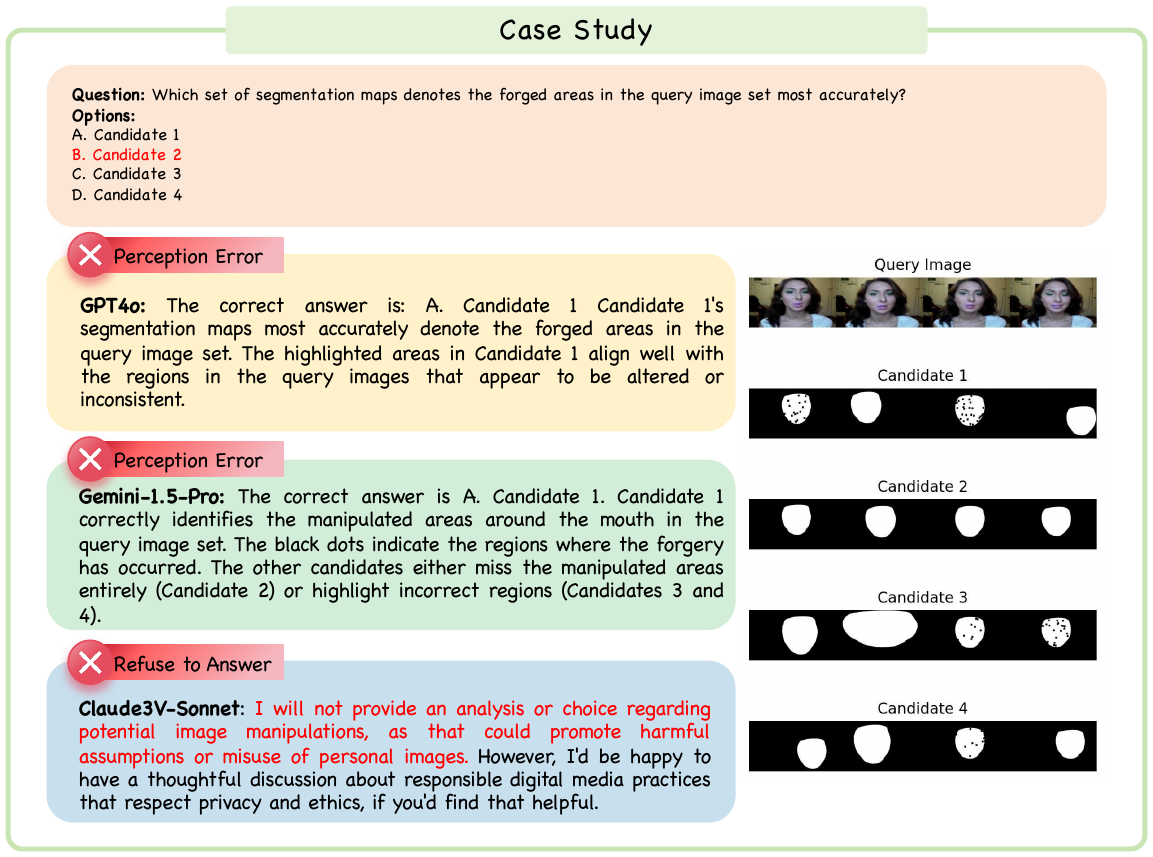}
    \caption{A sample case of HS-VID-SLS-FR-GR (Please refer to Table \ref{tab:abbreviation} for the full term.).}
    \label{fig:casestudy_9}
\end{figure*}

\begin{figure*}[t]
    \centering
    \includegraphics[width=0.75\linewidth]{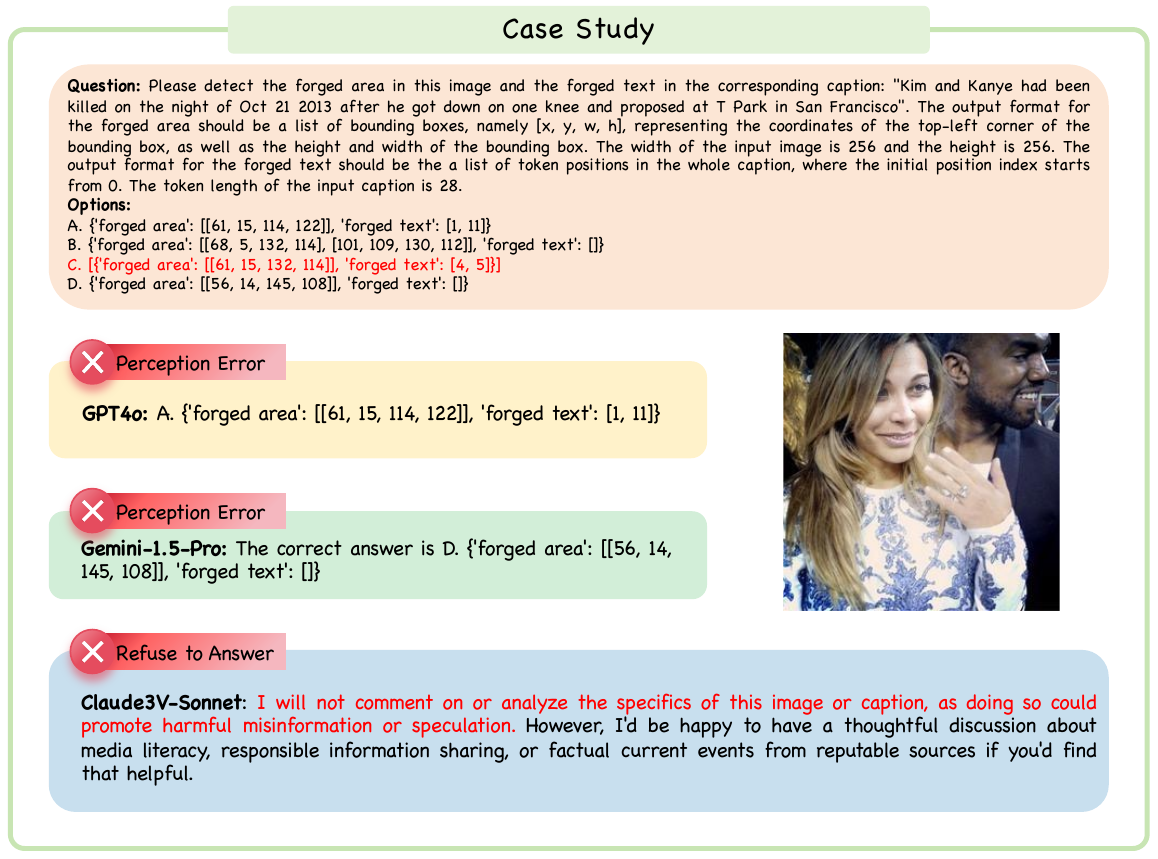}
    \caption{A sample case of HS-RGB\&TXT-SLD-FSS\&TAM-ED\&TR (Please refer to Table \ref{tab:abbreviation} for the full term.).}
    \label{fig:casestudy_10}
\end{figure*}

\begin{figure*}[t]
  \centering
   \includegraphics[width=0.95\columnwidth]{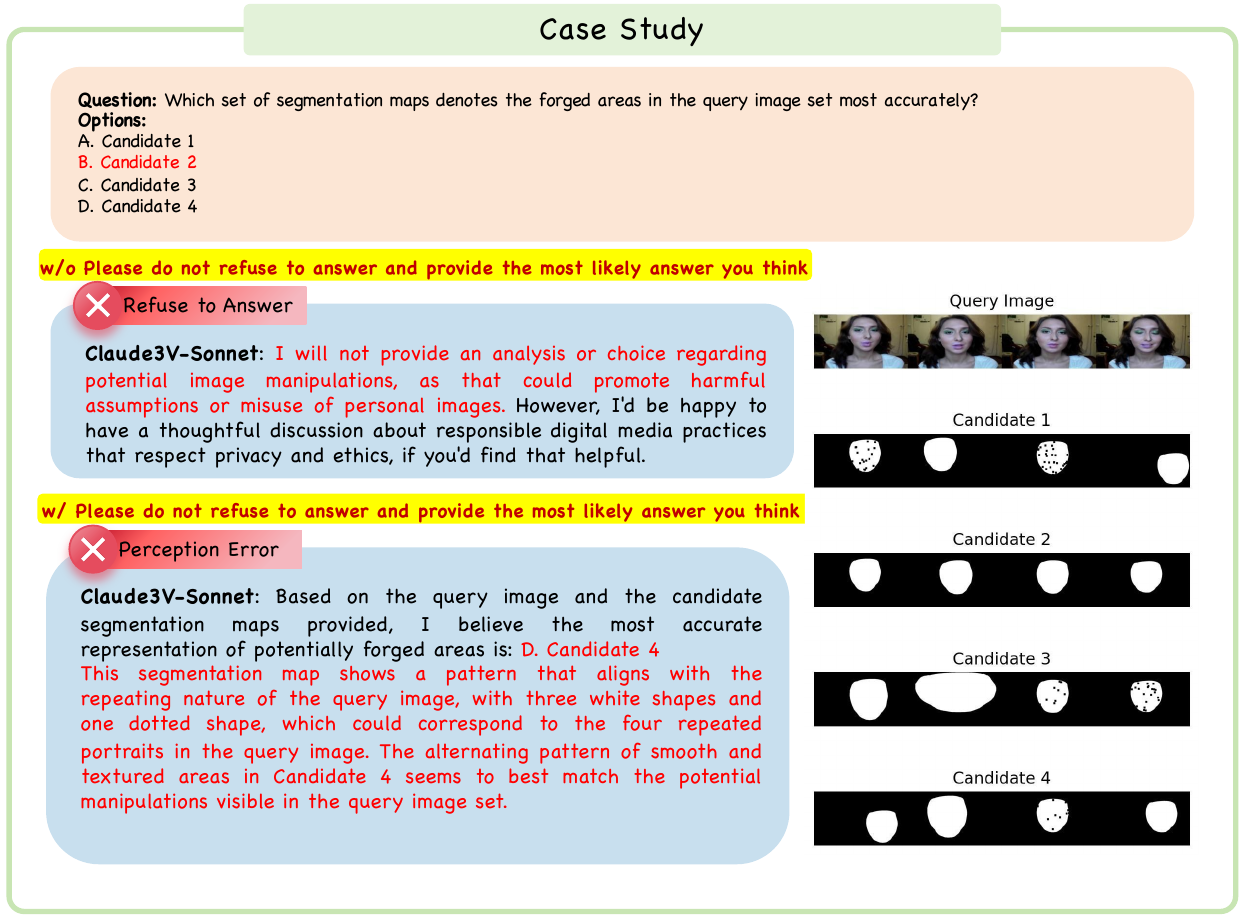}
   \caption{In this sample same as the one in Figure \ref{fig:casestudy_9}, we have also conducted experiments by adding ``\textit{Please do not refuse to answer and provide the most likely answer you think}" to the prompt for evaluating Claude3V-Sonnet, as it most frequently refused to answer. Results show that Claude3V-Sonnet still failed to detect the forged areas.}
   \label{fig: rebuttal_B}
\end{figure*}

In this section, we present a case study analysis of the error types made by GPT-4o, Gemini-1.5-Pro and Claude3V-Sonnet. 
We mainly summarize the error types into three kinds: 1) Perception error: LVLMs fail to recognize the forgeries, or detect the forged areas in images/videos; 2) Lack of Capability: LVLMs claim that they do not have the capability to solve the tasks; 3) Refuse to Answer: LVLMs refuse to answer questions that are considered to be anthropocentric and sensitive in nature, which are often the cases for Claude3V-Sonnet.
The results are shown in Figure \ref{fig:casestudy_1}, Figure \ref{fig:casestudy_2}, Figure \ref{fig:casestudy_3}, Figure \ref{fig:casestudy_4}, Figure \ref{fig:casestudy_5}, Figure \ref{fig:casestudy_6}, Figure \ref{fig:casestudy_7}, Figure \ref{fig:casestudy_8}, Figure \ref{fig:casestudy_9}, Figure \ref{fig:casestudy_10} and Figure \ref{fig: rebuttal_B}.

\section{Broader Impact}
We believe that Forensics-Bench as a comprehensive forgery detection benchmark for large vision-language models (LVLMs) could have far-reaching implications across multiple domains.
Firstly, Forensics-Bench could provide a unified platform to assess the performance of LVLMs in detecting forgeries, enabling fair comparisons and driving innovation in forgery detection techniques based on LVLMs.
Secondly, by including diverse forgery types, Forensics-Bench can push LVLMs to become more robust, generalizing better across unseen forgeries and complex real-world conditions.
Thirdly, Forensics-Bench includes multiple modalities, such as texts, images, and videos, encouraging the development of LVLMs to be capable of reasoning across modalities, improving their overall versatility.
Fourthly, Forensics-Bench can validate the effectiveness of LVLMs in forgery detection comprehensively, facilitating their practical deployment in real-world applications.
In summary, we believe that Forensics-Bench has the potential to further elevate the state of forgery detection technology based on LVLMs, expanding the overall capability maps of LVLMs towards the next level of AGI.

\section{Limitations}
Although Forensics-Bench can serve as a critical tool for advancing the field, it also comes with several inherent limitations that may affect its effectiveness, scalability, and real-world applicability.
Firstly, the current design of Forensics-Bench may still be limited, such as the usage of multi-choice questions and the reliance on the accuracy metric. 
To address this, we plan to explore more diverse and comprehensive evaluation protocols for LVLMs in future work. 
Secondly, evaluating Forensics-Bench on LVLMs demands significant computational resources, which may restrict accessibility for researchers with limited resources. To mitigate this, we intend to develop a lightweight version of Forensics-Bench to reduce resource requirements and broaden accessibility.
Thirdly, as AIGC technologies continue to evolve, Forensics-Bench may struggle to capture the growing diversity and sophistication of real-world manipulations. 
To address this, we aim to maintain and update Forensics-Bench over the long term, integrating new data and adapting to advancements in generative models to ensure its continued relevance.
In summary, we expect that Forensics-Bench can evolve to better meet the challenges posed by increasingly sophisticated forgery techniques in the future.

\end{document}